\documentclass[journal]{IEEEtran}
\ifCLASSINFOpdf
\else
\fi

\usepackage{graphicx}

\usepackage{cite}
\usepackage{comment}
\usepackage{amsmath,amssymb} 
\usepackage{xcolor}
\usepackage[normalem]{ulem}
\usepackage{multirow}
\usepackage{threeparttable}
\usepackage{booktabs}
\usepackage{floatrow}
\newfloatcommand{capbtabbox}{table}[][\FBwidth]
\usepackage{mathtools}
\usepackage{enumitem}
\DeclarePairedDelimiter{\ceil}{\lceil}{\rceil}
\def\figvspace{{\vspace{-3mm}}}

\begin{document}
%
\title{Video Frame Interpolation via Generalized Deformable Convolution}
%
%
%

\author{Zhihao Shi,~
		Xiaohong Liu,~\IEEEmembership{Graduate~Student~Member,~IEEE},
		Kangdi Shi,~
		Linhui Dai,~
		Jun Chen,~\IEEEmembership{Senior Member,~IEEE}
\thanks{Z.~Shi, X.~Liu (corresponding author), K.~Shi, L.~Dai and J.~Chen are with the Department of Electrical and Computer Engineering, McMaster University, Hamilton, ON L8S 4K1, Canada   (e-mail: \{shiz31, liux173, shik9, dail5\}@mcmaster.ca; junchen@ece.mcmaster.ca). This work was supported in part by the Natural Sciences and Engineering Research Council of Canada through a Discovery Grant.}}

\maketitle

\begin{abstract}
Video frame interpolation aims at {synthesizing intermediate frames from nearby source frames while maintaining spatial and temporal consistencies}. The existing {{deep-learning-based}} video frame interpolation methods can be roughly divided into two categories: flow-based methods and kernel-based methods.
The performance of flow-based methods is often jeopardized by
the inaccuracy of flow map estimation due to oversimplified motion models, while that of kernel-based methods tends to be constrained by the rigidity of kernel shape. To address these performance-limiting issues, a novel mechanism named generalized deformable convolution is proposed, which can effectively learn motion information in a data-driven manner and freely select sampling points in space-time. We further develop a new video frame interpolation method based on this mechanism. Our extensive experiments demonstrate that the new method performs favorably against the state-of-the-art, especially when dealing with complex motions. Code is available at https://github.com/zhshi0816/GDConvNet.
\end{abstract}


%
\IEEEpeerreviewmaketitle

\section{Introduction}


In recent years, owing to the hardware development and the availability of large-scale datasets, deep learning has achieved promising results in many computer vision and multimedia tasks \cite{qi2015guest} including, among others, super-resolution \cite{yang2018drfn, he2019mrfn, liu2020exploit}, optical flow estimation \cite{hu2018recurrent, sun2018pwc}, image dehazing \cite{li2019pdr, song2017single}, action recognition \cite{zhu2019cuboid}, and video frame interpolation (VFI) \cite{usman2016frame, ma2008error, lee2020adacof, xue2019video}. VFI is a classic problem in the multimedia area and has received significant attention with the rapid growth of streaming videos. It aims at synthesizing intermediate frames from nearby sources while maintaining spatial and temporal consistencies. VFI has two main use cases; one is to perform error concealment at the decoder side \cite{usman2016frame, ma2008error}, and the other one is to increase the frame rate of a given video for better visual performance \cite{lee2020adacof, xue2019video}. In general, VFI methods can be roughly divided into two categories: flow-based methods and kernel-based methods.

Flow-based methods generate the value of each pixel in the target intermediate frame by finding an associated optical flow. 
Accurate estimation of the flow map is essential for producing desirable VFI results. However, in some cases with complex motions, it is hard to obtain an accurate flow map regardless
whether traditional methods\cite{zhang2019refined, horn1981determining, lucas1981iterative} or deep-learning-based methods \cite{ranjan2017optical,dosovitskiy2015flownet,ilg2017flownet,sun2018pwc} are employed.
Flow-based methods \cite{ma2008error, qi2015guest, jiang2018super, bao2019depth, bao2019memc} typically adopt a linear model with the oversimplified assumption of uniform motion between neighboring frames. Recently, a more sophisticated approach was proposed in \cite{qvi_nips19} for estimating motion trajectories, where the naive linear model is replaced by a more accurate quadratic model that can take advantage of latent motion information by simultaneously exploiting four consecutive frames. Nevertheless, it is conceivable that the complexities and irregularities of real-world motions cannot be completely captured by a simple mathematical model. Moreover, the pixel-level displacement performed in flow-based methods is inherently inadequate for handling diffusion and dispersion effects, especially when such effects are not negligible over the time interval between two consecutive frames.


Kernel-based methods  directly generate the target intermediate frame by applying spatially-adaptive convolution kernels to the given frames. They circumvent the need for flow map estimation and consequently are not susceptible to the associated issues. On the other hand, the rigidity of the kernel shape \cite{niklaus2017video,niklaus2018video} severely limits the types of motions that such methods can handle. Indeed, one may need to choose a very large kernel size to ensure enough coverage, which is highly inefficient. As a partial remedy, reference \cite{bao2019depth} proposes adaptive deployment of convolution kernels guided by flow maps, but nevertheless, the receptive field is still constrained by the predetermined kernel shape. More recently, reference \cite{lee2020adacof} introduces a new approach known as AdaCoF, which utilizes spatially-adaptive deformable convolution (DConv) to select suitable sampling points needed for synthesizing each target pixel. Although this approach eliminates the constraint on the kernel shape in the spatial domain, it does not fully exploit the degrees of freedom available in whole space-time.


\begin{figure}[t]
	\centering
	\begin{minipage}[b]{0.4\linewidth}
		\centering
		\includegraphics[width=\linewidth]{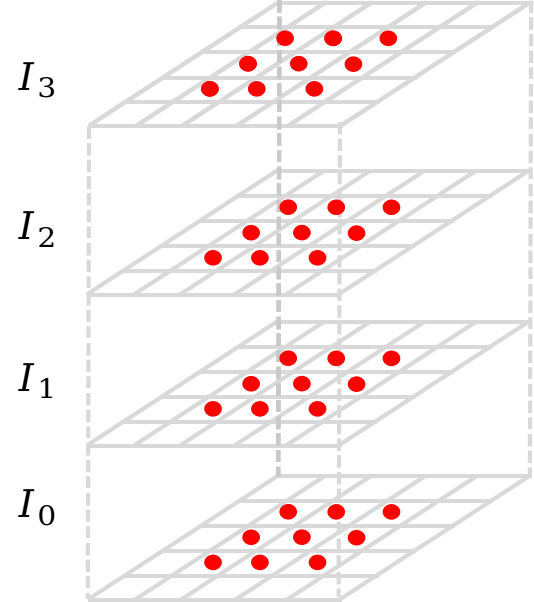}
		\scriptsize{(a)}
	\end{minipage}
	\hspace{5mm}
	\begin{minipage}[b]{0.32
			\linewidth}
		\centering
		\includegraphics[width=\linewidth]{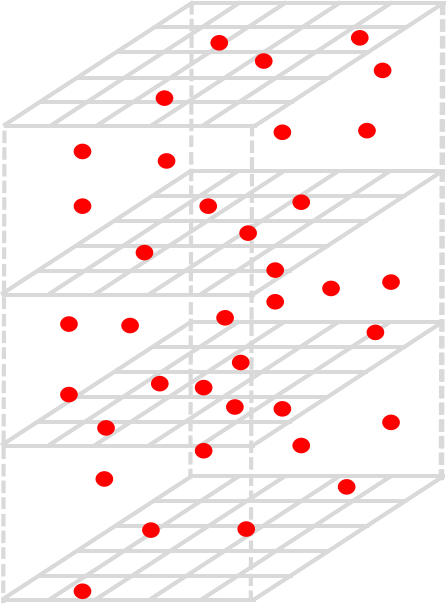}
		\scriptsize{(b)}
	\end{minipage}
	\\
	\vspace{8mm}
	\begin{minipage}[c]{\linewidth}
		\centering
		\includegraphics[width=\linewidth, height=0.15\textheight]{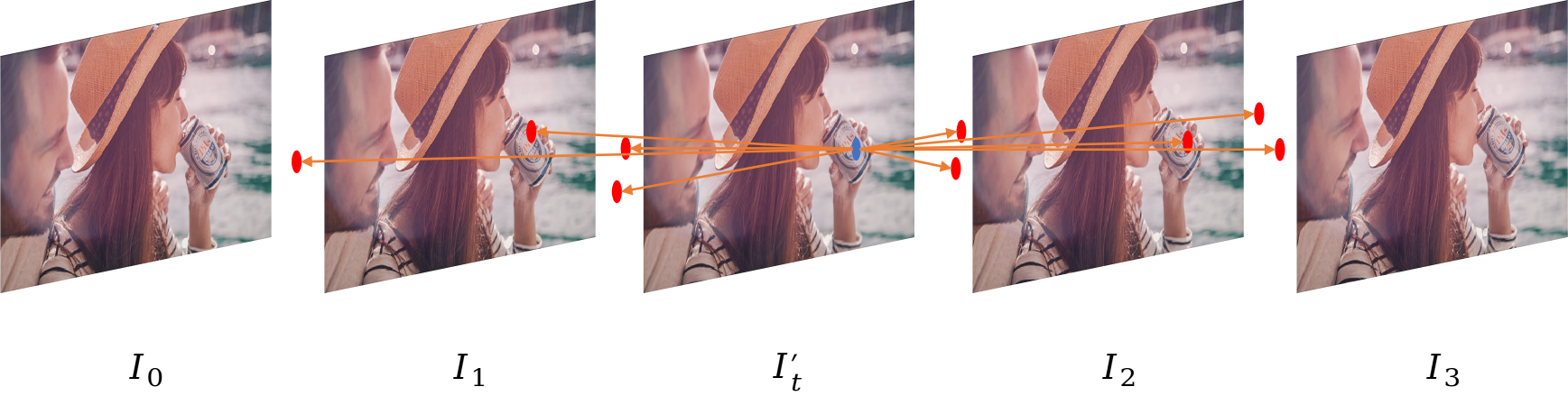}
		\scriptsize{(c)}
	\end{minipage}
	\caption{Illustration of (a) conventional convolution with $3 \times 3 \times 4 =36$ sampling points, (b) GDConv with the same number of sampling points, and (c) visualization of interpolating one frame with GDConv.}
	\label{fig:introduction}
\end{figure}

In summary, flow-based methods and kernel-based methods have their respective limitations. For flow-based methods, even with the aid of sophisticated mathematical models, flow map estimation is still a challenging task due to the intricacies of inter-frame motion trajectories. For kernel-based methods, the predetermined kernel shape lacks the flexibility to cope with a great variety of
motions in terms of range and pattern. While recent innovations have alleviated the rigidity issue to a certain extent, much remains to be done.



The main contribution of this paper is a new approach to VFI that overcomes the hurdles of the aforementioned methods and retains their desirable properties. The key mechanism underlying the proposed approach is generalized deformable convolution (GDConv). An illustration of the difference between conventional convolution and our GDConv in terms of the freedom to select sampling points can be found in Fig.~\ref{fig:introduction}(a) and (b).  Fig.~\ref{fig:introduction}(c) provides a rough idea of how GDConv can be leveraged for VFI: each pixel (e.g., the blue one) in the target intermediate frame is synthesized based on the corresponding sampling points (the red ones). It is worth noting that as the sampling points are allowed to move freely in the continuous space-time, the receptive field of GDConv is basically unconstrained, making it possible to handle all kinds of motions (say, large motions). Moreover, GDConv does not directly adopt a predetermined mathematical model (e.g., linear or quadratic model) for motion estimation. Instead, it is trained to learn real-world motion trajectories and patterns via a data-driven approach. In our design, GDConv is encapsulated in  a generalized deformable convolution module (GDCM). We integrate two GDCMs  with several other modules, including the source extraction module (SEM), the context extraction module (CEM) and the post-processing module (PM), to form a generalized deformable convolution network (GDConvNet) for VFI.
Our extensive experimental results demonstrate that owing to the effective design, the proposed GDConvNet performs favorably against the current state-of-the-art.

\section{Generalized Deformable Convolution Network}

The overall architecture of GDConvNet is shown in Fig.~\ref{fig:architecture}. Given a video clip that consists of $T+1$ source frames\footnote{ For notional simplicity, we assume that the source frames are equally spaced in time. However, the proposed framework can in fact handle the unequal spacing case as well.} $I_{0}$, $I_{1}$, $\cdots$, $I_{T}$, the task of GDConvNet is to synthesize an  intermediate frame $I_{t},~t \in [0,T]$. To this end, it first generates source features through SEM and extracts context maps $C_{0}$, $C_{1}$, $\cdots$, $C_{T}$ through CEM from $I_{0}$, $I_{1}$, $\cdots$, $I_{T}$. The input frames and context maps are then warped by two separate GDCMs according to the same source features. Finally, the warped frame ${I}'_{t}$ and the warped context map ${C}'_{t}$ are fed into the PM to produce the VFI result $\hat{I}_{t}$, which is an approximation\footnote{The accuracy of this approximation can be evaluated by using objective image quality metrics (to be detailed later) or subjective criteria.} of $I_{t}$. The proposed network accomplishes the VFI task by employing a novel GDConv mechanism.
Now we proceed to give a detailed description
of each module in Fig.~\ref{fig:architecture}, with a special emphasis on the GDCM where the GDConv mechanism is realized.

\subsection{Generalized Deformable Convolution Module}
\label{sec:2-A}
\begin{figure}[t]
	\centering
	\begin{minipage}[b]{0.46\linewidth}
		\centering
		\includegraphics[width=\linewidth]{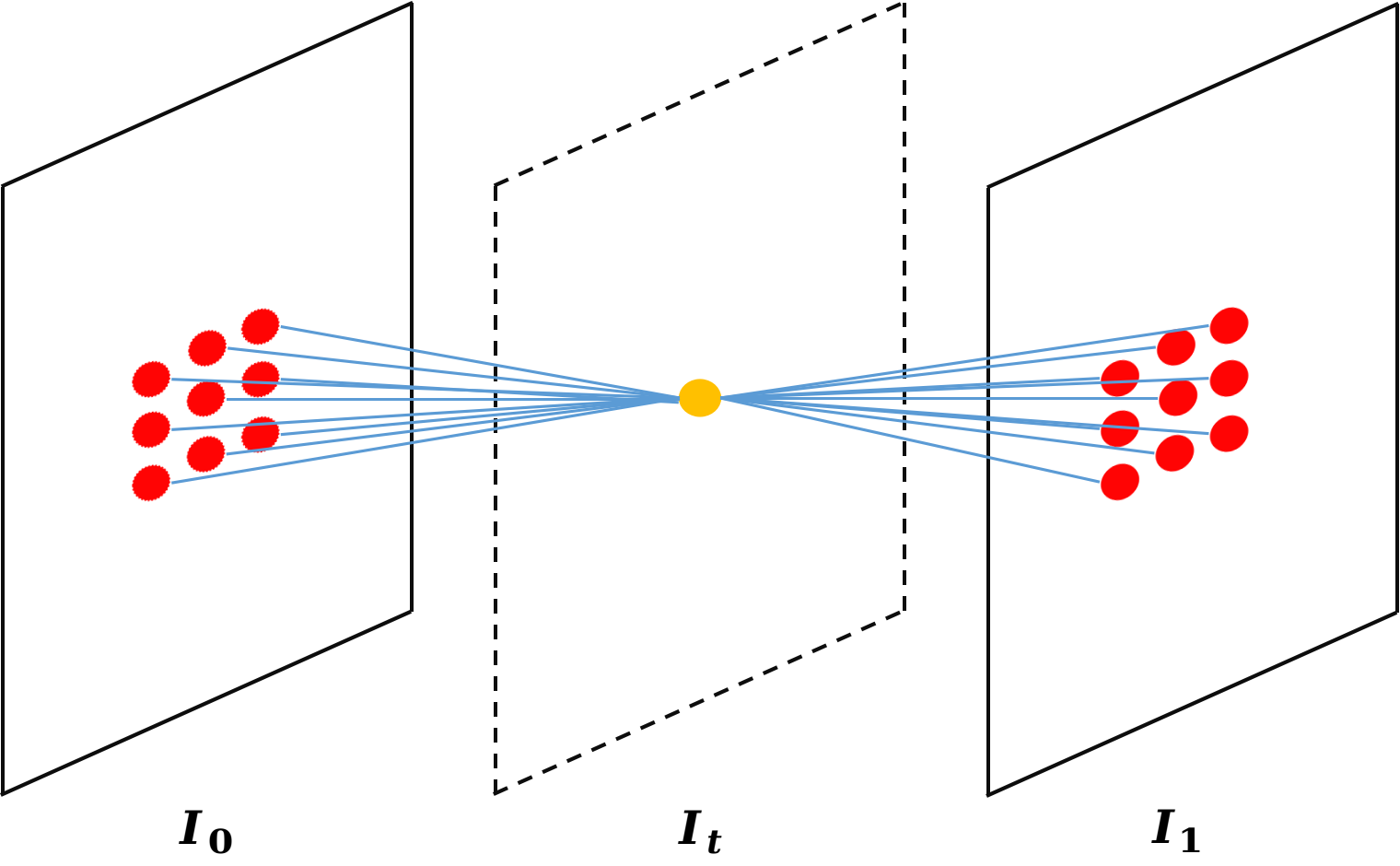}
		\scriptsize{(a)}
	\end{minipage}
	\hspace{2mm}
	\begin{minipage}[b]{0.46\linewidth}
		\centering
		\includegraphics[width=\linewidth]{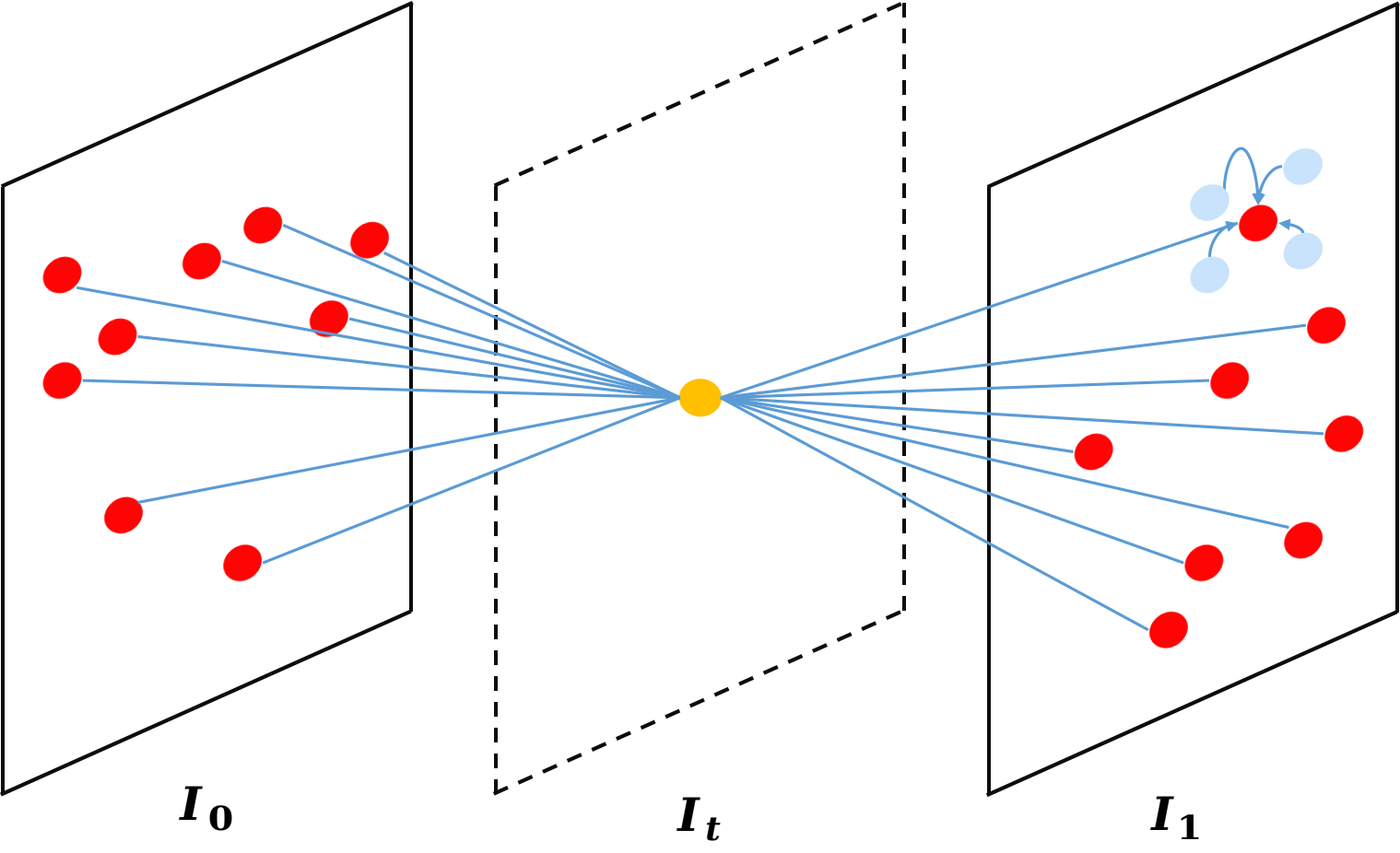}
		\scriptsize{(b)}
	\end{minipage}
	\\
	\vspace{2mm}
	\begin{minipage}[b]{\linewidth}
		\centering
		\includegraphics[width=\linewidth]{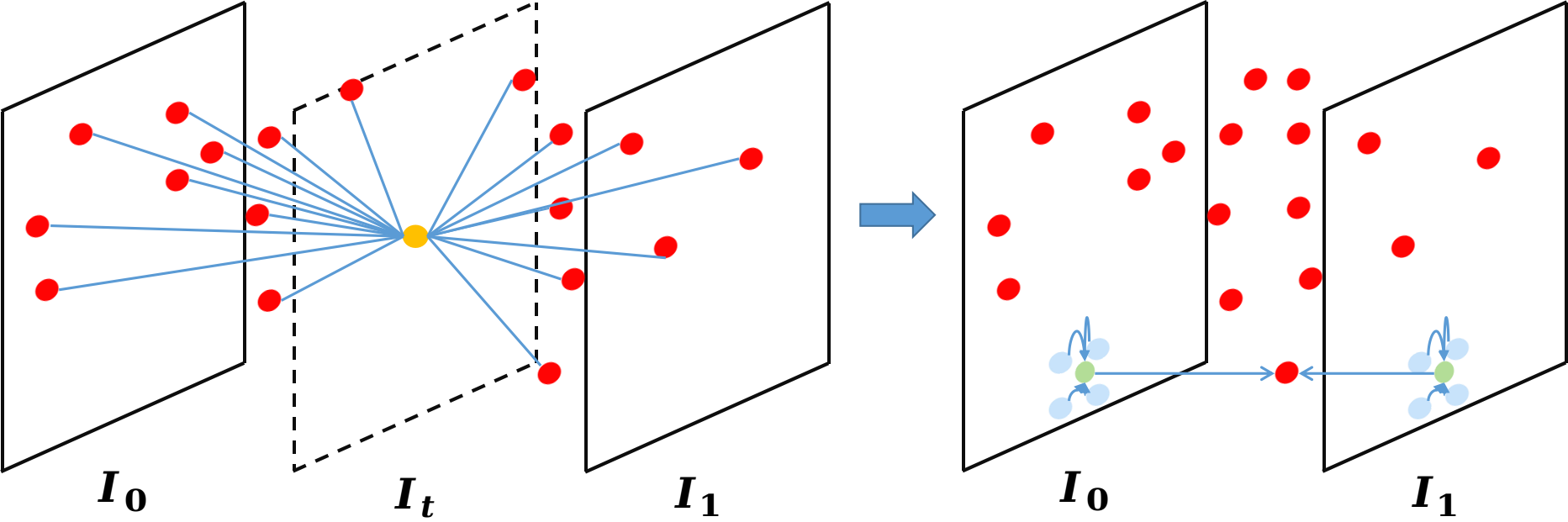}
		\scriptsize{(c)}
	\end{minipage}
	\\
	\vspace{2mm}
	\begin{minipage}[b]{\linewidth}
		\centering
		\includegraphics[width=\linewidth]{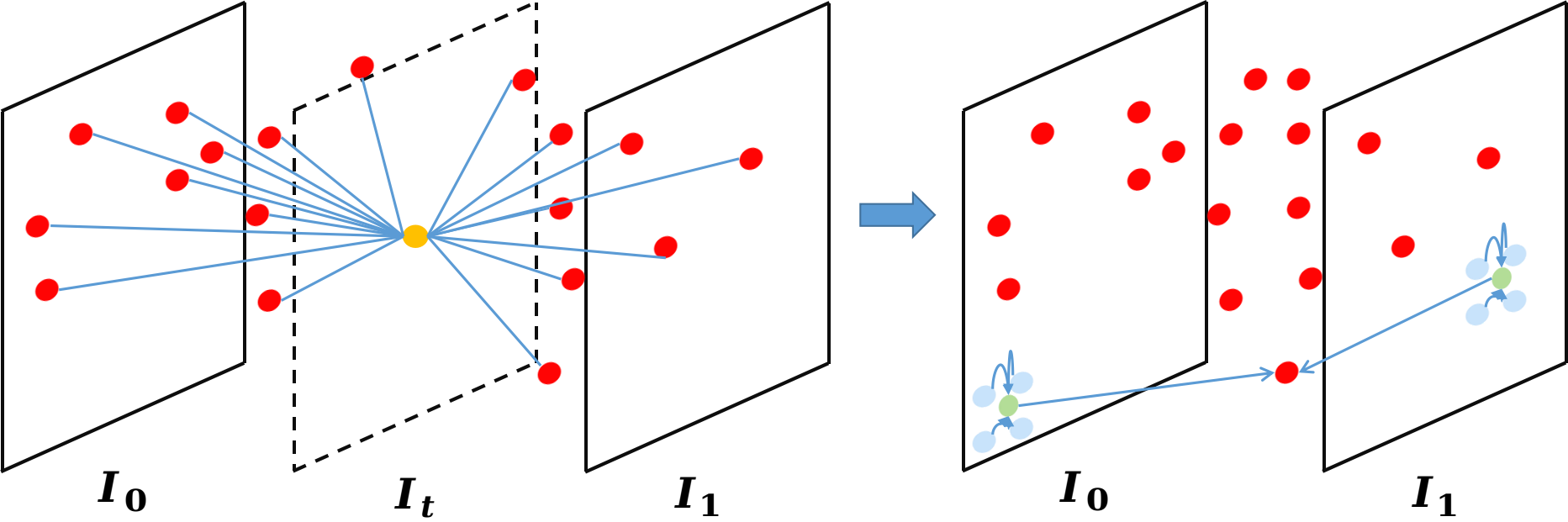}
		\scriptsize{(d)}
	\end{minipage}
	\figvspace
	\caption{Illustration of (a) conventional convolution, (b) AdaCoF, (c)  basic GDConv, (d) advanced GDConv with $T=1$. Here target pixels, sampling points, support points, and neighboring grid points are denoted by yellow, red, green, and blue dots, respectively. For AdaCoF, the value of each sampling point is specified via bilinear interpolation of its four neighboring grid points. For basic GDConv, the value of each sampling point is determined by its two support points via linear interpolation, or equivalently, by its eight associated grid points via trilinear interpolation.
	Advanced GDConv further removes the constraint that the support points need to be spatially aligned with the corresponding sampling point and allows more general numerical interpolation methods.}
	\label{fig:convolution}
\end{figure}

\begin{figure*}[t]
	\centering
	\begin{minipage}[h]{1\linewidth}
		\centering
		\includegraphics[width=\linewidth]{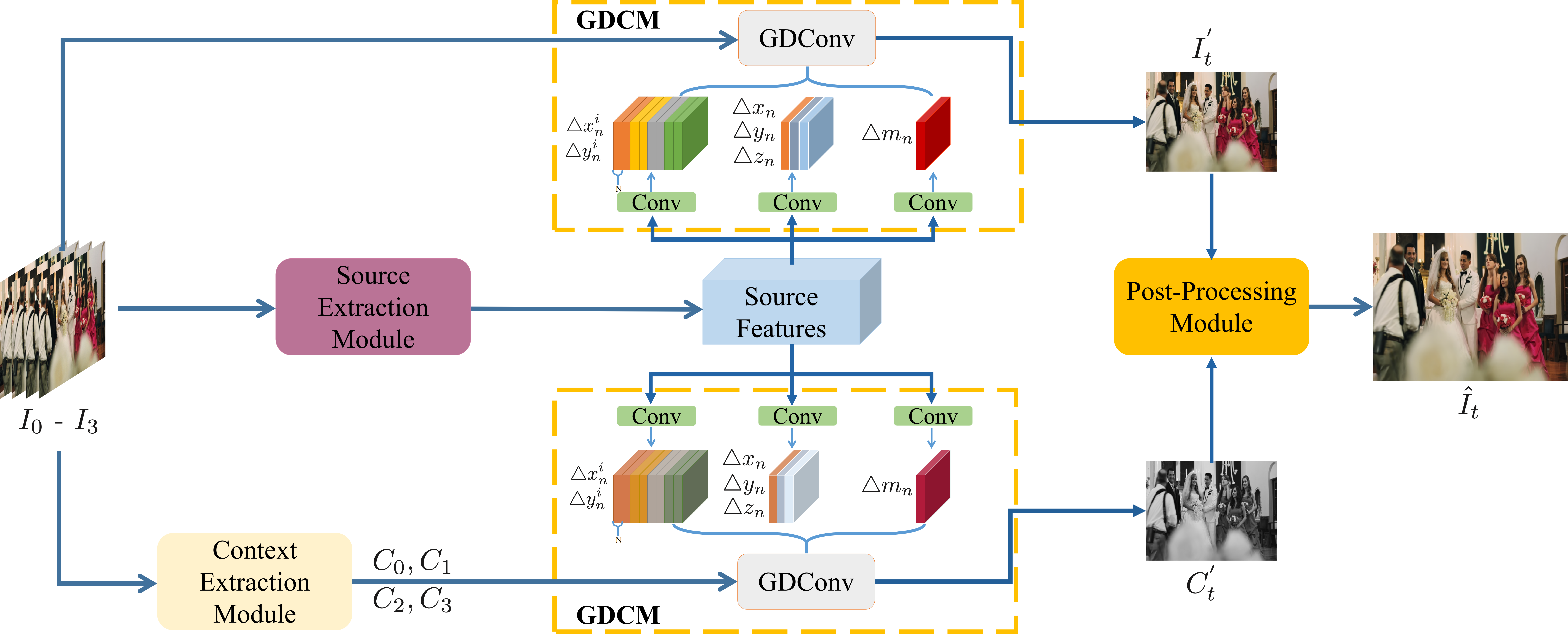}
	\end{minipage}
	\caption{Illustration of the architecture of GDConvNet with $T=3$. Here  $I_0, \cdots, I_3$ are input frames and $C_0, \cdots, C_3$ are their respective context maps; $\triangle x_n^i$ and $\triangle y_n^i$ are spatial offsets (horizontal and vertical) for support point; $\triangle x_n$, $\triangle y_n$ and $\triangle z_n$ are spatial offsets and temporal parameters for sampling points; $\triangle m_n$ is the modulation terms; ${I}'_t$ is a tentative prediction of the target frame $I_t$ while ${C}'_t$ denotes the predicted context map of $I_t$; $\hat{I}_{t}$ is the final output.}
	
	\label{fig:architecture}
\end{figure*}

The input to the GDCM consists of the $T+1$ source frames $I_{0}$, $I_{1}$, $\cdots$, $I_T$ (or the context maps $C_{0}$, $C_{1}$, $\cdots$, $C_T$) and the source features. As shown in Fig.~\ref{fig:architecture}, three different kinds of feature maps, which represent three different types of adaptive parameters, are generated through three different convolution layers, respectively. They are then fed to GDConv along with the  source frames $I_{0}$, $I_{1}$, $\cdots$, $I_T$ (or the context maps $C_{0}$, $C_{1}$, $\cdots$, $C_T$) to synthesize ${I}'_{t}$ (or ${C}'_{t}$). Since the two GDCMs are almost identical, here we only describe the upper one in detail. Moreover, as the operations on the three color channels are the same, we simply regard $I_{i}$ as a single-channel image. For ease of exposition, we first give a brief review of VFI techniques based on conventional convolution \cite{niklaus2017video} and AdaCoF \cite{lee2020adacof}, and then outline the improvements offered by the proposed GDConv.

Conventional convolution is employed in \cite{niklaus2017video} for VFI. This can be formulated as:
\begin{equation}
{I}'_{t} (x,y)= \sum_{i=0}^{T}\sum_{m=1}^{M}W^i_m(x, y) \cdot I_i(x+x_m, y+y_m),
\label{eq:cvconv}	
\end{equation}
where $W^i_m(x,y)$ is a spatially-adaptive convolution weight, and $\{(x_m, y_m)\}_{m=1}^N$ is a collection of pre-defined convolution sampling offsets. Fig.~\ref{fig:convolution}(a) provides an illustration for the special case with $T=1$, $M=9$ and $\{(x_m, y_m)\}_{m=1}^M = \{(-1,-1), (-1,0), \cdots, (1,1)\}$. 
Ideally, the object (pixel) movement should be confined within the coverage of the convolution kernel. As such, in the presence of large motions, this approach is memory-inefficient due to the need for a large number of sampling points to ensure sufficient coverage.

The inefficiency of conventional convolution is largely a consequence of the pre-defined kernel shape (typically, a rectangular grid). AdaCoF \cite{lee2020adacof} addresses this issue by adopting spatially-adaptive deformable convolution, resulting in the following formulation:
\begin{equation}
{I}'_{t} (x,y)= \sum_{i=0}^{T}\sum_{m=1}^{M}W^i_{m}(x, y) \cdot I_i(x+\triangle \alpha_m^i, y+\triangle \beta_m^i),
\label{eq:adacof}
\end{equation}
where $\{(\triangle \alpha_m^i, \triangle \beta_m^i)\}_{m=1}^M$ is a collection of adaptive sampling offsets. In the case where $\triangle \alpha_m^i$ and $\triangle \beta_m^i$ are not integers, $I_i(x+\triangle \alpha_m^i, y+\triangle \beta_m^i)$ is specified through bilinear interpolation. As a result of the introduction of {adaptive sampling offsets, the kernel shape becomes adjustable, as shown in Fig.~\ref{fig:convolution}(b). For this reason, AdaCoF is able to cope with large motions using a relatively small number of sampling points. On the other hand, AdaCoF only exploits the degrees of freedom in the spatial domain. As a result, the sampling points are evenly split among the input frames.
However, this is clearly suboptimal since the frames that are closer to the target intermediate frame in the temporal domain are more relevant and consequently should be allocated with more sampling points.

\begin{figure}[t]
	\centering
	\begin{minipage}[h]{0.8\linewidth}
		\centering
		\includegraphics[width=0.8\linewidth]{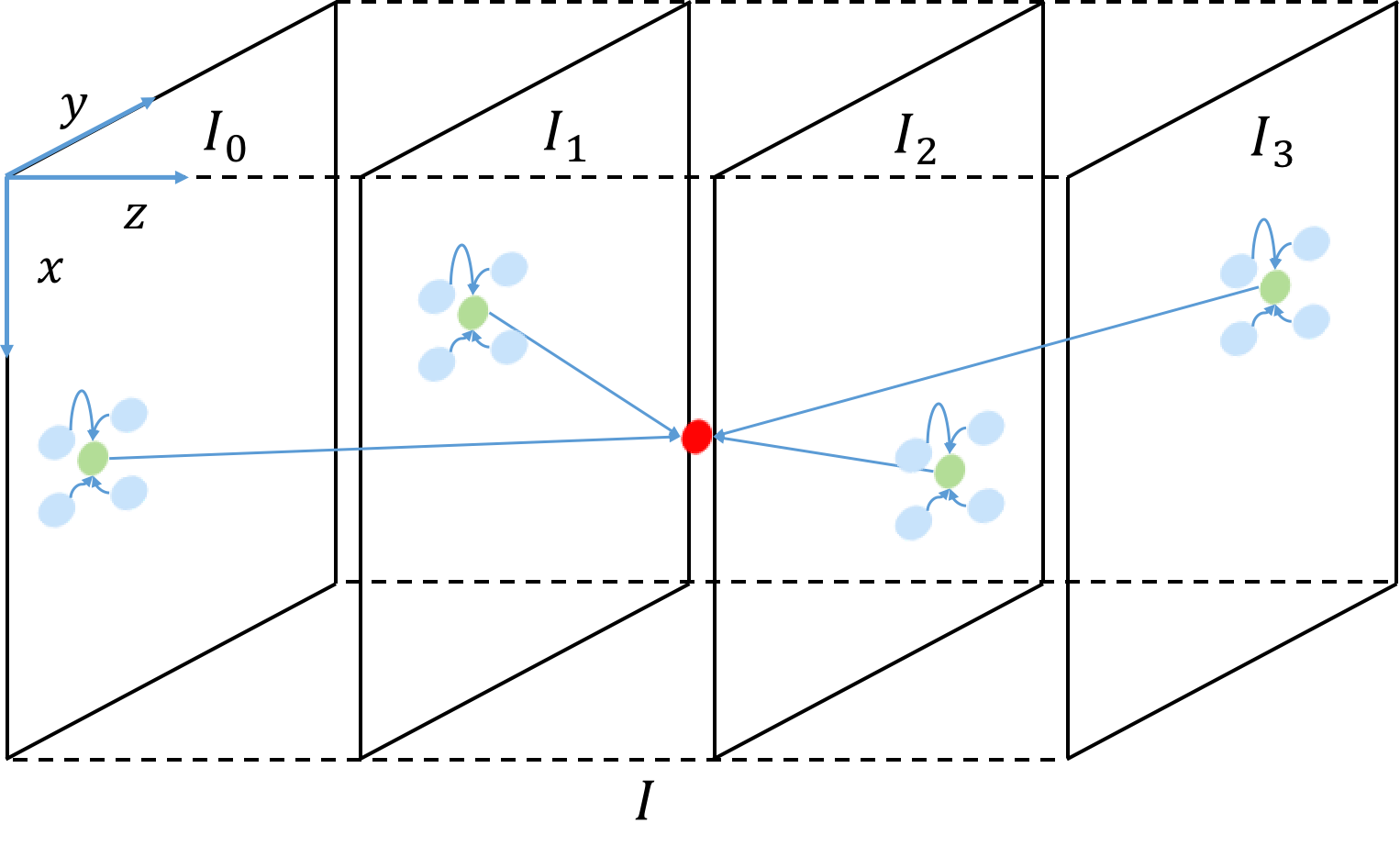}
	\end{minipage}
	\caption{Construction of function $I$ for the special case $T=3$ with a sampling point $(x+\triangle x_n, y+\triangle y_n, z_n)$, its associated support points $(x+\triangle x_n^i, y+\triangle y_n^i, i), i \in \{0, 1, 2, 3\}$, and their neighboring grid points  highlighted in red, green, and blue, respectively.}	
	\label{fig:I_space}
\end{figure}

We shall develop a mechanism that enables flexible allocation of the sampling points across the input frames. In fact, we go one step further by allowing the sampling points to be freely distributed in whole space-time. The key idea is to associate each sampling point with an adaptive temporal parameter $z_n \in [0,T]$, leading to the following formulation:
\begin{equation}
{I}'_{t} (x,y)= \sum_{n=1}^{N}W_{n}(x, y) \cdot I(x+\triangle x_n, y+\triangle y_n, z_n).\label{eq:gdc}
\end{equation}
Here, $I$ is a function (defined on a 3D space) obtained via a judicious extension of  
$I_0$, $I_1$, $\cdots$, $I_T$ to be detailed below (see Fig.~\ref{fig:I_space} for an illustration of the special case in which $T=3$). Note that $z_n$ is allowed to be any real number in $[0,T]$  to facilitate  end-to-end training.
If $z_n$ is an integer, we set
$I(x+\triangle x_n, y+\triangle y_n, z_n)=I_{z_n}(x+\triangle x_n, y+\triangle y_n)$. (Following \cite{lee2020adacof, dai2017deformable, zhu2019deformable}, in the case where $\triangle x_n$ and $\triangle y_n$ are not integers, $I_{z_n}(x+\triangle x_n, y+\triangle y_n)$ is specified via bilinear interpolation of four neighboring grid points.) 
It can be seen that Eq.~(\ref{eq:gdc}) reduces to Eq.~(\ref{eq:adacof}) when $N=(T+1)M$ and each value in $\{0,1,\cdots, T\}$ is taken by the same number of $z_n$. Now it remains to deal with non-integer valued $z_n$, which occurs when the associated sampling point is not exactly located on an input frame.
One simple solution is to set $I(x+\triangle x_n, y+\triangle y_n, z_n)$ as  $(\ceil{z_n}-z_n) \cdot (I(x+\triangle x_n, y+\triangle y_n, \lfloor{z_n}\rfloor)+(z_n-\lfloor{z_n}\rfloor) \cdot I(x+\triangle x_n, y+\triangle y_n, \ceil{z_n})$. (See Fig.~\ref{fig:convolution}(c) for an illustration of the special case in which $T=1$.)
More generally, we attach a set of support points $(x+\triangle x_n^i, y+\triangle y_n^i, i)$, $i \in \{0, 1, \cdots, T\}$, to each sampling point $(x+\triangle x_n, y+\triangle y_n, z_n)$, and use their values $I(x+\triangle x_n^i, y+\triangle y_n^i, i)$ (denoted as $s^i_n$ for short), $i \in \{0, 1, \cdots, T\}$, and their relative positions, to specify $I(x+\triangle x_n, y+\triangle y_n, z_n)$ (denoted as $s_n$ for short) via a numerical interpolation function $G$: 
 \begin{equation}
 \begin{aligned}
 s_n = G(\triangle x_n, \triangle y_n, z_n, \{s_n^i, \triangle x_n^i, \triangle y_n^i\}_{i=0}^T).
 \label{eq:numerical interpolation}
 \end{aligned}
 \end{equation}
Illustrations of special cases with $T=1$ and $T=3$ can be found in  Fig.~\ref{fig:convolution}(d) and Fig.~\ref{fig:I_space}, respectively. Note that each support point has its own adaptive spatial offset $(\triangle x_n^i, \triangle y_n^i)$, which is not necessarily the same as $(\triangle x_n, \triangle y_n)$. Moreover,
there is considerable freedom in the choice of $G$ as long as the differentiability condition needed for end-to-end training is satisfied. We will discuss several candidate numerical interpolation methods in Section \ref{sec:interpolation}. Finally, 
inspired by modulated deformable convolution  \cite{zhu2019deformable}, we rewrite Eq.~(\ref{eq:gdc}) in the following equivalent form:
\begin{equation}
\begin{aligned}
{I}'_{t}(x,y)=\sum_{n=1}^{N}W_{n} \cdot I(x+\triangle x_n, y+\triangle y_n, z_n)\cdot \triangle m_n(x,y),
\label{eq:GDConv}
\end{aligned}
\end{equation}
where $\triangle m_n(x,y)\in [0,1]$ is an adaptive modulation term.

As illustrated in Fig.~\ref{fig:architecture}, three types of feature maps are generated in GDCM via three different convolution layers. The first $2(T + 1)N$ feature maps represent the spatial offsets
(horizontal and vertical) for the support points (i.e., $\triangle x_n^i$, $\triangle y_n^i$), and the next $3N$
feature maps represent the spatial offsets and temporal parameters
for the sampling points (i.e., $\triangle x_n$, $\triangle y_n$, $z_n$), and the last $N$ feature maps represent the modulation terms (i.e., $\triangle m_n$).
We set the initial values of the adaptive parameters $\triangle x_n$, $\triangle y_n$, $z_n$, $\triangle m_n$, $\triangle x_n^i$ and $\triangle y_n^i$  as $0$, $0$, $0$, $1$, $0$, and $0$, respectively.

\subsection{Other Modules}
Now we proceed to give a brief description of the remaining modules in the proposed GDConvNet. 

\begin{figure}[t]
	\centering
	\begin{minipage}[b]{\linewidth}
		\centering
		\includegraphics[width=\linewidth]{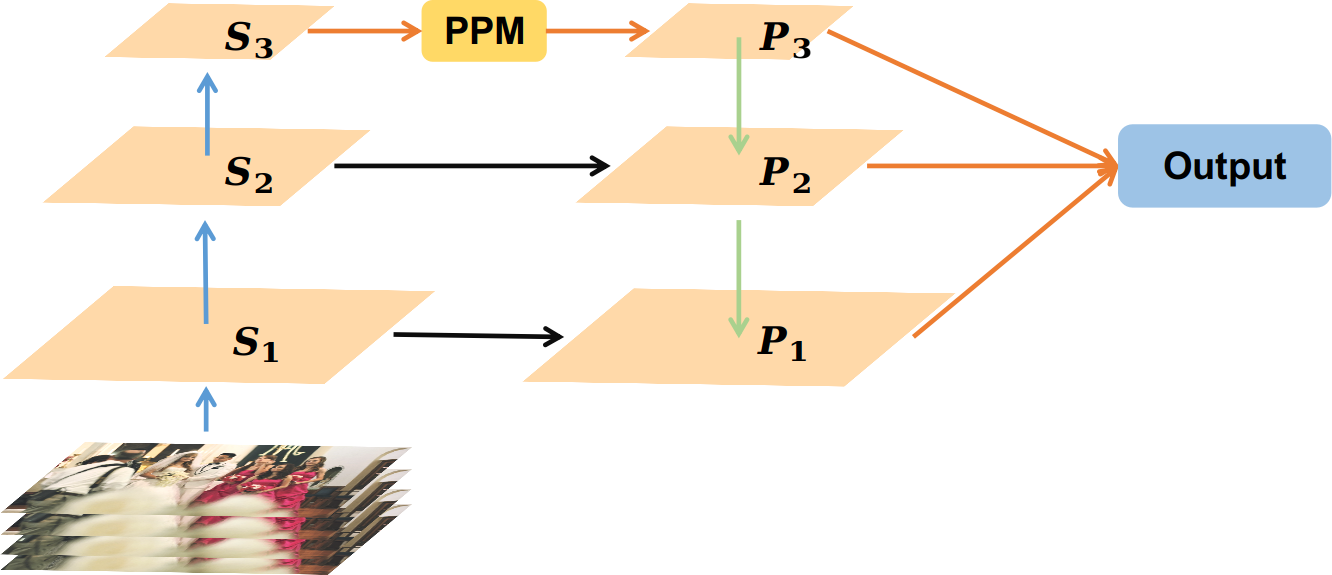}
	\end{minipage}
	\caption{Illustration of the architecture of SEM.}
	\label{fig:SEM}
\end{figure}

\noindent \textbf{Source Extraction Module:} As shown in Fig.~\ref{fig:SEM}, we adopt the FPN backbone \cite{lin2017feature} to generate hierarchical features. In the bottom-up pathway, there are three levels (each consisting of two residual blocks and one convolution layer) and the associated feature maps (which are of different scales) are denoted as $S_1$, $S_2$, and $S_3$. The input $P_3$ to the top level of the top-down pathway is generated from $S_3$ through a pyramid pooling module \cite{zhao2017pyramid}. $P_3$ is then upsampled and merged with $S_2$ via element-wise addition to generate $P_2$, which is further upsampled and merged with $S_1$ to generate $P_1$. Finally, $P_2$ and $P_3$ are upsampled and concatenated with $P_1$ to form the output. 

\noindent \textbf{Context Extraction Module:}
It is demonstrated in \cite{niklaus2018context} that context information is very important for VFI. We use one convolution layer and two residual blocks \cite{he2016deep} to sequentially extract contextual features. A SEblock \cite{hu2018squeeze} is then used to rearrange these feature maps, and finally its output is smoothed by a convolution layer.

\noindent \textbf{Post-Processing Module:} To refine the warped image, we adopt the GridDehazeNet architecture \cite{liu2019griddehazenet},  where each row is associated with a different scale and contains five RDB blocks \cite{zhang2018residual}, while each column can be considered as a bridge connecting different scales through downsampling or upsampling modules. (which decrease or increase the size of feature maps by a factor of two.) Instead of employing the hard attention mechanism in \cite{liu2019griddehazenet}, we use  SEBlocks \cite{hu2018squeeze} to adaptively rebalance the incoming information flows at the junctions of GridDehazeNet.

\section{Understanding Generalized Deformable Convolution in VFI}

In this section, we shall place generalized deformable convolution in a board context and explain why it is an effective mechanism for VFI.

\subsection{Related Works}
Generalized deformable convolution is conceptually related to several existing ideas in the literature.

\noindent \textbf{Deformable Convolution:} There are many works on variants of conventional convolution with improved performance,
including active convolution \cite{jeon2017active}, dynamic filter \cite{jia2016dynamic}, atrous convolution \cite{holschneider1990real}, among others. A culminating achievement of this line of research is deformable convolution \cite{dai2017deformable,zhu2019deformable}. 
Our generalized deformable convolution degenerates to conventional deformable convolution \cite{dai2017deformable,zhu2019deformable} if the temporal dimension is not present, and its basic form, shown in Fig.~\ref{fig:convolution}(c), can be viewed as a $3D$-version of deformable convolution.


\noindent \textbf{Non-Local Network:} In deep learning, non-locality means that the receptive field is not restricted to a certain local region and can capture long-range context information. The receptive field of conventional convolution is typically a fixed grid and consequently is local in nature. Significant efforts have been devoted to addressing this issue \cite{zhao2018psanet,zhao2017pyramid,holschneider1990real,wang2018non}. Arguably the most successful one is \cite{wang2018non}, which takes all possible spatial positions into consideration. However, this comes at the cost of high memory usage. In contrast, generalized deformable convolution is memory-efficient as it is able to achieve non-local coverage and capture long-range context information with a relatively small kernel by adaptively and intelligently selecting sampling points in space-time.

\noindent \textbf{Attention Mechanism:} An attention mechanism enables differentiated treatment of different input features according to their relative importance, which has shown to yield significant performance gain in many vision tasks. Traditionally, it can be divided into spatial-wise attention \cite{wang2019spatial} and channel-wise/temporal-wise attention \cite{hu2018squeeze}. Recently, there have also been attempts \cite{park2018bam,woo2018cbam} to combine these two types of attention. Nevertheless, in these approaches the spatial-wise and channel-wise/temporal-wise attention maps are still generated separately. It is interesting to note that generalized deformable convolution offers a natural way to consolidate these two types of attention by suitably modulating the sampling points at different locations in space-time.

\noindent \textbf{Non-Linearity:} The conventional approach to increasing the non-linearity of convolutional neural networks~\cite{lecun1998gradient,simonyan2014very,krizhevsky2012imagenet}  is 
by stacking more non-linear modules~\cite{krizhevsky2012imagenet,he2015delving}. However, it has been recognized that a more effective approach is to allow the functionalities of  constituent modules to be input-dependent~\cite{hu2018squeeze,jaderberg2015spatial,wang2019spatial}.
From this perspective, generalized deformable convolution converts a linear convolution operation into a highly non-linear operation by adaptively adjusting its kernel according to the input, and by doing so it yields enhanced learning capabilities.

\subsection{Comparison with State-of-the-Art VFI Algorithms}
The state-of-the-art VFI methods can be divided into two categories: flow-based methods and kernel-based methods. For illustrative purposes, we shall consider the simple scenario where two source frames $I_1$ and $I_2$ are used to predict one target frame $I_{1.5}$, unless specified otherwise.

\begin{figure}[t]
	\center
	\includegraphics[width=\linewidth]{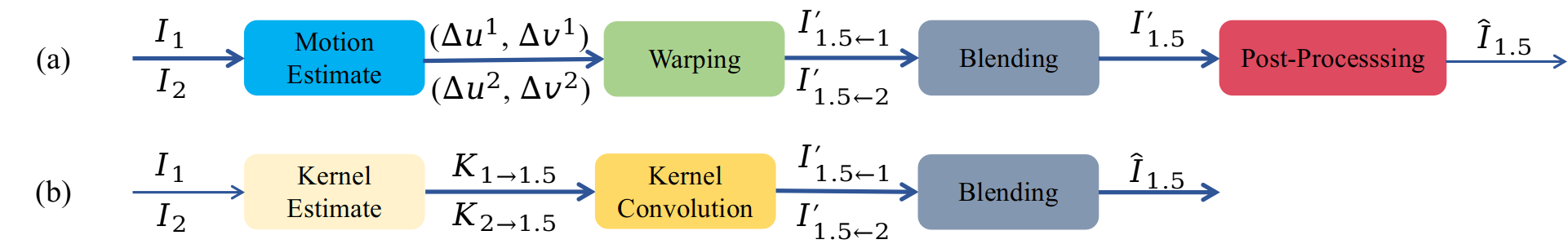}
	\caption{Illustration of (a) flow-based VFI pipeline and (b) kernel-based VFI pipeline.}
	\label{fk pipeline}
\end{figure}

\noindent \textbf{Flow-based:} These methods admit a common mathematical formulation as follows:

\begin{equation}
I_{1.5\leftarrow 1}^{'}(x,y) = I_1(x + \triangle u^1, y + \triangle v^1),
\label{eq:flow 1}
\end{equation}
\begin{center}
	or
\end{center}
\begin{equation}
I_{1.5\leftarrow 2}^{'}(x,y) = I_2(x + \triangle u^2, y + \triangle v^2),
\label{eq:flow 2}
\end{equation}
where $(\triangle u^1, \triangle v^1)$ and $(\triangle u^2, \triangle v^2)$ are respectively optical flow fields from $I_{1.5}$ to $I_1$ and $I_2$, while $I_{1.5\leftarrow 1}^{'}$ and  $I_{1.5\leftarrow 2}^{'}$ denote the warped images from each direction. The pipeline of flow-based methods is illustrated in Fig.~\ref{fk pipeline}(a). First, two input frames are used to estimate optical flow maps, typically with the help of traditional optical flow estimation methods \cite{zhang2019refined, horn1981determining, lucas1981iterative} or convolution neural network \cite{ranjan2017optical,dosovitskiy2015flownet,ilg2017flownet,sun2018pwc}. The input frames are then warped according to these optical flow maps. Finally, blending and post-processing operations are performed to generate the final output. The linear motion model is widely adopted in flow map estimation. However, this model is not accurate for describing accelerated and curvilinear motions. To handle such complex motions, a quadratic model is proposed in \cite{qvi_nips19}, where $(\triangle u^1, \triangle v^1)$ and $(\triangle u^2, \triangle v^2)$ are estimated based on four frames $I_0$, $I_1$, $I_2$, and $I_3$ instead of just $I_1$ and $I_2$. To understand the connection with our method, it is instructive to consider a special case of Eq.~(\ref{eq:GDConv}) with $N=1$, where $z_1=1$, $(\triangle x_1, \triangle y_1)=(\triangle x_1^1, \triangle y_1^1)$, or $z_1=2$, $(\triangle x_1, \triangle y_1)=(\triangle x_1^2, \triangle y_1^2)$:
\begin{equation}
\begin{aligned}
I_{1.5 \leftarrow 1}^{'}(x,y)&=W_{1} \cdot I(x+\triangle x_1, y+\triangle y_1, 1)\cdot \triangle m_1\\
&=W_{1} \cdot I_1(x+\triangle x_1^1, y+\triangle y_1^1)\cdot \triangle m_1,
\label{eq:ours flow 1}
\end{aligned}
\end{equation}
\begin{center}
	or
\end{center}
\begin{equation}
\begin{aligned}
I_{1.5 \leftarrow 2}^{'}(x,y)
&=W_{1} \cdot I(x+\triangle x_1, y+\triangle y_1, 2)\cdot \triangle m_1\\
&=W_{1} \cdot I_2(x+\triangle x_1^2, y+\triangle y_1^2)\cdot \triangle m_1.
\label{eq:ours flow 2}
\end{aligned}
\end{equation}
One can readily recover Eq.~(\ref{eq:flow 1}) and  Eq.~(\ref{eq:flow 2}) from Eq.~(\ref{eq:ours flow 1}) and Eq.~(\ref{eq:ours flow 2}) by setting $W_{1}=\triangle m_1=1$ and interpreting $(\triangle x^i_1, \triangle y^i_1)$ as $(\triangle u^i, \triangle v^i)$, $i=1,2$. Similarly to the case with $(\triangle u^1, \triangle v^1)$ and $(\triangle u^2, \triangle v^2)$  in
\cite{qvi_nips19},  the estimation of the offsets $(\triangle x^1_1, \triangle y^1_1)$ and $(\triangle x^2_1, \triangle y^2_1)$ can also benefit from more than two source frames. 
More importantly, in our method, the offset estimation does not directly resort to any predetermined mathematical model and is carried out in a completely data-driven manner.
As such, it can cope with real-world motions more flexibly and accurately. Furthermore, for the general version of our method, the number of sampling points can be set to be greater than 1 (i.e., $N>1$), which, together with the freedom in choosing the space-time coordinates of the sampling points and the relaxation of the constraint $(\triangle x_n, \triangle y_n)=(\triangle x_n^i, \triangle y_n^i)$, makes it possible to capture complex diffusion and dispersion effects. Finally, we would like to point out that the space-time numerical interpolation operation in our method plays a role similar to that of the blending operation in some existing flow-based methods \cite{bao2019memc,jiang2018super,xue2019video} (see also Fig.~\ref{fk pipeline}(a)), but requires fewer parameters, as it is performed at the sampling point level.

\noindent \textbf{Kernel-based:}  These methods \cite{niklaus2017video,niklaus2018video, lecun1998gradient} generate two sets of spatially-adaptive convolution kernels and use them to convolve with source frame patches to get the predicted target frames $I_{1.5\leftarrow 1}^{'}$, $I_{1.5\leftarrow 2}^{'}$ from two sides, which are then blended at the pixel level to get final VFI result:
\begin{equation}
\begin{aligned}
\hat{I}_{1.5}(x,y)&=I_{1.5\leftarrow 1}^{'}(x,y) + I_{1.5\leftarrow 2}^{'}(x,y)\\
&=K_1(x,y)*I_1(x,y) + K_2(x,y)*I_2(x,y).
\end{aligned}
\end{equation}
The pipeline of kernel-based methods is shown in Fig.~\ref{fk pipeline}(b). Note that in the presence of complex motions, the technique in \cite{niklaus2017video} and \cite{niklaus2018video} need to adopt large kernels (specifically, the size of convolutional kernels used in \cite{niklaus2017video} and \cite{niklaus2018video} are $41 \times 41$ and $51 \times 51$, respectively) to ensure sufficient coverage, which is inflexible and memory-inefficient. AdaCoF \cite{lee2020adacof} addresses this issue by adopting  deformable convolution. Nevertheless, the sampling points in AdaCoF are only spatially adaptive. In contrast, the proposed method can make more effective use of the sampling points by freely exploring space-time (not just in the spatial domain). As such, it often suffices to employ small kernels, even when dealing with very complex motions. Our method also has the additional advantage of blending images 
at the sampling point level (in the form of space-time numerical interpolation), which is more efficient than blending at the pixel level in kernel-based methods.

\section{Four-Frame VFI Experiments}
Due to its flexibility, our method can leverage an arbitrary number of frames for VFI. Here we focus on the four-frame VFI case. 
The experimental results for two-frame VFI will be presented in Section \ref{Sec:2-frame}.

\subsection{Implementation Details}
We use four source frames $I_0$, $I_1$, $I_2$, and $I_3$ to synthesize the target frame $I_{1.5}$. In GDConv, the number of sampling points for each warped pixel is set to $25$. The loss function, the training dataset, and the training strategy are described below.

\noindent \textbf{Loss Function:}
In addition to the supervision provided at the output end, we introduce intermediate supervision to ensure proper training of the GDCM (which is the key component of GDConvNet). Note that without intermediate supervision, we have no direct control of the training of the GDCM  due to the fact that the downstream post-processing module, which is a relatively large and complex network, tends to dilute the impact of the supervisory signal.
The overall loss function can be formulated as:
\begin{equation}
\begin{aligned}
\mathcal{L} &= \mathcal{L}_{r} + \lambda \mathcal{L}_{w}\\
&=\sum_{x}||\hat{I}_t(x) - I_{GT}(x)||_1 + \lambda \sum_{x} ||I_t^{'}(x)  - I_{GT}(x)||_1,
\end{aligned}
\end{equation}
where $I_{GT}$ is the ground-truth frame, and $\lambda$ is a hyper-parameter to balance the warped loss $\mathcal{L}_{w}$ and the refined loss $\mathcal{L}_{r}$. (Experimentally, we found that $\lambda=0.5$ yields the best performance.) We use the $\ell_1$ norm instead of the $\ell_2$ norm because the latter is known to produce blurry results in image synthesis tasks. Following \cite{liu2017video, lee2020adacof, bao2019memc, bao2019depth}, we use the Charbonnier Function $\Phi(x) = \sqrt{x^2 + \epsilon ^2}$ to smoothly approximate the $\ell_1$ norm and set $\epsilon=10^{-6}$.

\noindent \textbf{Training DataSet:} 
The Vimeo90k Septuplet training dataset \cite{xue2019video} is used to train our model. This training dataset is composed of $64,612$ seven-frame sequences with a resolution of $256 \times 448$. 
We use the first, the third, the fifth, and the seventh frames (corresponding to $I_0$, $I_1$, $I_2$, and $I_3$ in our notation, respectively) of each sequence to predict the fourth one (corresponding to $I_{1.5}$). We randomly crop image patches of size $256 \times 256$ for training. Horizontal and vertical flipping, as well as temporal order reversal, are performed for data augmentation.

\noindent \textbf{Training Strategy:} 
Different from \cite{bao2019depth,bao2019memc,xue2019video}, our network can be trained from scratch without relying on any pre-trained model. We adopt the Adam optimizer \cite{kingma2014adam}, where $\beta_1$ and $\beta_2$ are set as the default values $0.9$ and $0.999$, respectively. We set the training batch size as $8$ and train our network for $14$ epochs (nearly $11,300$ iterations) in total. The initial learning rate is set as $10^{-3}$, and the learning rate is reduced by a factor of two every $4$ epochs for the first $8$ epochs and by a factor of five every $2$ epochs for the last $6$ epochs. The training is carried out on four NVIDIA GTX 1080Ti GPUs, and takes about $58$ hours to converge.

\begin{table}[t]
	\vspace{-3mm}
	\begin{tabular}{c|cccc}
		\toprule
		Dataset        & avg disp. & $>15$ & $>20$ & $> 25$\\ \hline
		Vimeo90k Dataset & $6.1$& $9.1\%$ & $5.0\%$ & $3.0\%$\\
		Gopro Dataset & $6.1$& $7.0\%$ &$2.7\%$ & $1.0\%$ \\
		Adobe240 Dataset & $8.2$& $13.0\%$ & $9.0\%$ & $6.1\%$\\
		\bottomrule
	\end{tabular}
	\caption{The statistics of pixel displacement within different datasets. This table shows the average pixel displacement, the percentage of pixels with displacement larger than $15$, $20$ and $25$ respectively for three datasets.}
	\label{tab:pixel move}
\end{table}

\subsection{Evaluation Datasets}
The following three datasets are used for performance evaluation.

\noindent \textbf{Vimeo90K Septuplet Test Set \cite{xue2019video}:} This dataset consists of $7,824$ video sequences, each with $7$ frames. As in the case of the Vimeo90K Septuplet training dataset, the first, the third, the fifth, and the seventh frames of each sequence are leveraged to synthesize the fourth one. The image resolution of this dataset is $256 \times 448$.

\noindent \textbf{Gopro Dataset \cite{nah2017deep}:} 
This dataset is composed of $33$ high-resolution videos recorded by hand-held cameras. The frame rate of each video is $240$ fps, and the image resolution is $720 \times 1,280$. The dataset was released in an image format, consisting of a total of $35,782$ images. We successively group every $25$ consecutive images as a test sequence, and resize the images to $360 \times 480$. Finally, $1,392$ test sequences are selected. For each sequence, the first, the ninth, the seventeenth, and the twenty-fifth frames (corresponding to $I_0$, $I_1$, $I_2$, and $I_3$, respectively) are used to synthesize the thirteenth frame (corresponding to $I_{1.5}$). This dataset is rich with non-linear camera motions and dynamic object motions, posing significant challenges to VFI methods in these respects.

\noindent \textbf{Adobe240 Dataset \cite{su2017deep}:}
This dataset consists of $133$ $240$ fps videos in total, where the resolution of each video is $720 \times 1,280$. These videos are recorded by hand-held cameras, and mainly contain outdoor scenes. Different from the Gopro dataset, this dataset is released in a video format. We extract $7,479$ non-overlapped test sequences, each with $25$ frames. This dataset is rich with large motions. Indeed, it has the largest average pixel displacement among the three datasets under consideration according to Table.~\ref{tab:pixel move}. Therefore, it can be used to examine the strength of a VFI method in handling such motions.
	
\begin{table*}[t]
	\label{tab:numerical interpolation performance}
	\caption{Quantitative comparisons of GDConvNet with different numerical interpolation methods on Viemo-90k test dataset, Gopro dataset, and Adobe240 dataset.}
	\vspace{-2mm}
	\footnotesize
	\center
	\renewcommand\tabcolsep{15.0pt}
	\resizebox{\textwidth}{!}{\begin{tabular}{ccccccccc}
			\toprule
			\multicolumn{1}{c}{\multirow{2}*{Method}} 
			&{\#Parameters} 
			&\multicolumn{2}{c}{Vimeo-90k} 
			&\multicolumn{2}{c}{Gopro}
			&\multicolumn{2}{c}{Adobe240} \\  
			\cmidrule(r){3-4} 
			\cmidrule(r){5-6}
			\cmidrule(r){7-8}
			& (million) & PSNR & SSIM & PSNR & SSIM & PSNR & SSIM\\
			\hline
			Ours-Linear &$5.1$& $34.96$& $0.9534$& $30.06$& $0.9092$& $34.20$&$0.9422$\\
			Ours-3D Inv &$5.1$& $35.01$& $0.9535$& $30.12$& $0.9099$& $34.27$&$0.9427$\\
			Ours-1D Inv &$5.1$& $35.08$& $0.9541$& $30.16$& $0.9099$& $34.36$&$0.9436$\\
			\hline
			Ours-Poly 	&$5.1$ &$\textbf{35.58}$ &$\textbf{0.9580}$ &$\textbf{30.49}$ &$\textbf{0.9180}$ &$\textbf{34.53}$ &$\textbf{0.9456}$\\
			Ours-Poly-clamping 	&$5.1$ &$35.10$ &$0.9548$ &$30.18$ &$0.9072$ &$34.33$ &$0.9442$\\
			\bottomrule
			
	\end{tabular}}
	
\end{table*}

\subsection{Numerical Interpolation Methods} \label{sec:interpolation}
As described in Section \ref{sec:2-A}, a numerical interpolation function $G$ is used to specify the value  $s_n=I(x+\triangle x_n, y+\triangle y_n, z_n)$ of a sampling point
in accordance with its position and the corresponding support points $s_n^i=I(x+\triangle x_n^i, y+\triangle y_n^i, i)$, $i\in \{0, 1, \cdots, T\}$, when it does not exactly lie on an input frame (i.e., when $z_n$ is not an integer). In principle, any numerical interpolation function satisfying the differentiability condition can be leveraged for this purpose. However, different numerical interpolation functions may generate different values for the same sampling point and consequently lead to different final outputs. Therefore, it is important to understand how the choice of the numerical interpolation function affects the overall system performance. To this end, 
we investigate the following representatives: linear interpolation, 3D and 1D versions of inverse-distance-weighted interpolation, and polynomial interpolation.

\subsubsection{Linear Interpolation}
This is one of the simplest interpolation methods. It can be be formulated as:
\begin{equation}
s_n=\sum_{i=0}^{T}max(0, 1-|z_n-i|) \cdot s_n^i.
\label{eq:linear}
\end{equation}
Note that even if $T>1$, only two adjacent support points are taken into consideration in Eq.~(\ref{eq:linear}) for interpolating $s_n$. (The maximum operation suppresses the contribution of other support points.) We regard this interpolation method as the baseline in comparisons.

\begin{table}[h]
	\caption{\label{tab:mean of distance} Mean of the squared distance.}
	\center
	\begin{tabular}{ccc}
		\hline
		$(d_x)^2$ & \hspace{5mm} $(d_y)^2$ &\hspace{5mm} $(d_z)^2$\\    %
		\hline
		$0.0025$& \hspace{6mm} $0.0010$ \hspace{6mm} & $0.2009$\\
		\hline
	\end{tabular}
\end{table}

\subsubsection{3D Version of Inverse-Distance-Weighted Interpolation (3D Inv)}
In contrast to linear interpolation, this  method makes use of all support points (see Fig.~\ref{fig:I_space}) as follows:

\begin{equation}
	s_n=\frac{\sum_{i=0}^{T}w_i \cdot s_n^i}{\sum_{i=0}^{T}w_i},\label{eq:inversedistance}
\end{equation}
where $w_i=1/((d^i_x)^2+(d_y^i)^2+(d_z^i)^2)$,  $d^i_x=|\triangle x_n-\triangle x_n^i|/H$, $d^i_y=|\triangle y_n-\triangle y_n^i|/W$, and $d^i_z=|z_n-i|/T$. The quantitative comparisons in Table \ref{tab:numerical interpolation performance} indicate that leveraging all support points instead of just two adjacent points yields better performance. 
Table \ref{tab:mean of distance} shows the means of $(d^i_x)^2$, $(d^i_y)^2$, and $(d^i_z)^2$ (averaged over $i$),  denoted as $(d_x)^2$, $(d_y)^2$ and $(d_z)^2$, respectively. It is clear that $(d_x)^2$ and $(d_y)^2$ are about two orders of magnitude smaller than $(d_z)^2$. This implies that it might suffice to set the weights based on the temporal information alone, which naturally suggests the following interpolation method.

\subsubsection{1D Version of Inverse-Distance-Weighted Interpolation (1D Inv)}

Setting $w_i=1/(d^i_z)^2$ in Eq.~(\ref{eq:inversedistance}) leads to the 1D version of inverse distance weighted interpolation (see Fig.~\ref{fig:interpolation curve}(a) for an example with $T=3$). The quantitative results of this interpolation method are shown in Table \ref{tab:numerical interpolation performance}. Somewhat surprisingly, the 1D version slightly outperforms its 3D counterpart. The reason is that focusing on the dominant dimension enables more effective use of the training data and consequently yields more accurate VFI results. This suggests that it might be possible to further improve the performance by employing more advanced 1D interpolation methods.

\begin{figure}[h]
	\centering
	\begin{minipage}[b]{0.48\linewidth}
		\centering
		\includegraphics[width=\linewidth]{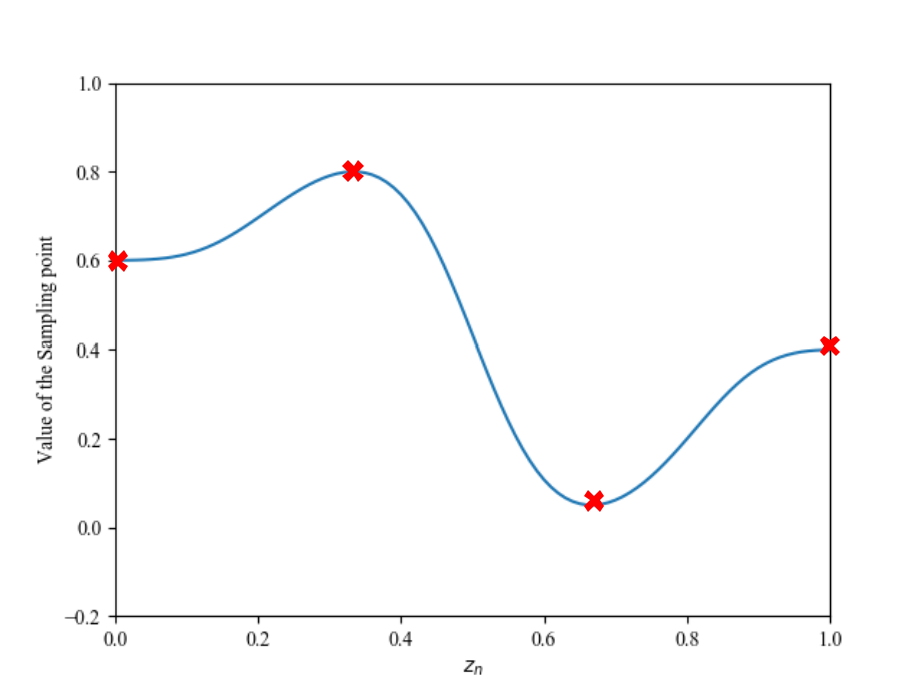}
		\scriptsize{(a)}
	\end{minipage}
	\begin{minipage}[b]{0.485\linewidth}
		\centering
		\includegraphics[width=\linewidth]{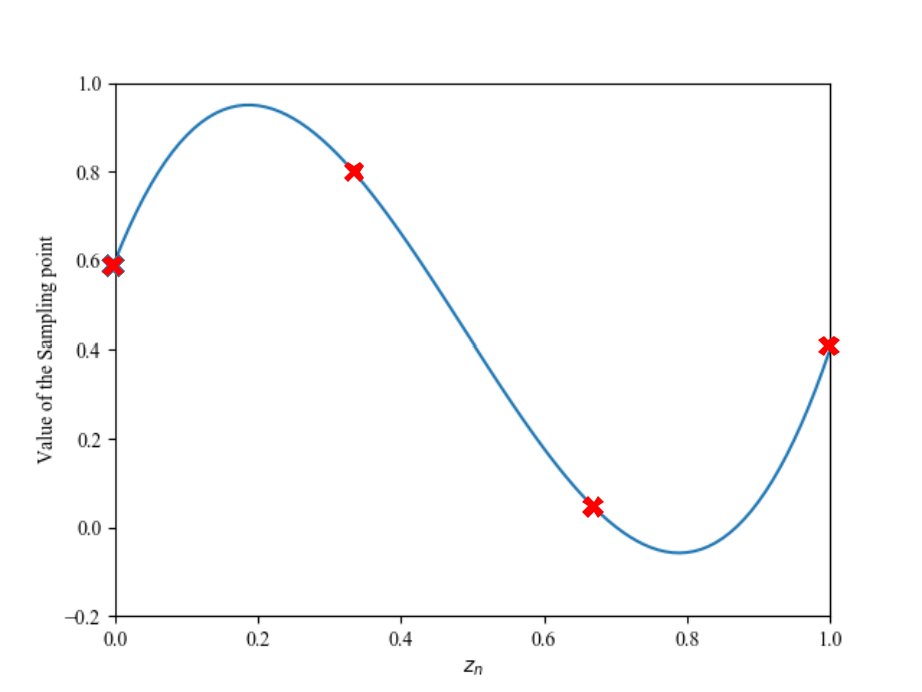}
		\scriptsize{(b)}
	\end{minipage}
	\caption{Illustration of (a) 1D version of inverse distance weighted interpolation and (b) polynomial interpolation with support points highlighted in red. 
	Here $s_n^0=0.6$, $s_n^1=0.8$, $s_n^2=0.05$, and $s_n^3=0.4$, respectively.}
	\label{fig:interpolation curve}
\end{figure}

\subsubsection{Polynomial Interpolation (Poly)}
This method uses a polynomial function of degree $T$ to perform interpolation. More specifically, we have:

\begin{equation}
\begin{aligned}
G= a_0+a_1 z_n + \cdots + a_Tz_n^T, z\in[0, T],
\end{aligned}
\end{equation}
where the coefficients $a_0$, $a_1$, $\cdots$, and $a_T$ can be uniquely determined by jointly solving $T+1$ linear equations  $G|_{z_n=i}=s_n^i$, $i \in \{0, 1, \cdots, T\}$. It should be emphasized that sampling points and their associated support points are still selected in 3D space-time even if a 1D interpolation method is adopted; as such, the overall method is intrinsically 3D.

Fig.~\ref{fig:interpolation curve}(b) provides an example of polynomial interpolation with $T=3$. In contrast to 1D Inv, polynomial interpolation is able to generate values beyond the upper and lower limits of $s_n^i$, $i \in \{0, 1, \cdots, T\}$. This extra freedom might be the reason why polynomial interpolation leads to $0.5$ dB improvement over 1D Inv as shown in Table~\ref{tab:numerical interpolation performance}.

\begin{table}[t]
	\begin{tabular}{c|cc}
		\toprule
		Method        & Beyond Upper & Beyond Lower\\ \midrule
		Upper GDCM & $10.8\%$& $22.6\%$ \\
		Lower GDCM & $9.3\%$& $10.0\%$ \\
		\bottomrule
	\end{tabular}
	\caption{The statistical distribution of sampling points beyond limits.}
	\label{tab:beyond limits}
\end{table}

\begin{figure}[t]
	
	\centering
	\begin{minipage}[b]{0.45\linewidth}
		\centering
		\includegraphics[width=\linewidth]{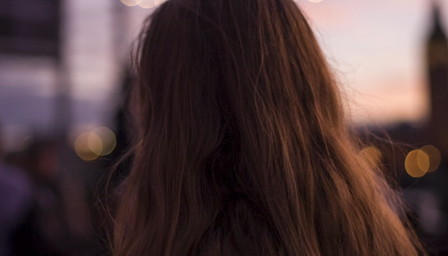}
		\centerline{(a) ground truth}
		\vspace{0.0001cm}	
	\end{minipage}
	\begin{minipage}[b]{0.45\linewidth}
		\centering
		\includegraphics[width=\linewidth]{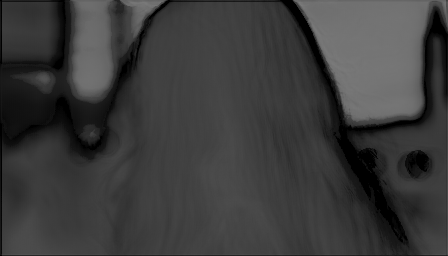}
		\centerline{(b) ER map w/o clamping}
		\vspace{0.0001cm}	
	\end{minipage}
	\\
	\begin{minipage}[b]{0.45\linewidth}
		\centering
		\includegraphics[width=\linewidth]{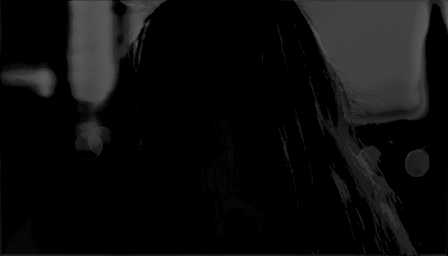}
		\centerline{(c) ER map w clamping}
		\vspace{0.0001cm}	
	\end{minipage}
	\begin{minipage}[b]{0.45\linewidth}
		\centering
		\includegraphics[width=\linewidth]{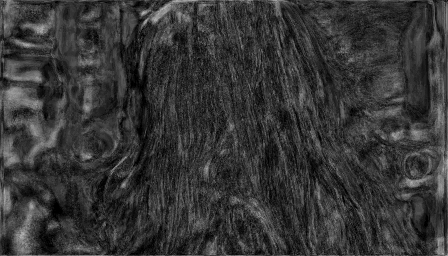}
		\centerline{(d) Distribution map}
		\vspace{0.0001cm}	
	\end{minipage}
	\figvspace
	\caption{Visualization  of (a) ground truth, (b) error residual (ER) map generated by GDConvNet with standard polynomial (c) error residual map generated by GDConvNet with clamped polynomial, (d) distribution map of sampling points beyond limits. The error residual map is calculated by $ER = MSE_{1\_inv} - MSE_{poly}$, where $MSE_{1\_inv}$ denotes the mean squared error map between the ground truth and the result generated by GDConvNet with 1D Inv, and $MSE_{poly}$ is similarly defined for polynomial interpolation.}
	\label{fig:beyond limits}
\end{figure}

To provide supporting evidence for our conjecture, we count the number of sampling points whose values are beyond the upper or lower limit. As shown in Table~\ref{tab:beyond limits}, for the upper GDCM used to synthesize intermediate frame ${I}'_t$, there are $10.8\%$ and $22.6\%$ sampling points beyond the upper limit and lower limit respectively. As for the lower GDCM used to predict the context map ${C}'_t$, $9.3\%$ and $10.0\%$ points are beyond the upper limit and  the lower limit, respectively. We then clamp those values to their associated limits and reevaluate the model on the test datasets. As shown in Table~\ref{tab:numerical interpolation performance}, indeed, forcing the values of sampling points to stay in the range set by support points jeopardizes the performance.

\begin{table*}[t]
	\renewcommand\arraystretch{1}
	\caption{\label{tab:comparison with SOTA}Quantitative comparisons of different VFI methods on Vimeo90K Septulet test set, Gopro dataset and Adobe240 dataset, where the first place and second place are highlighted in red and blue, respectively.}
	\vspace{-2mm}
	\footnotesize
	\center
	\renewcommand\tabcolsep{15.0pt}
	\resizebox{\textwidth}{!}{\begin{tabular}{cccccccc}
			\toprule
			\multicolumn{1}{c}{\multirow{2}*{Method}} 
			&{\#Parameters} 
			&\multicolumn{2}{c}{Vimeo90K} 
			&\multicolumn{2}{c}{Gopro}
			&\multicolumn{2}{c}{Adobe240} \\  
			\cmidrule(r){3-4} 
			\cmidrule(r){5-6}
			\cmidrule(r){7-8}
			& (million) & PSNR & SSIM & PSNR & SSIM & PSNR &SSIM\\
			\hline
			Ours-Poly	&$5.1$ &$\textbf{{\color{red}{35.58}}}$	&$\textbf{{\color{red}{0.9580}}}$	&$\textbf{{\color{red}{30.49}}}$	&${{\color{blue}{0.9180}}}$ & $\textbf{{\color{red}{34.53}}}$ & $\textbf{{\color{red}{0.9456}}}$\\
			Ours-Poly* &$5.1$ & $35.01$ & $0.9558$ & $30.12$ & $0.9100$ &${\color{blue}{34.12}}$ & ${\color{blue}{0.9422}}$\\
			\hline
			AdaCoF &$21.8$ &$33.92$& $0.9453$ &$28.45$ &$0.8734$ &$33.17$ &$0.9305$\\
			QVI &$29.2$ &${\color{blue}{35.19}}$ &${\color{blue}{0.9563}}$ &	${\color{blue}{30.24}}$ &$\textbf{\color{red}{0.9230}}$ &$33.06$ &$0.9393$\\
			Slomo 	&$39.6$ & $33.73$ &$0.9453$ &$28.50$ &$0.8827$ &$31.94$ &$0.9264$\\
			SepConv	&$21.6$ &$33.65$ &$0.9435$ &$28.66$ &$0.8798$ &$33.41$ &$0.9349$\\
			DVF &$3.8$ & $30.79$	&$0.8912$ &$25.13$ &$0.7633$ &$22.33$ &$0.6159$\\
			Phase &$-$ & $30.52$ &$0.8854$ &$26.17$	&$0.8135$ &$31.20$ &$0.8930$\\
			\bottomrule
			
	\end{tabular}}
	
\end{table*}
\begin{figure*}[!htb]
	\begin{minipage}[htb]{0.135\linewidth}
		\vspace{1mm}
		\centering
		{\includegraphics[width=\linewidth]{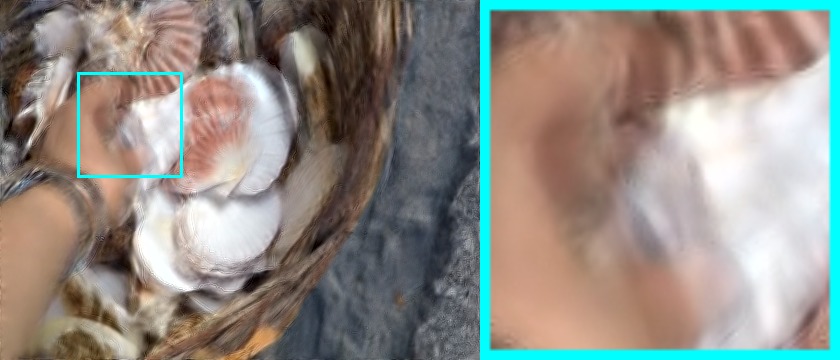}\vspace{1mm}}
		{\includegraphics[width=\linewidth]{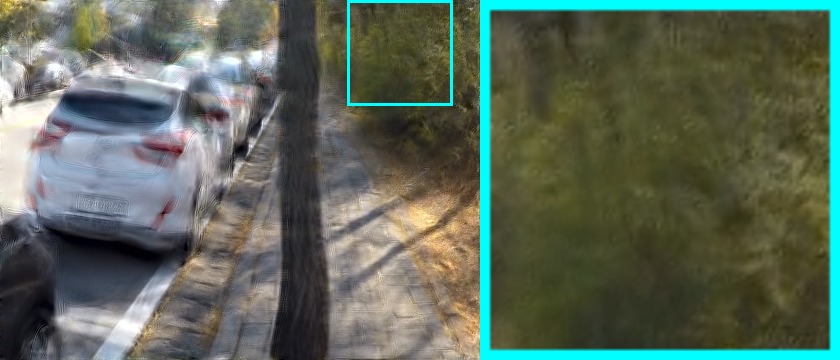}\vspace{1mm}}
		{\includegraphics[width=\linewidth]{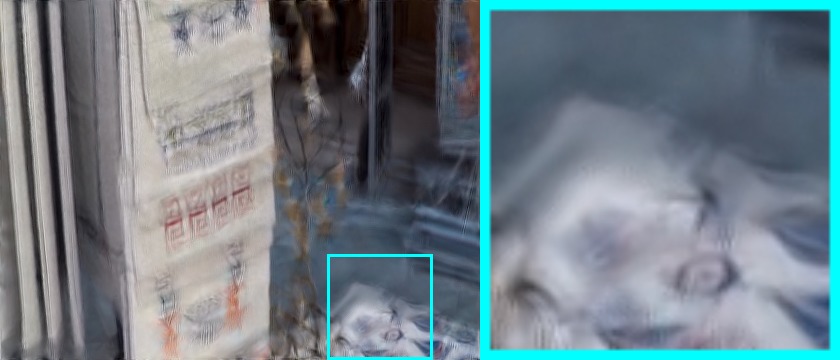}\vspace{1mm}}
		{\includegraphics[width=\linewidth]{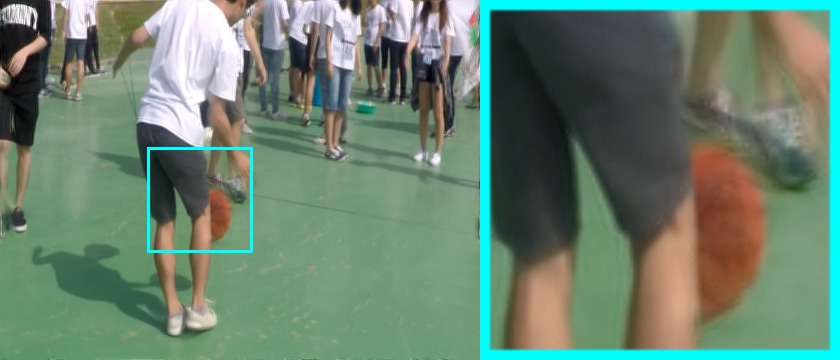}\vspace{1mm}}
		{\includegraphics[width=\linewidth]{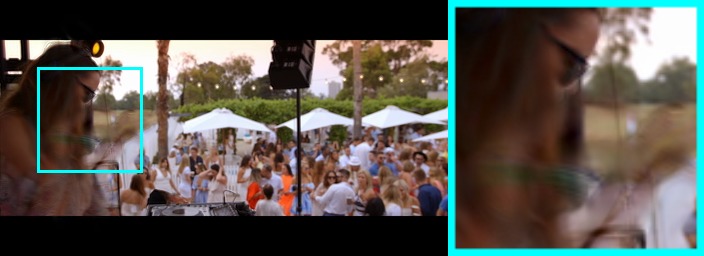}\vspace{1mm}}
		{\includegraphics[width=\linewidth]{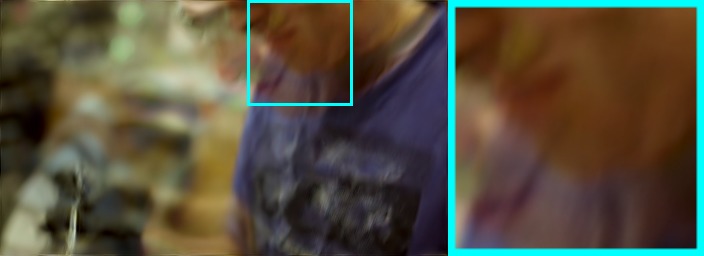}\vspace{1mm}}
		{\includegraphics[width=\linewidth]{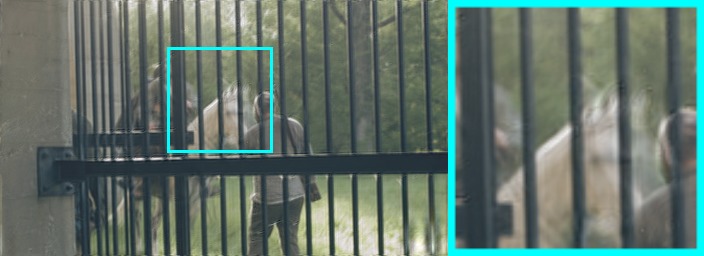}\vspace{1mm}}
		{\includegraphics[width=\linewidth]{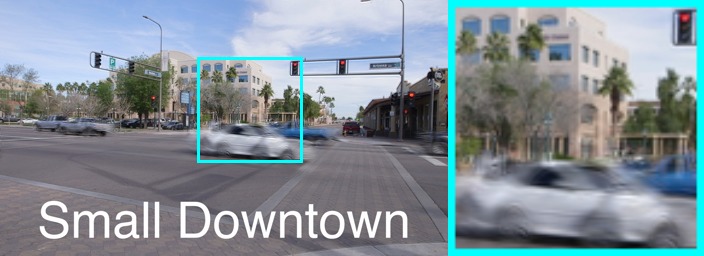}\vspace{1mm}}
		{\includegraphics[width=\linewidth]{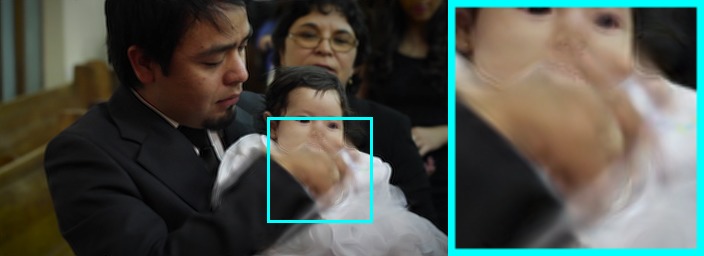}\vspace{1mm}}
		{\includegraphics[width=\linewidth]{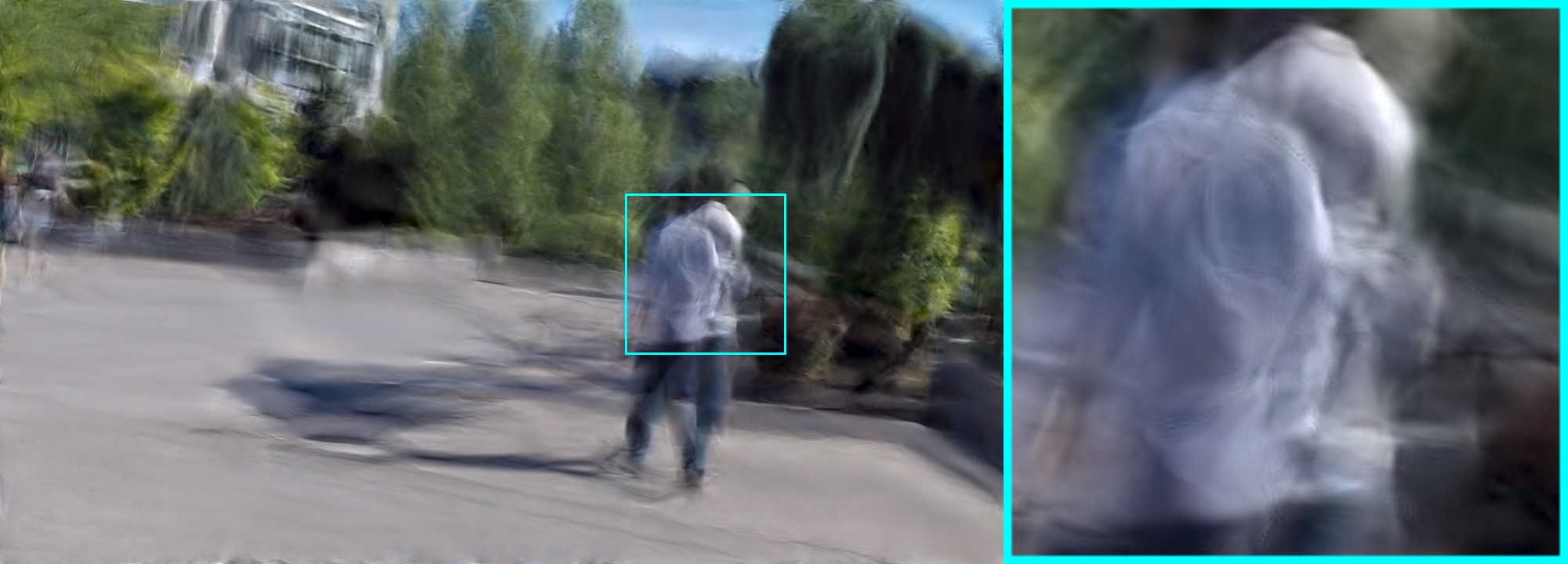}\vspace{1mm}}
		{\includegraphics[width=\linewidth]{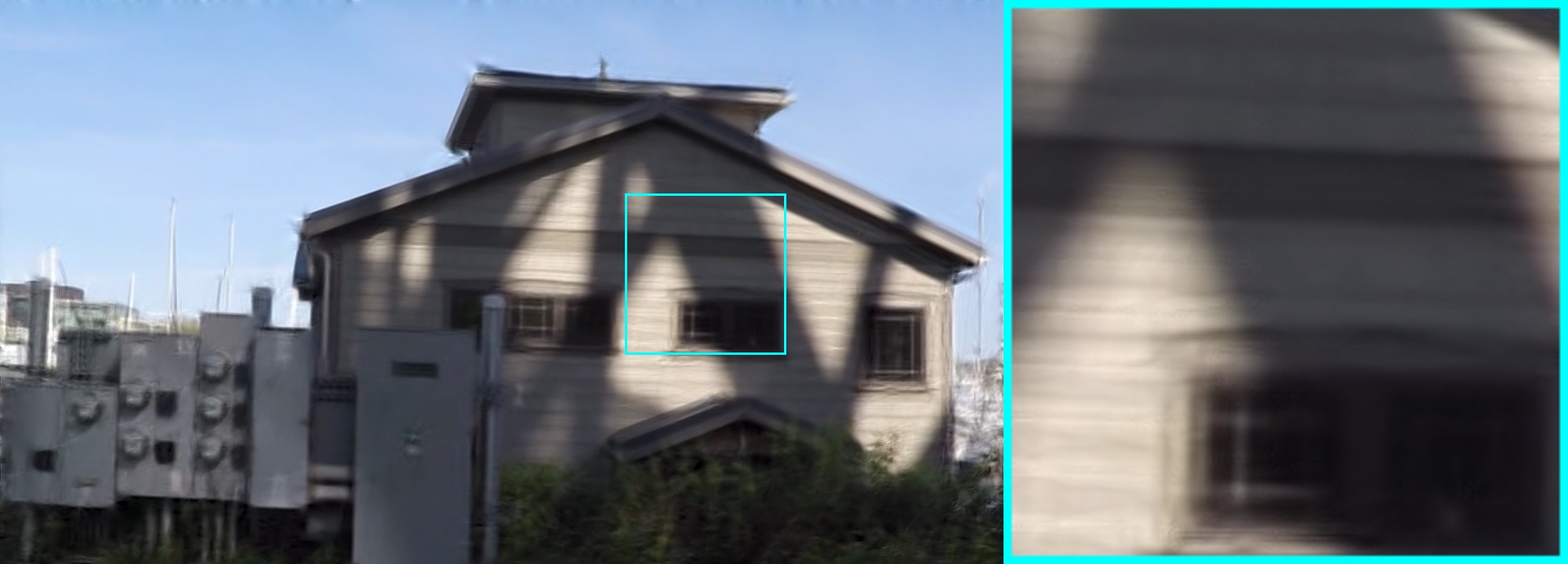}\vspace{1mm}}
		{\includegraphics[width=\linewidth]{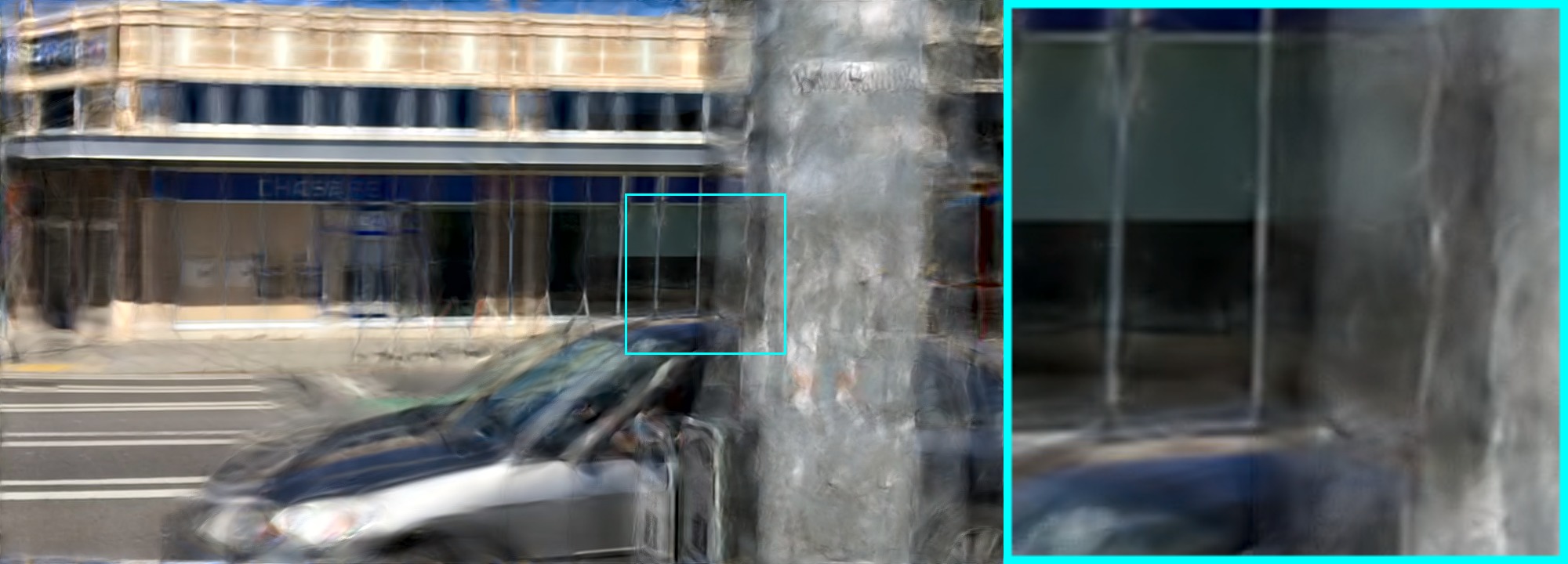}}
		\vspace{1cm}
		\centerline{Phase}
		\vspace{0.1cm}
	\end{minipage}
	\begin{minipage}[htb]{0.135\linewidth}
		\vspace{1mm}
		\centering
		{\includegraphics[width=\linewidth]{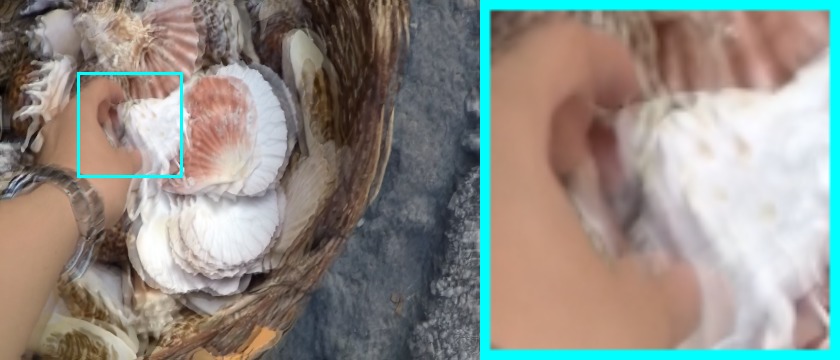}\vspace{1mm}}
		{\includegraphics[width=\linewidth]{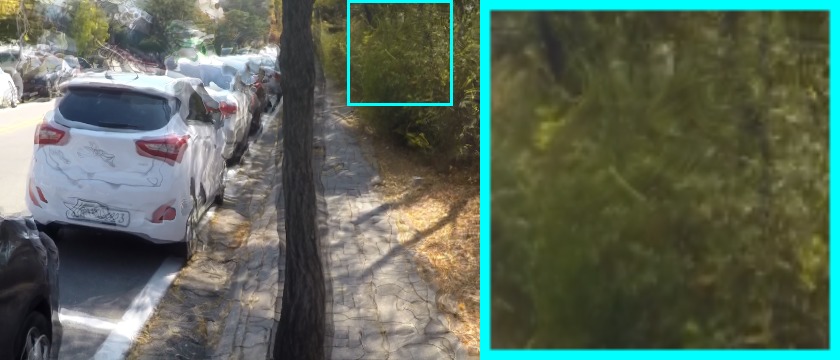}\vspace{1mm}}
		{\includegraphics[width=\linewidth]{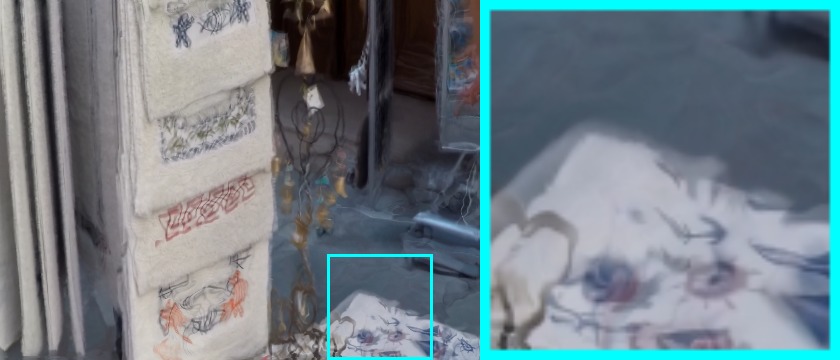}\vspace{1mm}}
		{\includegraphics[width=\linewidth]{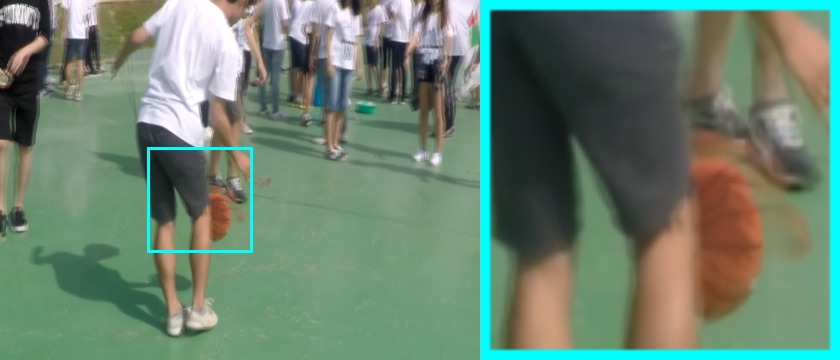}\vspace{1mm}}
		{\includegraphics[width=\linewidth]{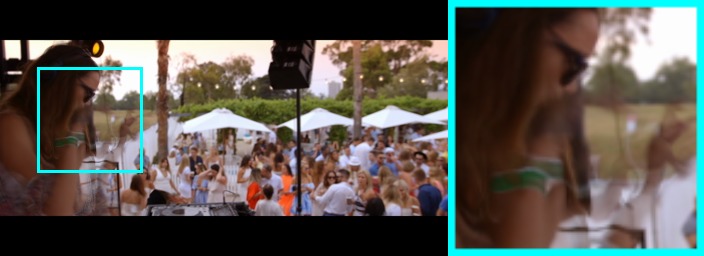}\vspace{1mm}}
		{\includegraphics[width=\linewidth]{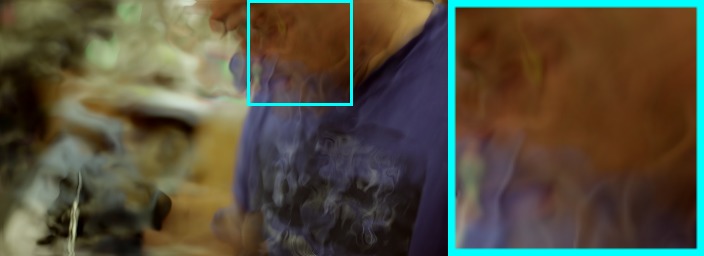}\vspace{1mm}}
		{\includegraphics[width=\linewidth]{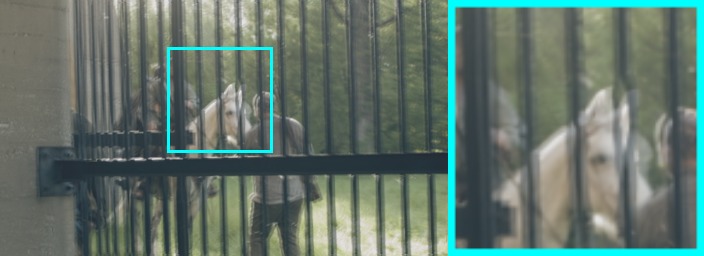}\vspace{1mm}}
		{\includegraphics[width=\linewidth]{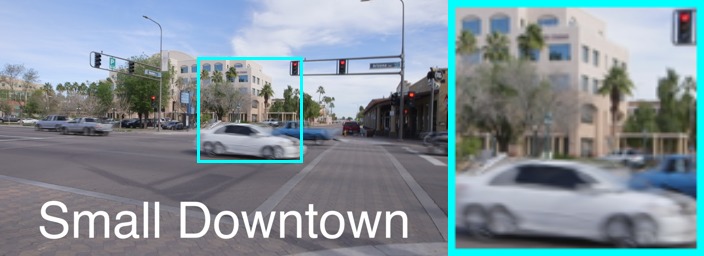}\vspace{1mm}}
		{\includegraphics[width=\linewidth]{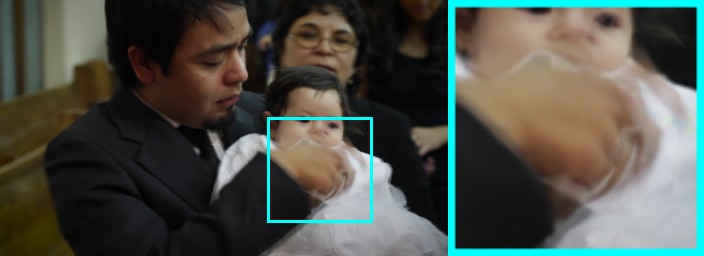}\vspace{1mm}}
		{\includegraphics[width=\linewidth]{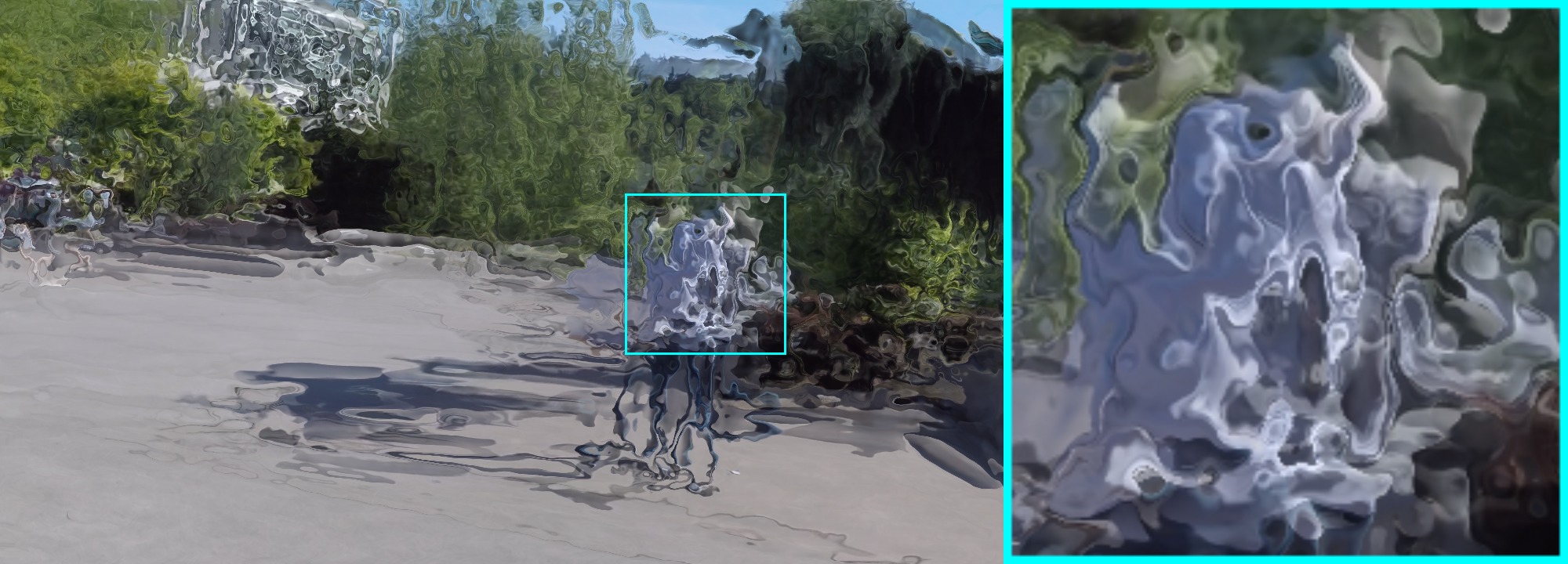}\vspace{1mm}}
		{\includegraphics[width=\linewidth]{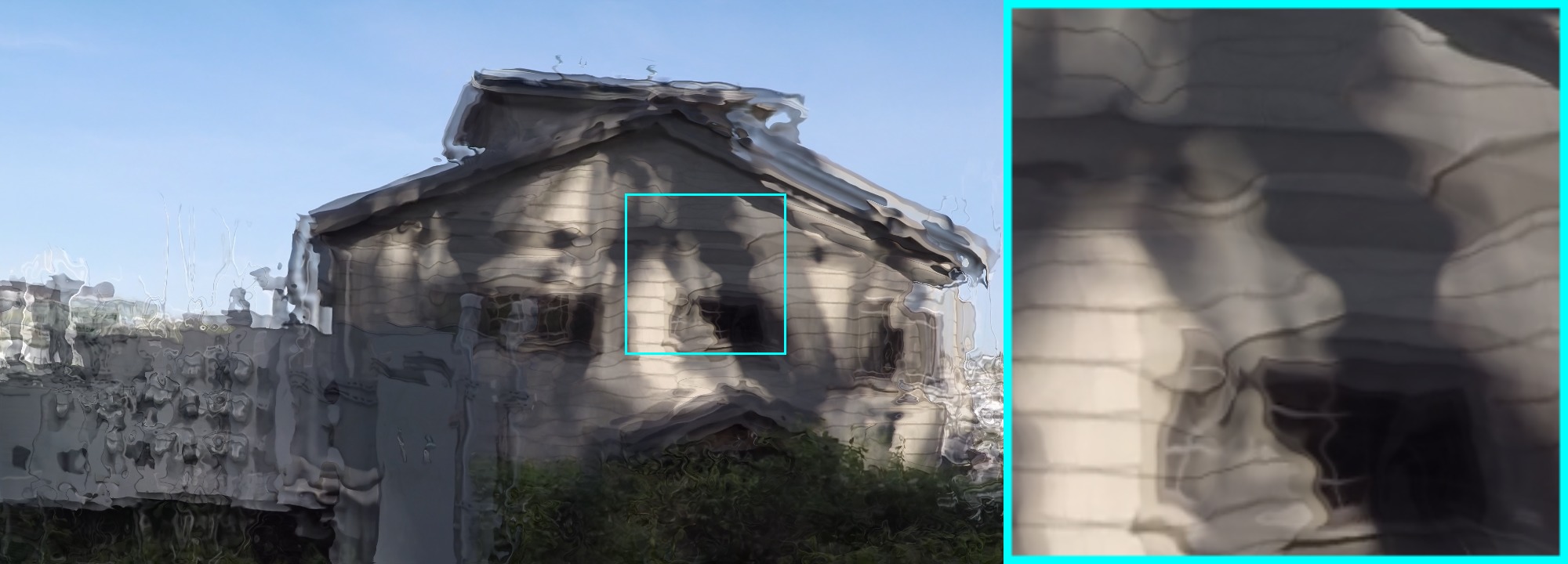}\vspace{1mm}}
		{\includegraphics[width=\linewidth]{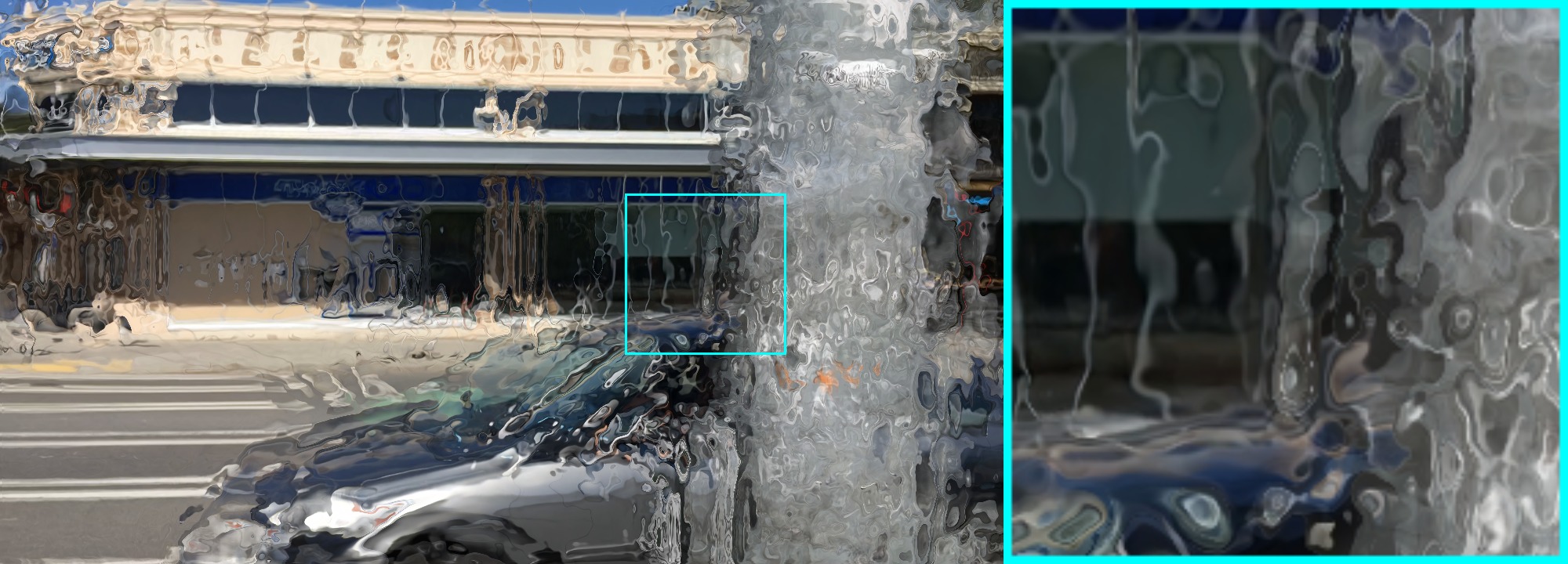}}
		\vspace{1cm}
		\centerline{DVF}
		\vspace{0.1cm}
	\end{minipage}
	\begin{minipage}[htb]{0.135\linewidth}
		\vspace{1mm}
		\centering
		{\includegraphics[width=\linewidth]{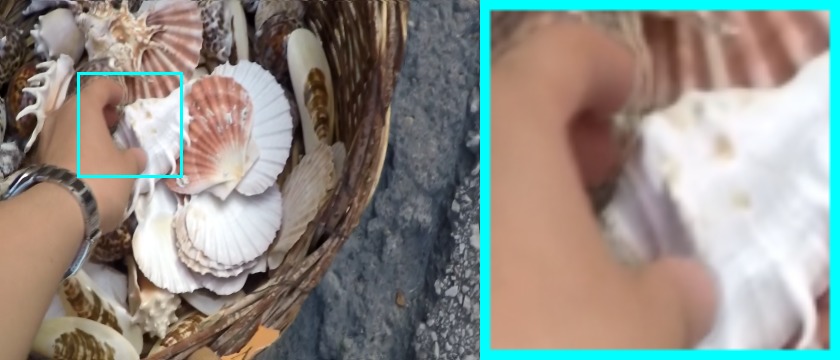}\vspace{1mm}}
		{\includegraphics[width=\linewidth]{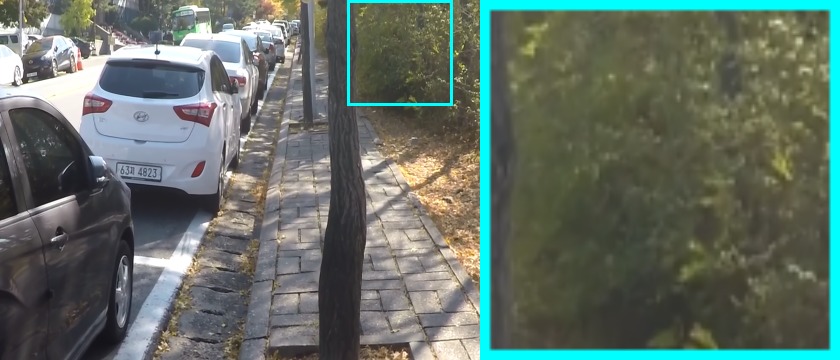}\vspace{1mm}}
		{\includegraphics[width=\linewidth]{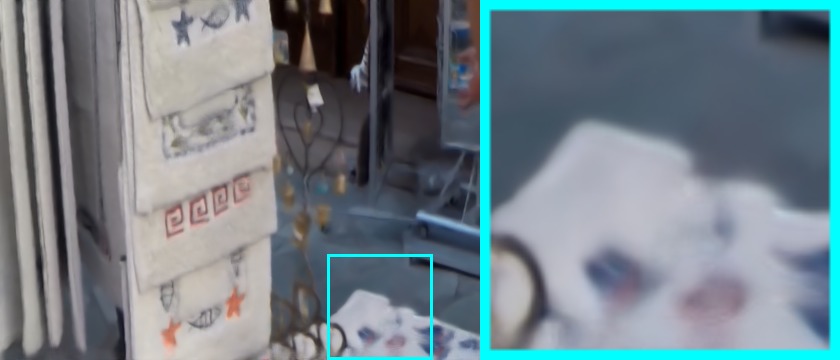}\vspace{1mm}}
		{\includegraphics[width=\linewidth]{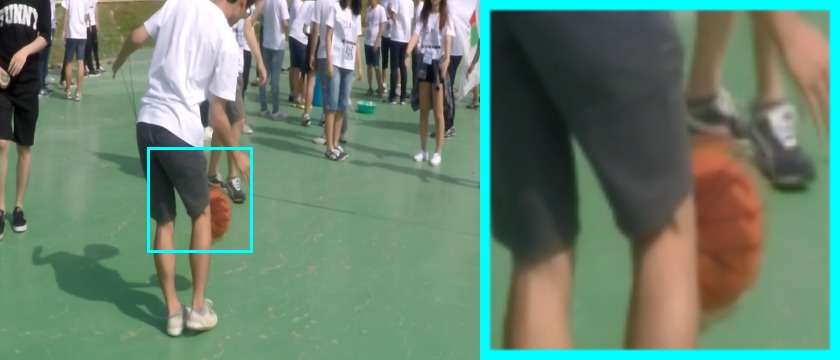}\vspace{1mm}}
		{\includegraphics[width=\linewidth]{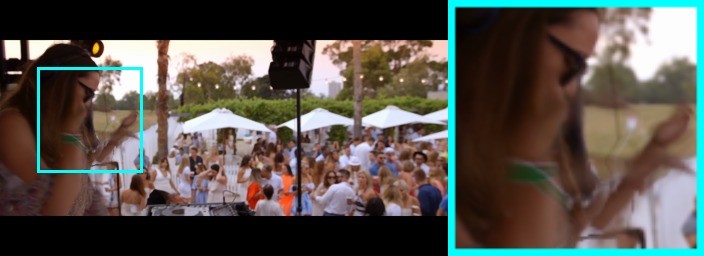}\vspace{1mm}}
		{\includegraphics[width=\linewidth]{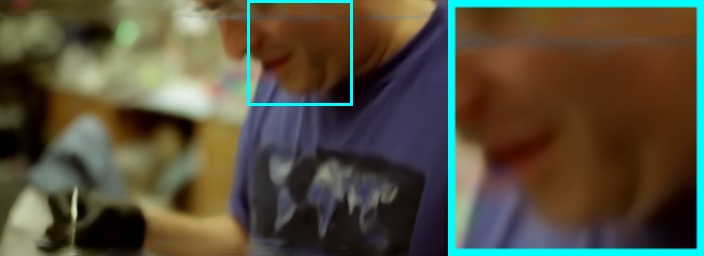}\vspace{1mm}}
		{\includegraphics[width=\linewidth]{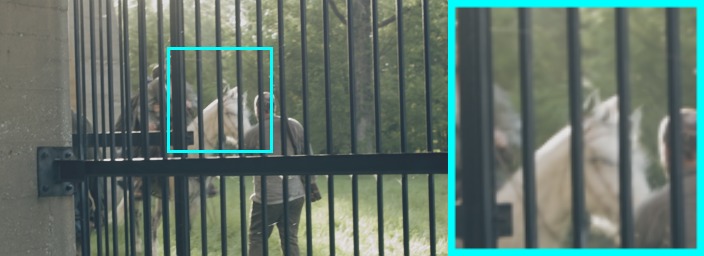}\vspace{1mm}}
		{\includegraphics[width=\linewidth]{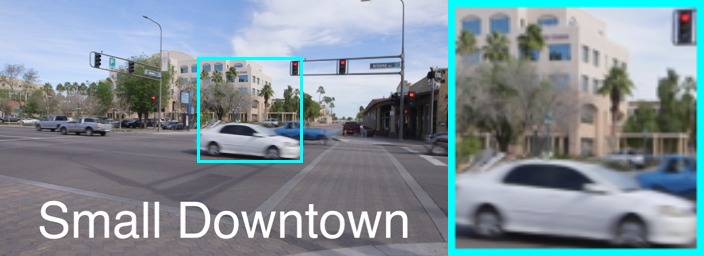}\vspace{1mm}}
		{\includegraphics[width=\linewidth]{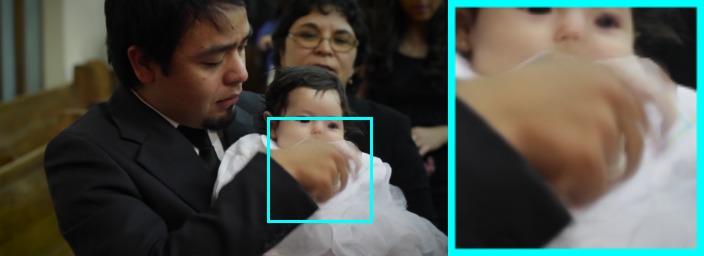}\vspace{1mm}}
		{\includegraphics[width=\linewidth]{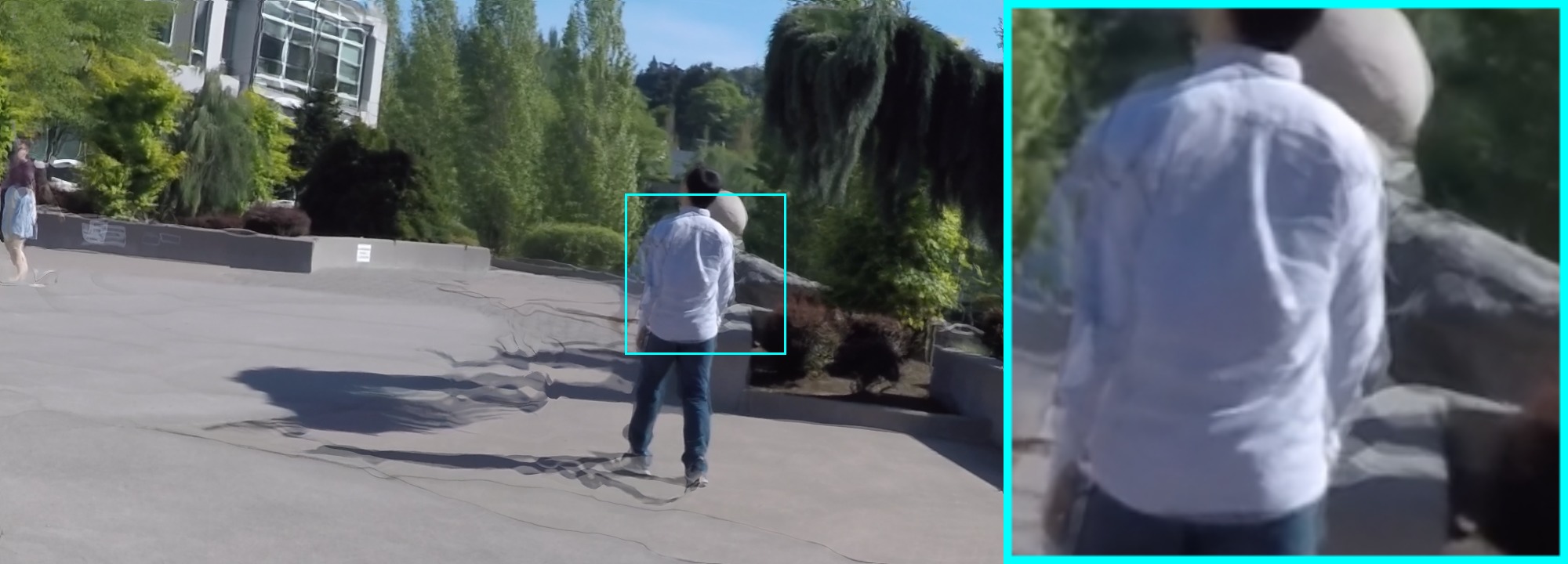}\vspace{1mm}}
		{\includegraphics[width=\linewidth]{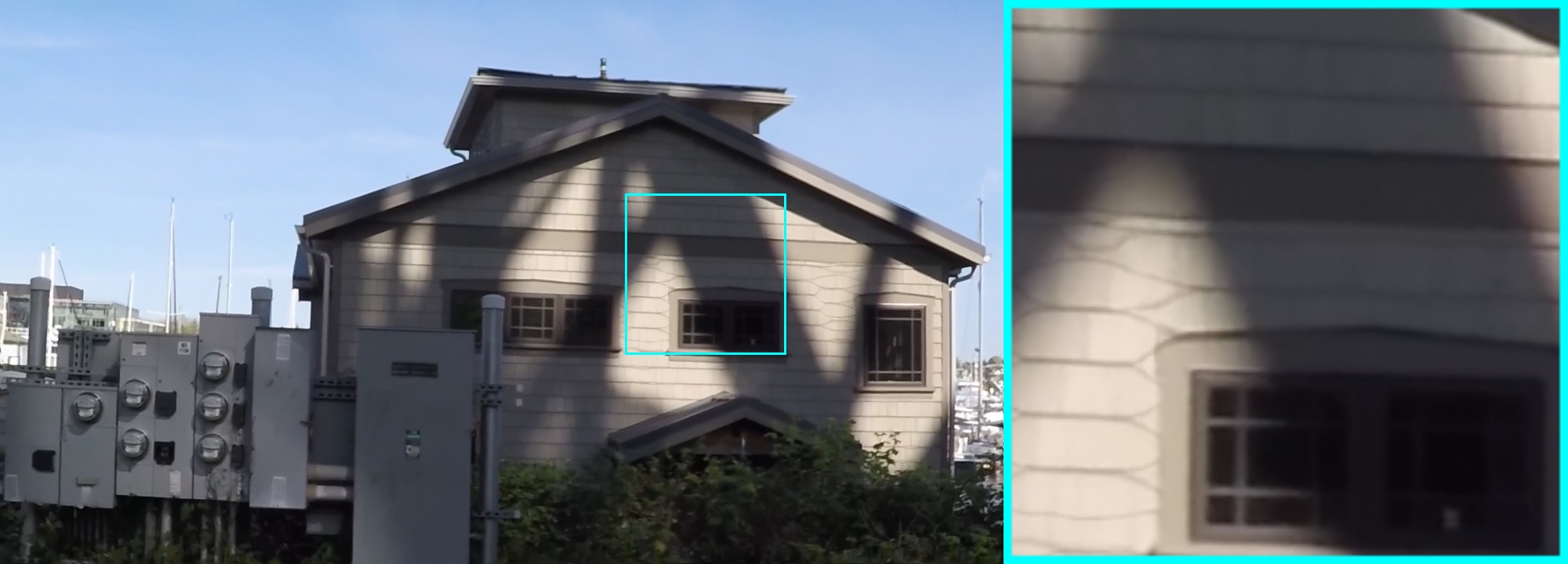}\vspace{1mm}}
		{\includegraphics[width=\linewidth]{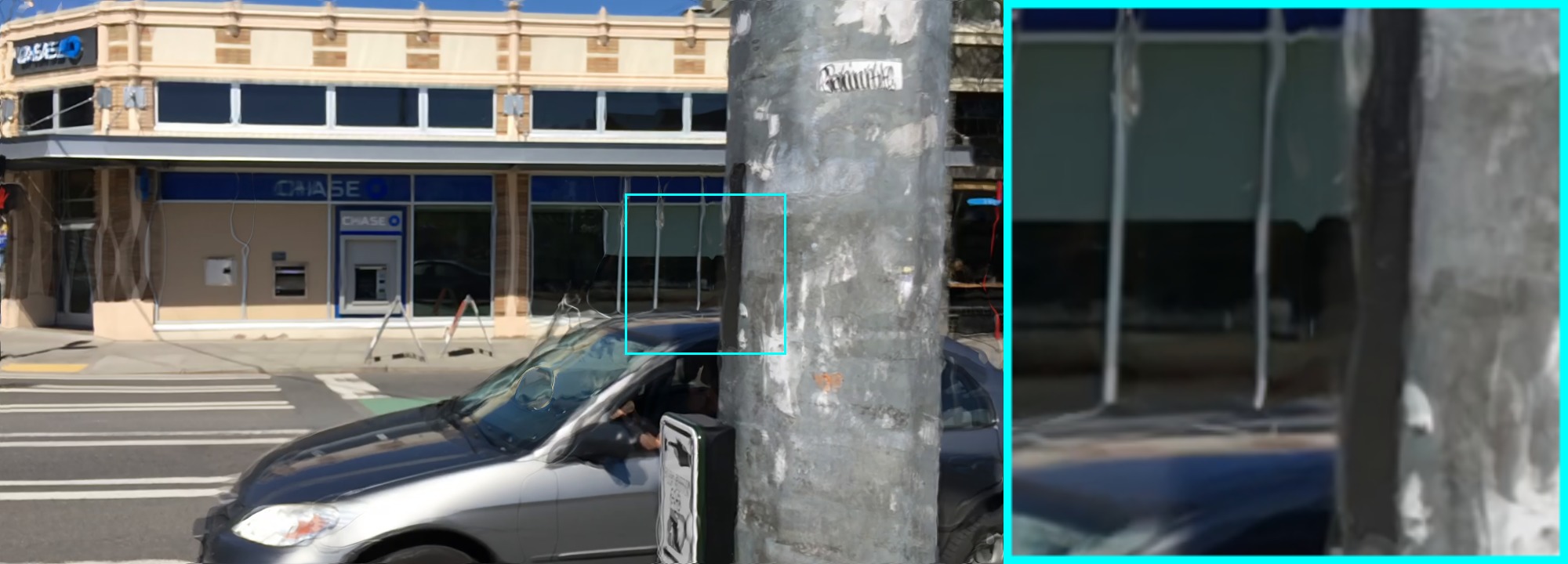}}
		\vspace{1cm}
		\centerline{Slomo}
		\vspace{0.1cm}
	\end{minipage}
	\begin{minipage}[htb]{0.135\linewidth}
		\vspace{1mm}
		\centering
		{\includegraphics[width=\linewidth]{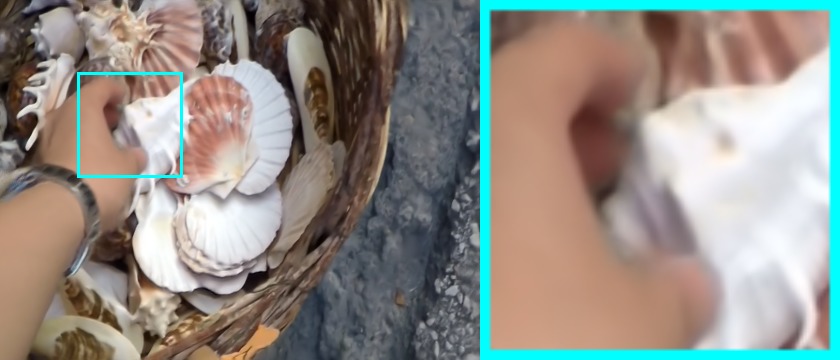}\vspace{1mm}}
		{\includegraphics[width=\linewidth]{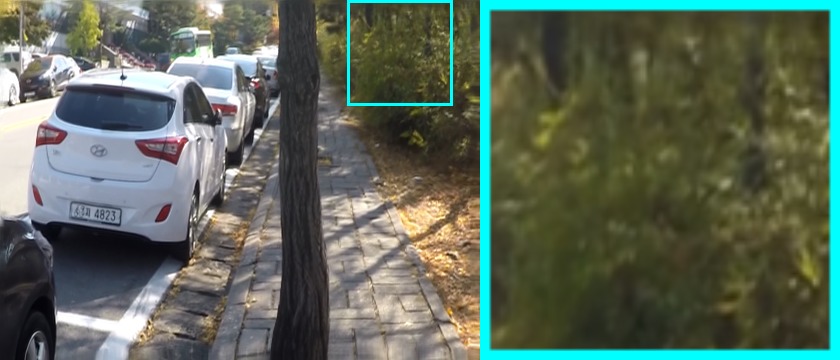}\vspace{1mm}}
		{\includegraphics[width=\linewidth]{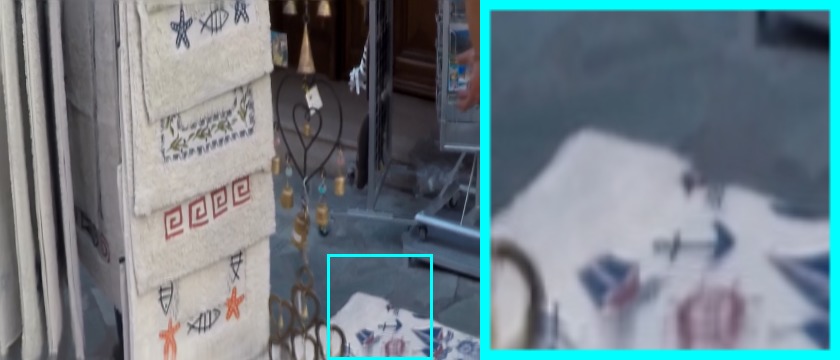}\vspace{1mm}}
		{\includegraphics[width=\linewidth]{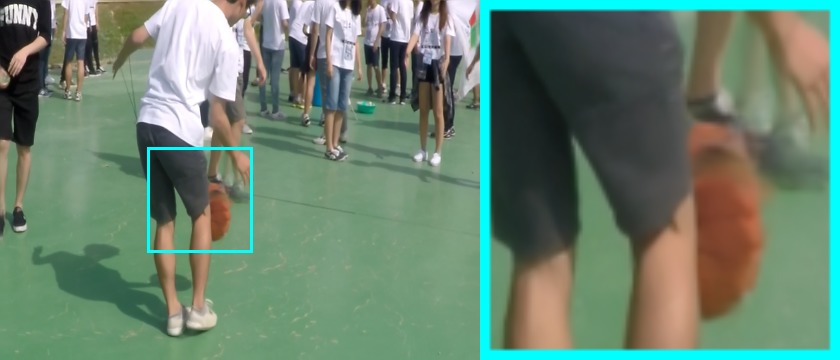}\vspace{1mm}}
		{\includegraphics[width=\linewidth]{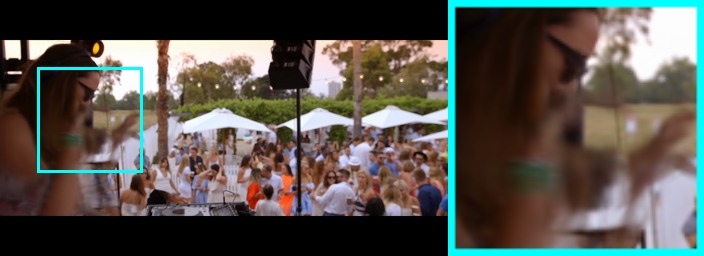}\vspace{1mm}}
		{\includegraphics[width=\linewidth]{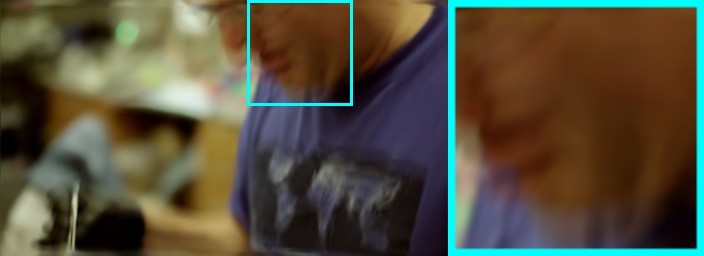}\vspace{1mm}}
		{\includegraphics[width=\linewidth]{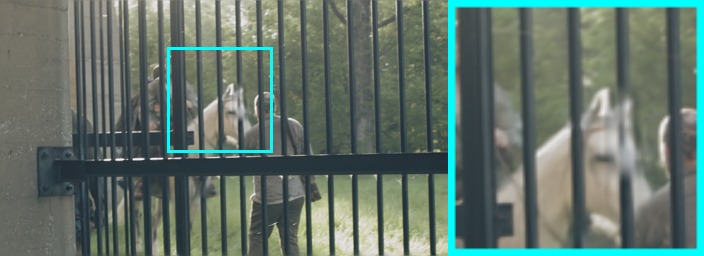}\vspace{1mm}}
		{\includegraphics[width=\linewidth]{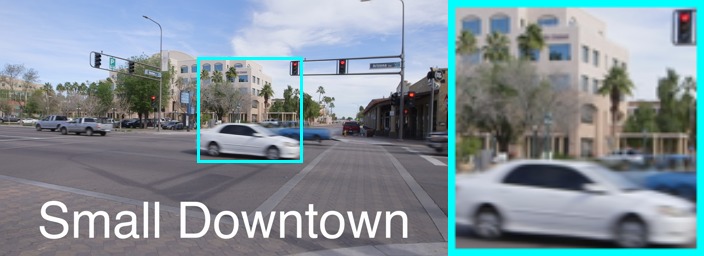}\vspace{1mm}}
		{\includegraphics[width=\linewidth]{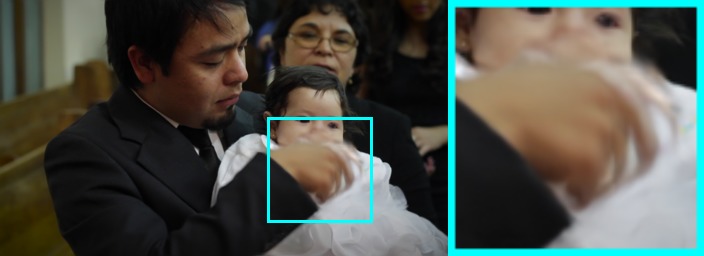}\vspace{1mm}}
		{\includegraphics[width=\linewidth]{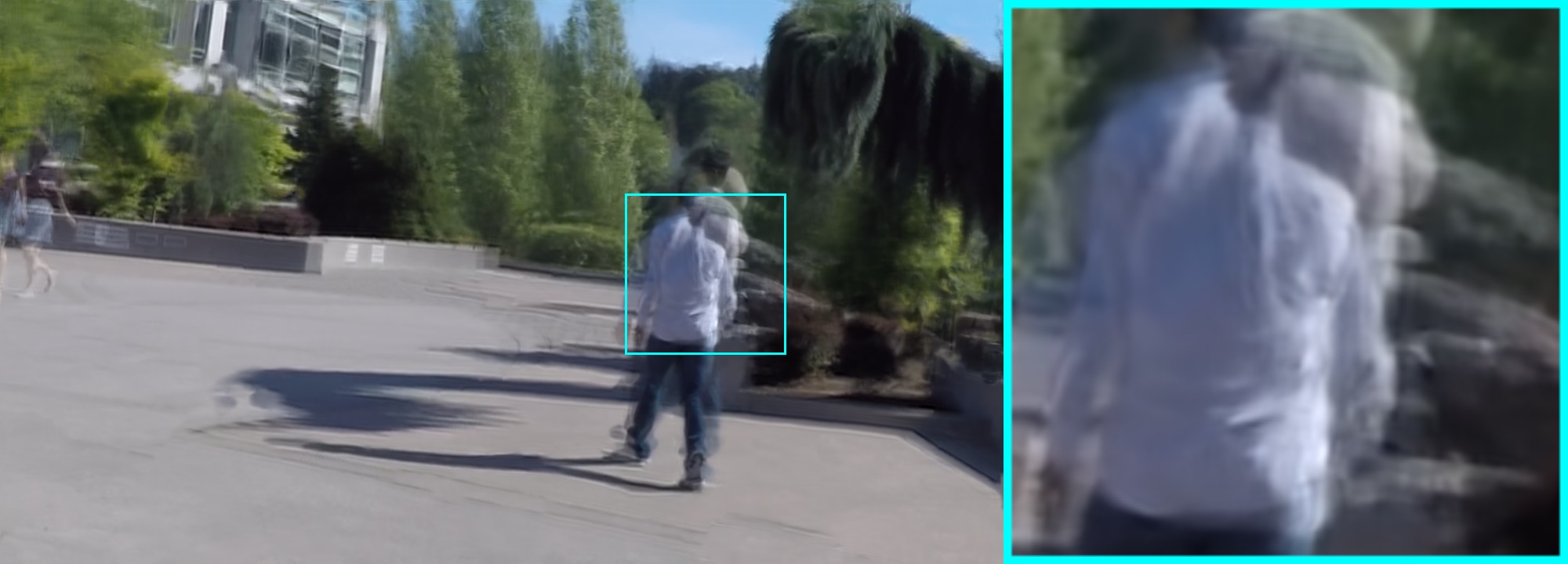}\vspace{1mm}}
		{\includegraphics[width=\linewidth]{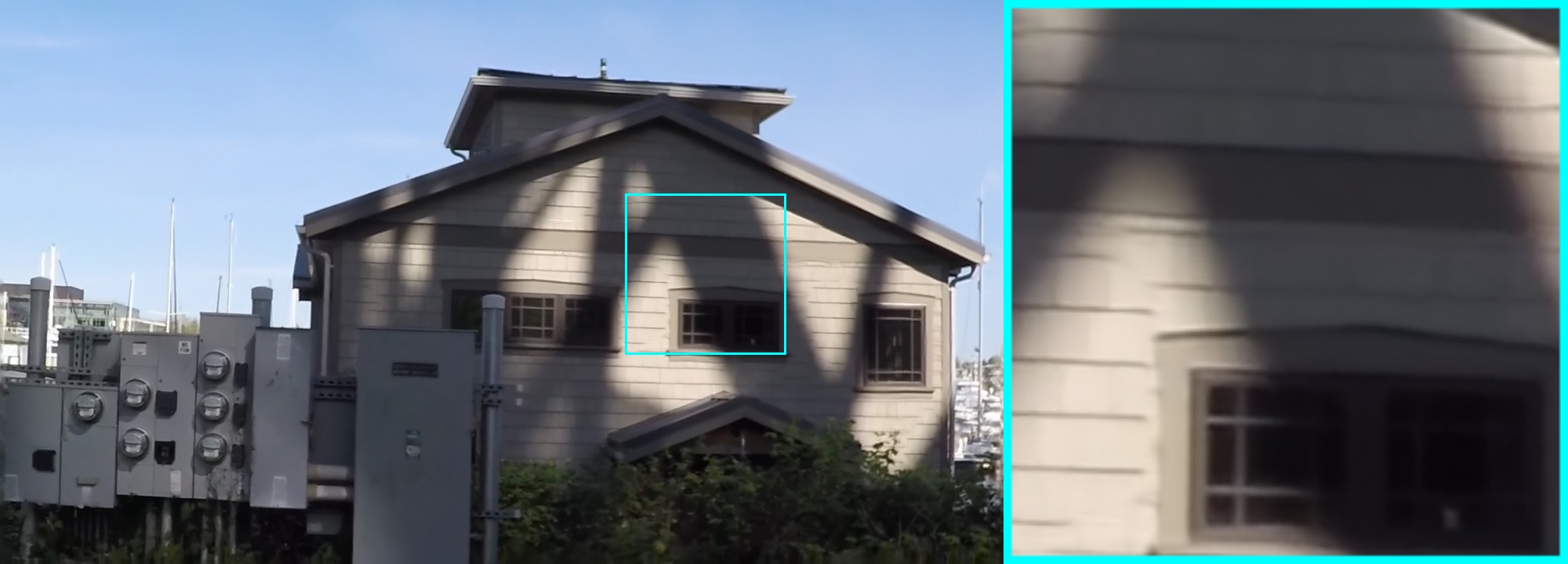}\vspace{1mm}}
		{\includegraphics[width=\linewidth]{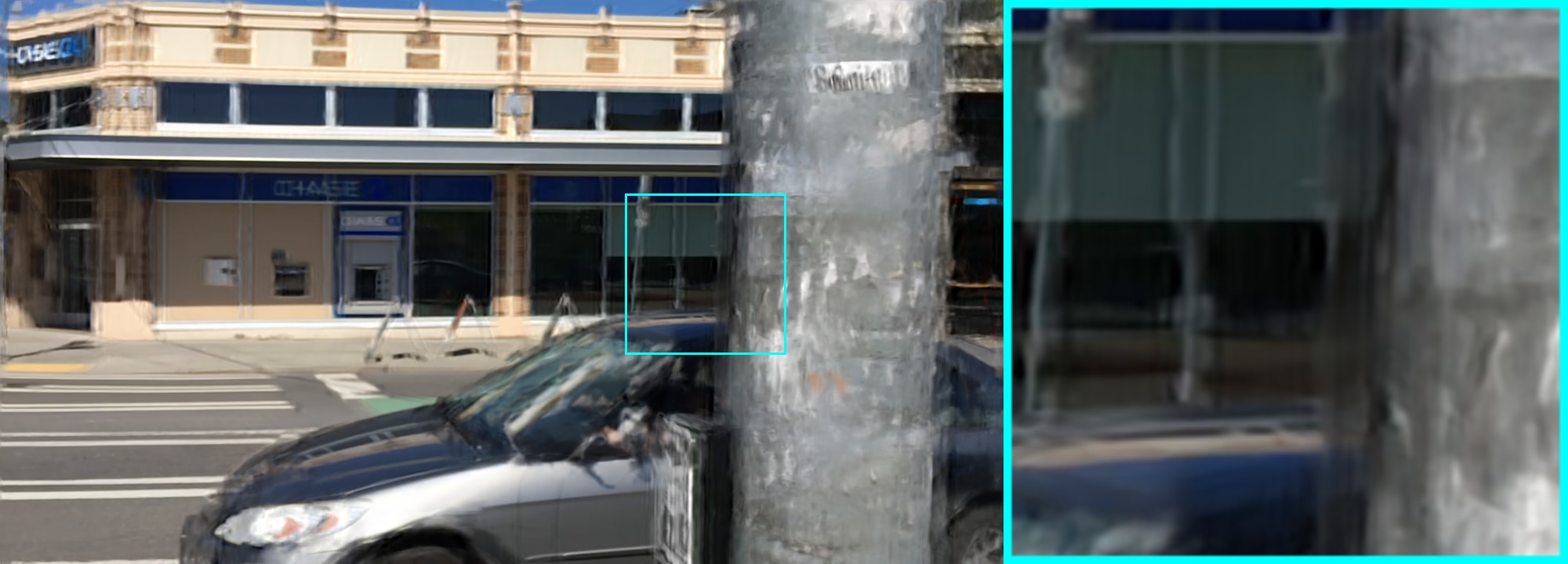}}
		\vspace{1cm}
		\centerline{SepConv}
		\vspace{0.1cm}
	\end{minipage}
	\begin{minipage}[htb]{0.135\linewidth}
		\vspace{1mm}
		\centering
		{\includegraphics[width=\linewidth]{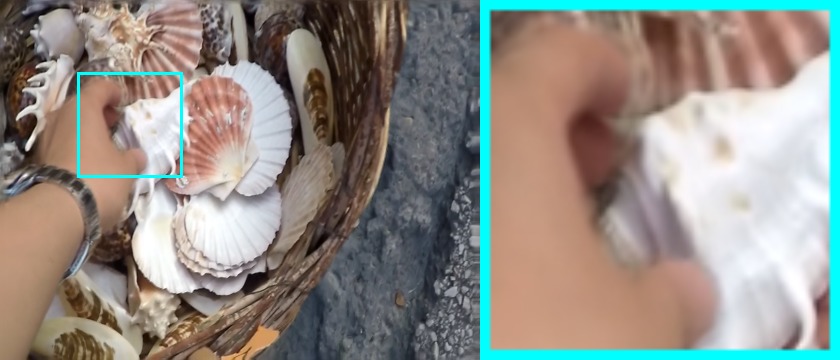}\vspace{1mm}}
		{\includegraphics[width=\linewidth]{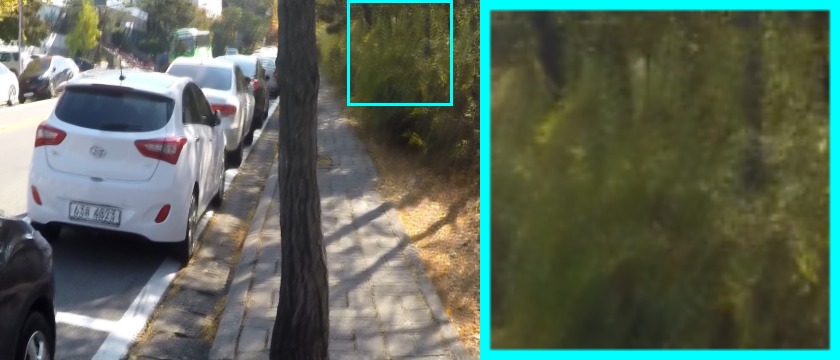}\vspace{1mm}}
		{\includegraphics[width=\linewidth]{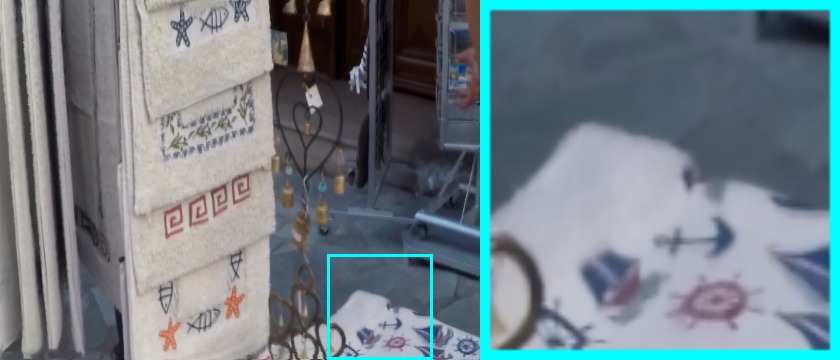}\vspace{1mm}}
		{\includegraphics[width=\linewidth]{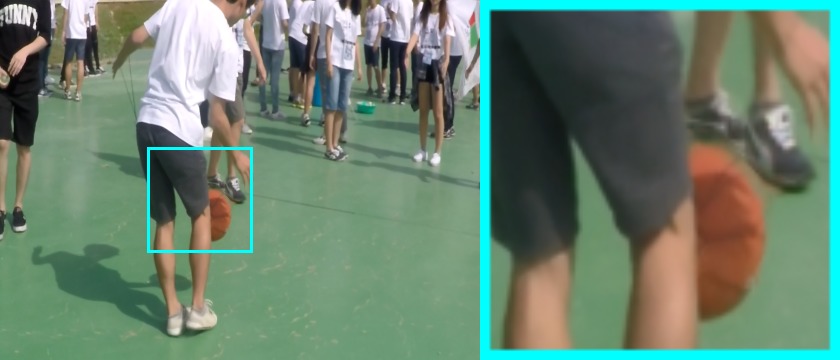}\vspace{1mm}}
		{\includegraphics[width=\linewidth]{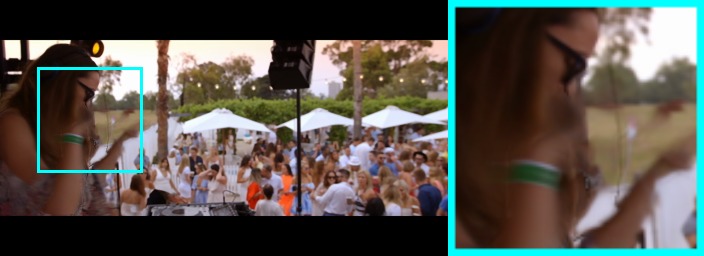}\vspace{1mm}}
		{\includegraphics[width=\linewidth]{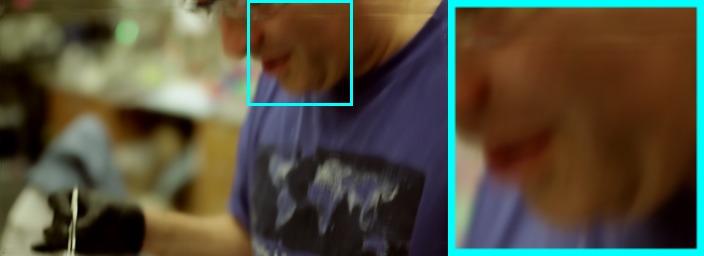}\vspace{1mm}}
		{\includegraphics[width=\linewidth]{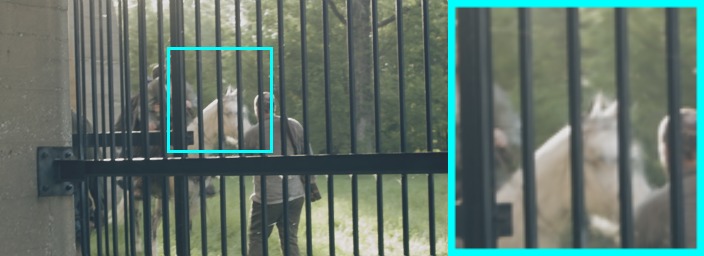}\vspace{1mm}}
		{\includegraphics[width=\linewidth]{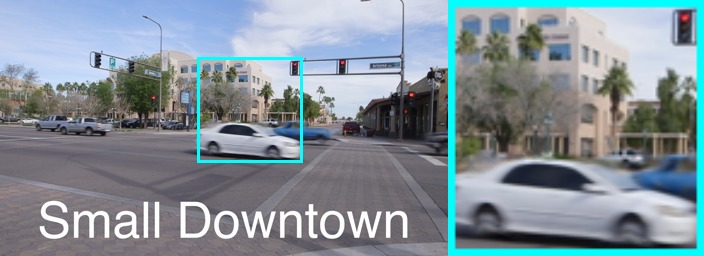}\vspace{1mm}}
		{\includegraphics[width=\linewidth]{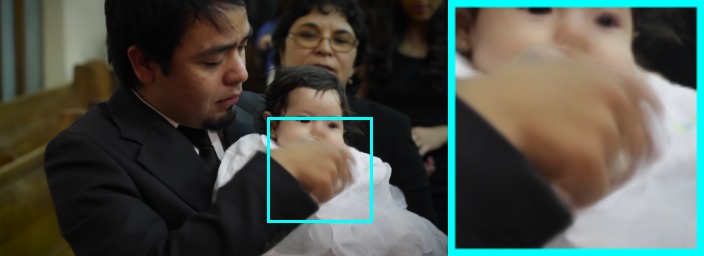}\vspace{1mm}}
		{\includegraphics[width=\linewidth]{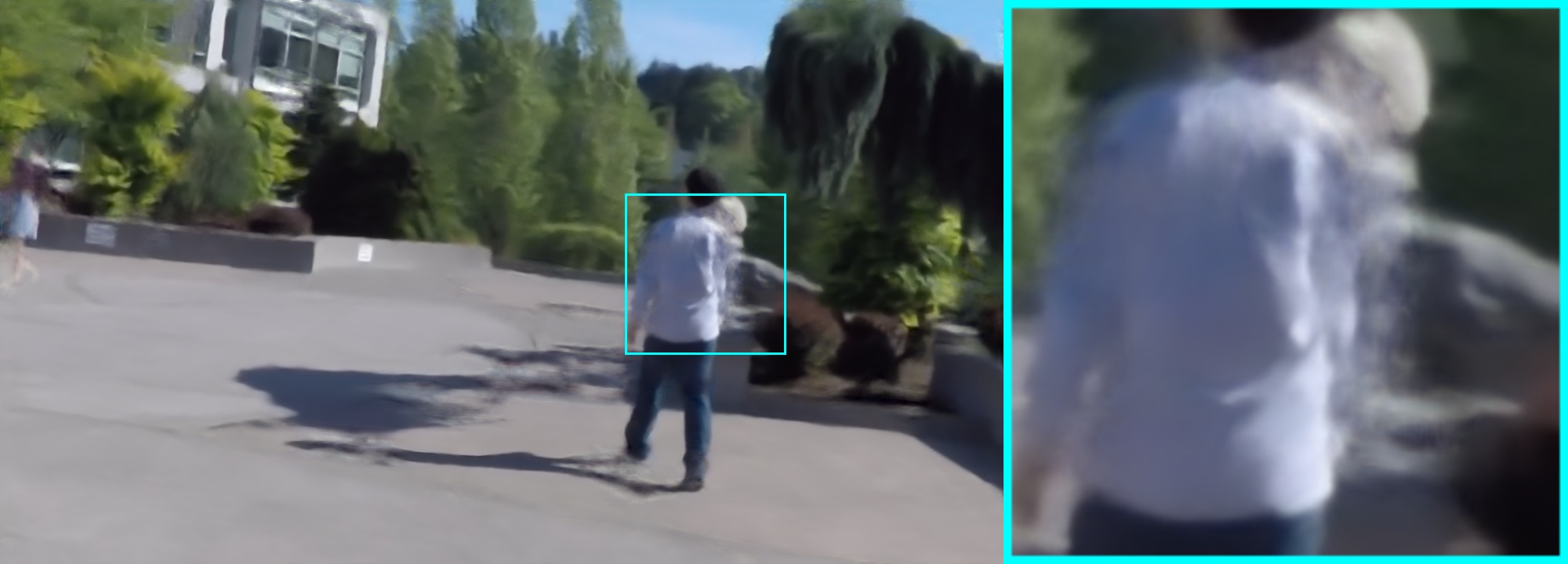}\vspace{1mm}}
		{\includegraphics[width=\linewidth]{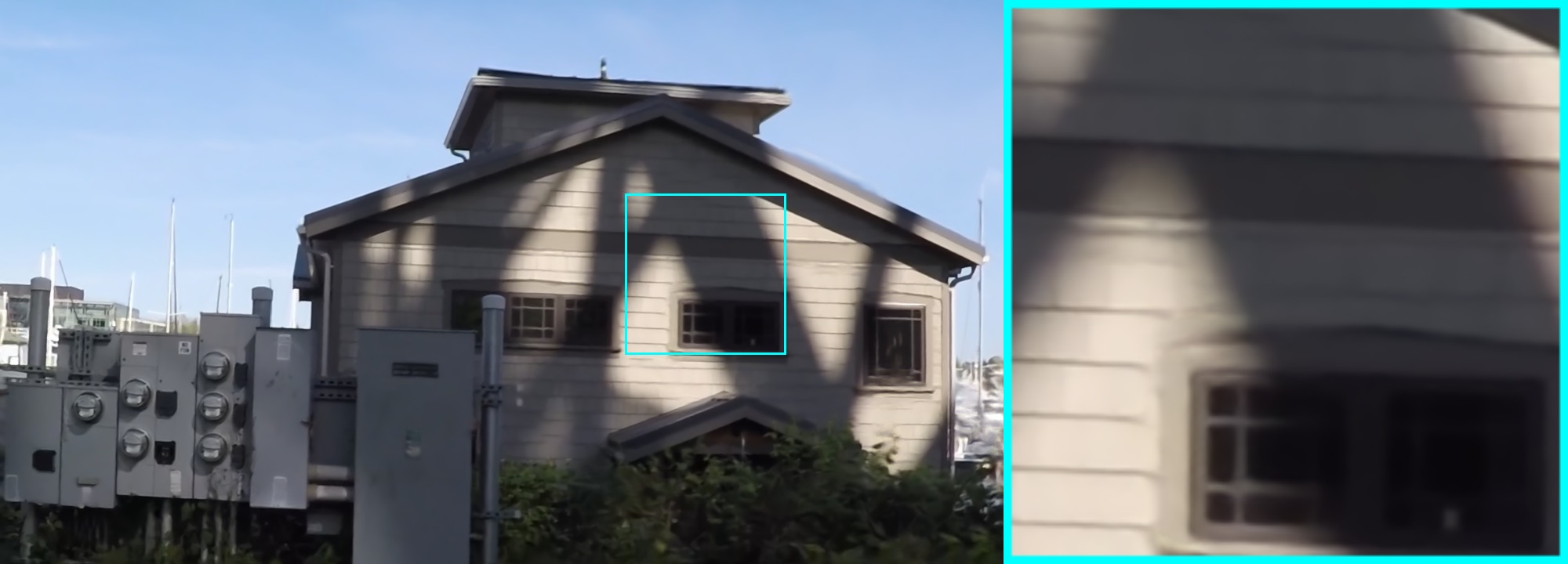}\vspace{1mm}}
		{\includegraphics[width=\linewidth]{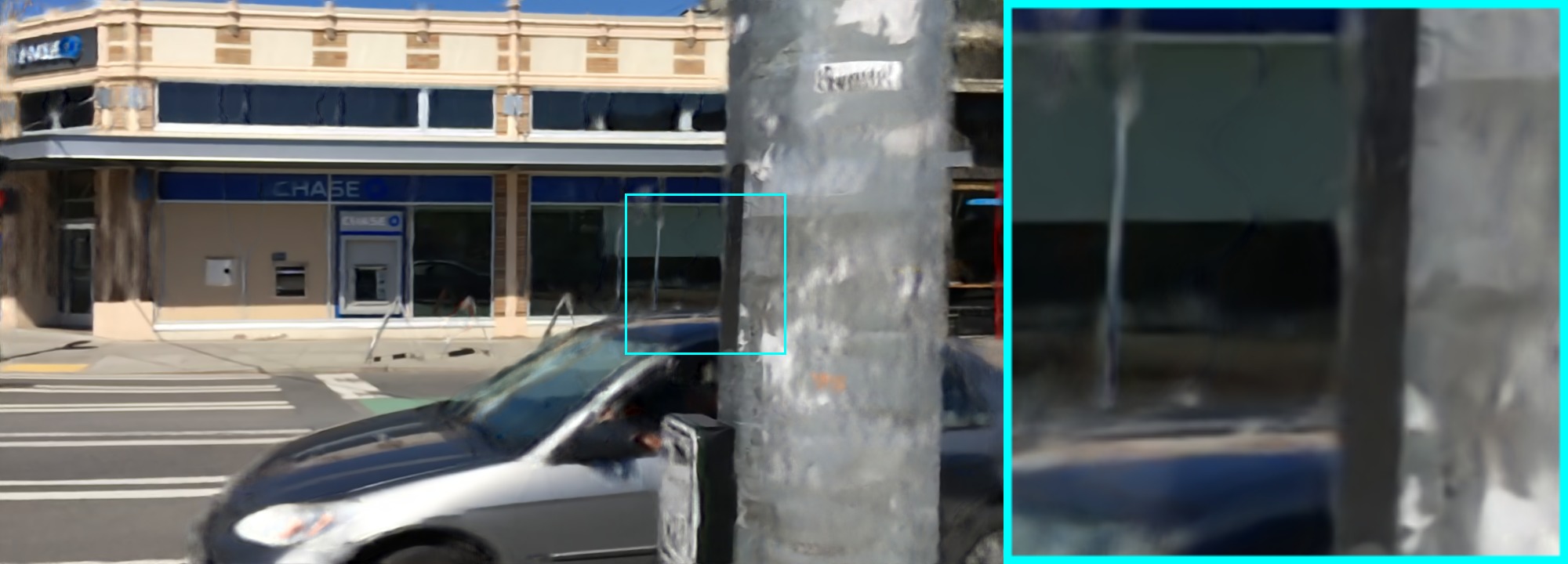}}
		\vspace{1cm}
		\centerline{QVI}
		\vspace{0.1cm}
	\end{minipage}
	\begin{minipage}[htb]{0.135\linewidth}
		\vspace{1mm}
		\centering
		{\includegraphics[width=\linewidth]{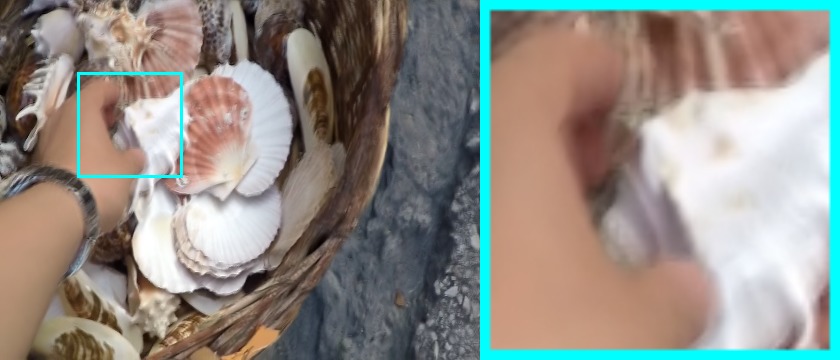}\vspace{1mm}}
		{\includegraphics[width=\linewidth]{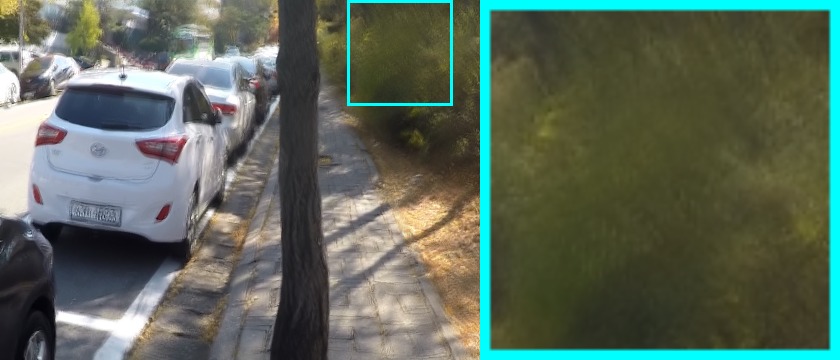}\vspace{1mm}}
		{\includegraphics[width=\linewidth]{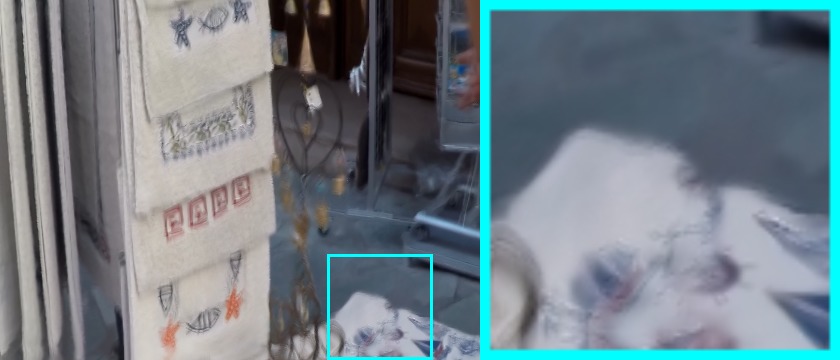}\vspace{1mm}}
		{\includegraphics[width=\linewidth]{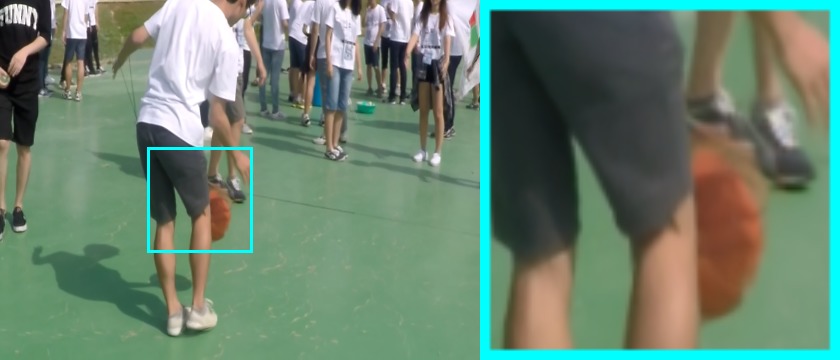}\vspace{1mm}}
		{\includegraphics[width=\linewidth]{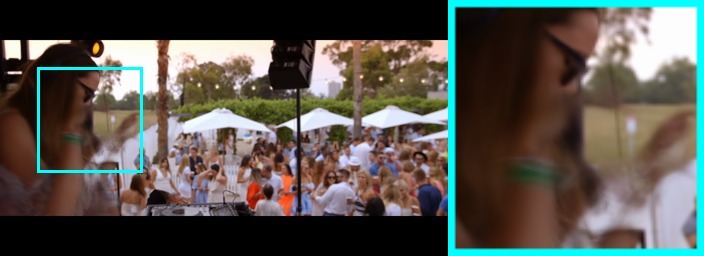}\vspace{1mm}}
		{\includegraphics[width=\linewidth]{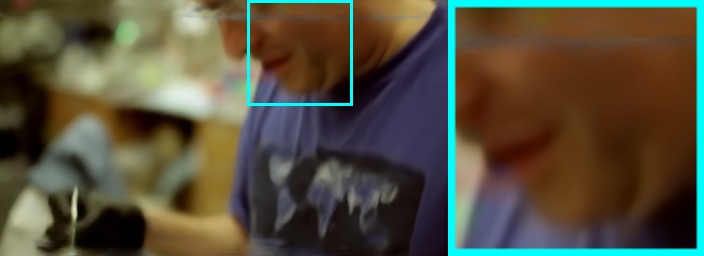}\vspace{1mm}}
		{\includegraphics[width=\linewidth]{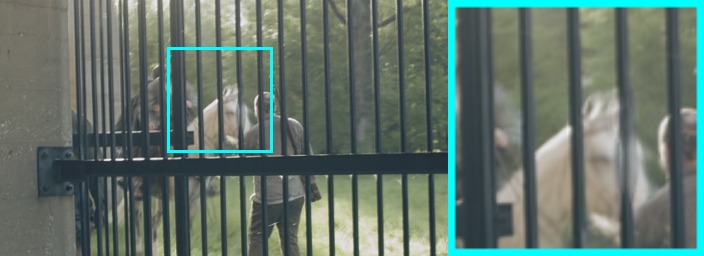}\vspace{1mm}}
		{\includegraphics[width=\linewidth]{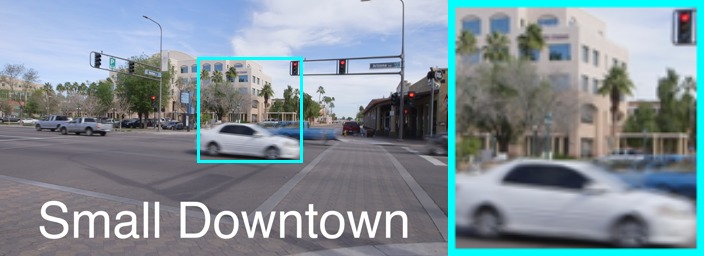}\vspace{1mm}}
		{\includegraphics[width=\linewidth]{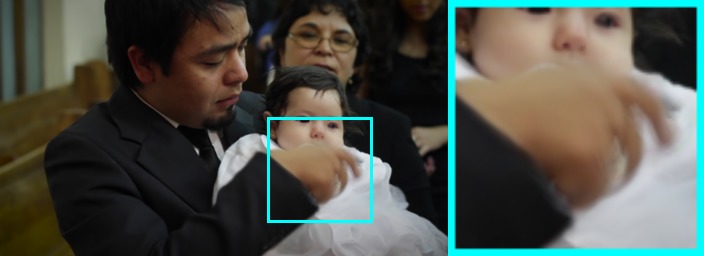}\vspace{1mm}}
		{\includegraphics[width=\linewidth]{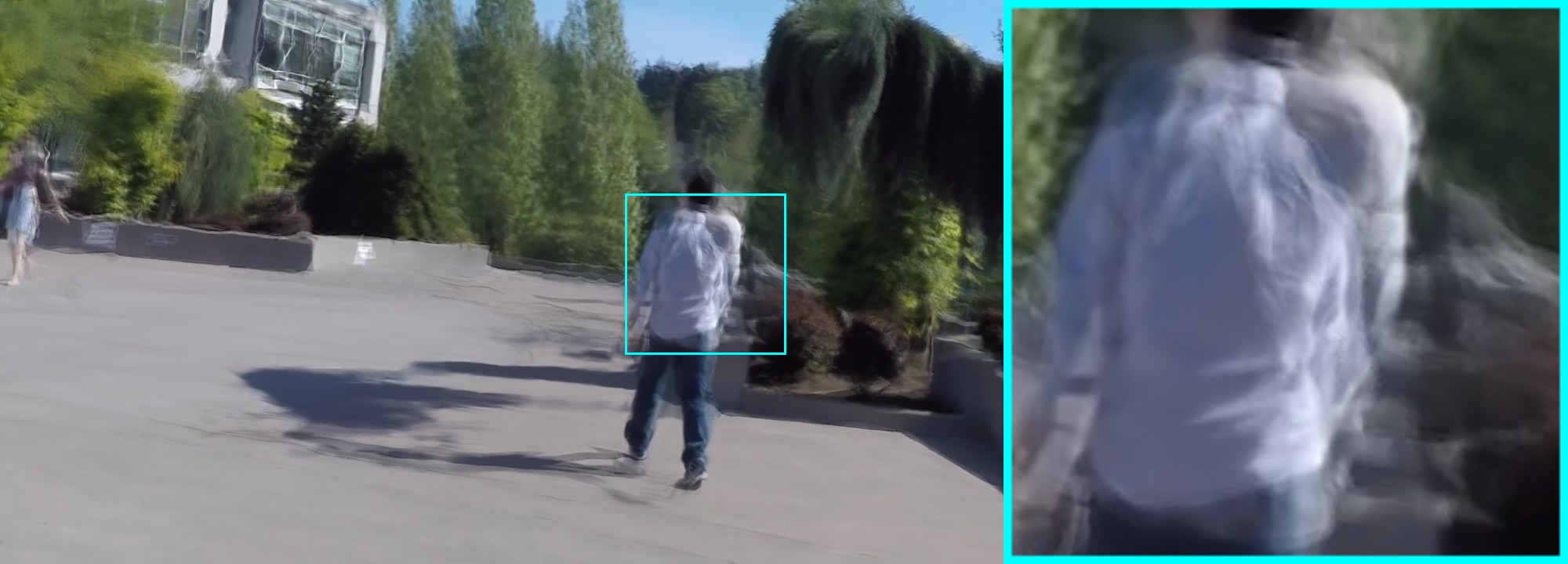}\vspace{1mm}}
		{\includegraphics[width=\linewidth]{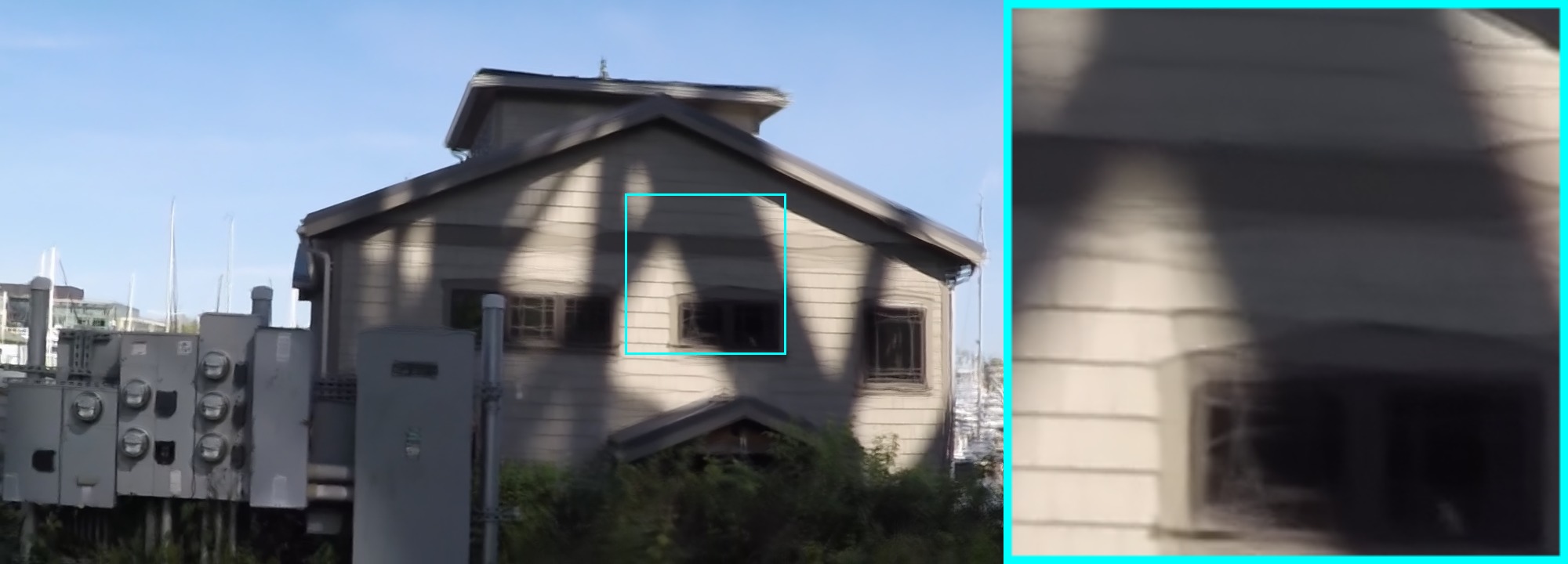}\vspace{1mm}}
		{\includegraphics[width=\linewidth]{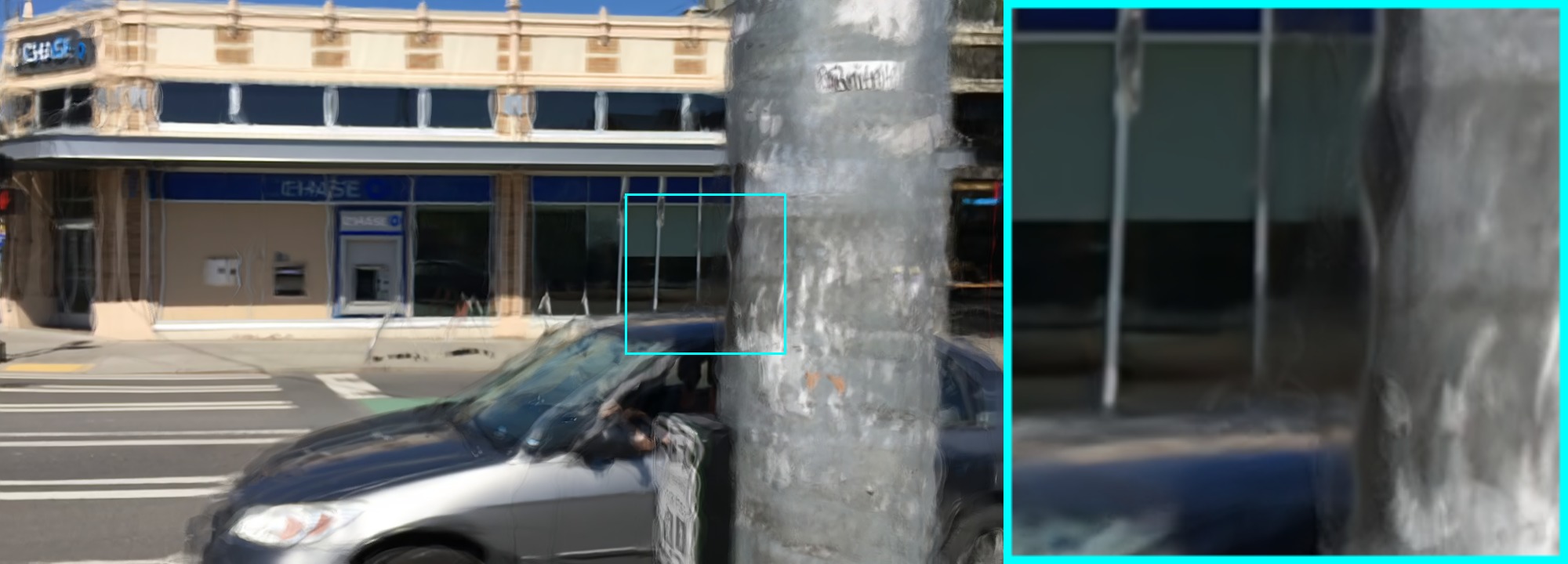}}
		\vspace{1cm}
		\centerline{AdaCoF}
		\vspace{0.1cm}
	\end{minipage}
	\begin{minipage}[htb]{0.135\linewidth}
		\vspace{1mm}
		\centering
		{\includegraphics[width=\linewidth]{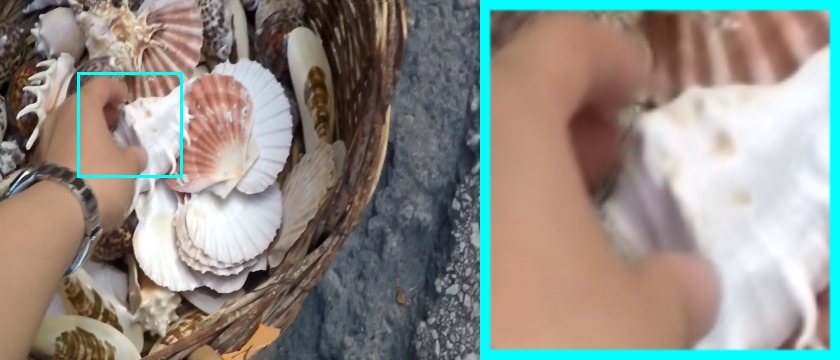}\vspace{1mm}}
		{\includegraphics[width=\linewidth]{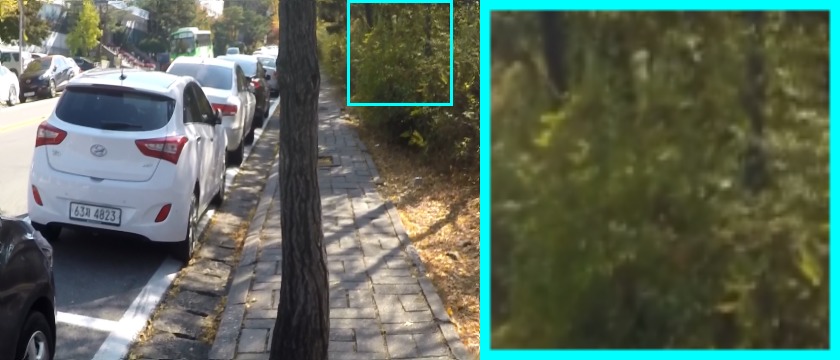}\vspace{1mm}}
		{\includegraphics[width=\linewidth]{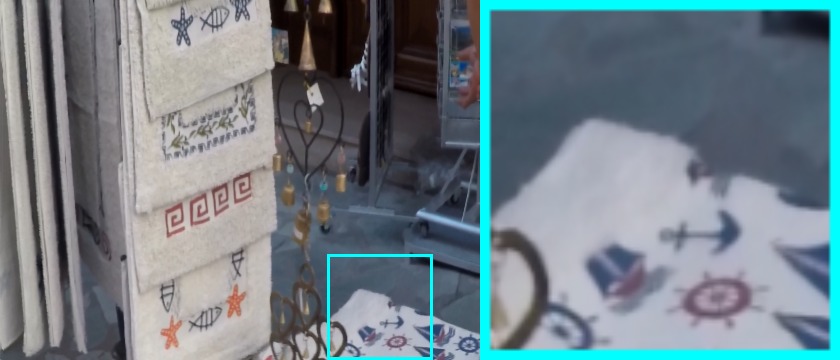}\vspace{1mm}}
		{\includegraphics[width=\linewidth]{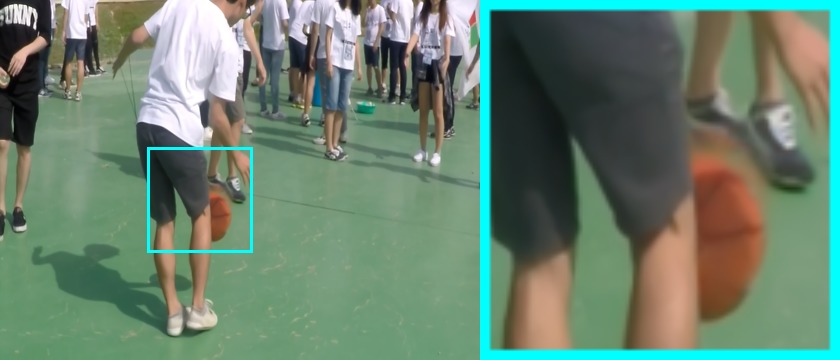}\vspace{1mm}}
		{\includegraphics[width=\linewidth]{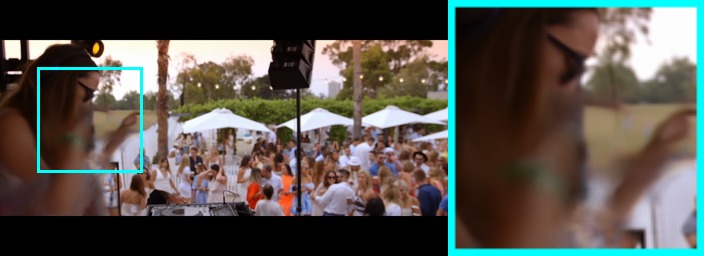}\vspace{1mm}}
		{\includegraphics[width=\linewidth]{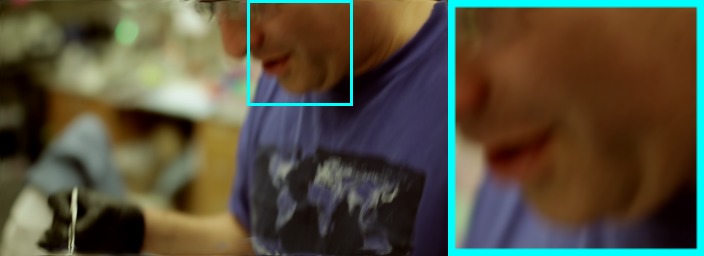}\vspace{1mm}}
		{\includegraphics[width=\linewidth]{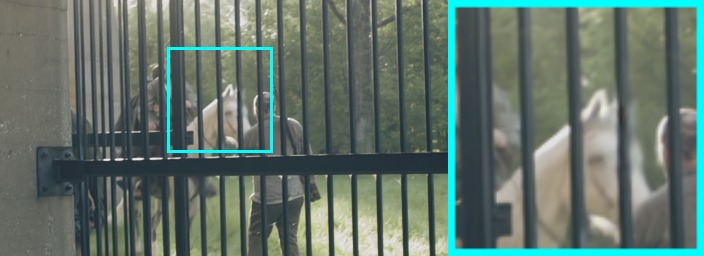}\vspace{1mm}}
		{\includegraphics[width=\linewidth]{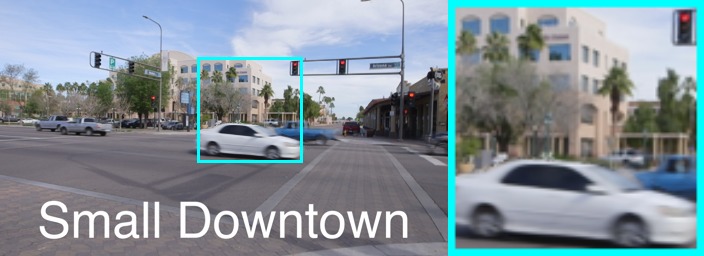}\vspace{1mm}}
		{\includegraphics[width=\linewidth]{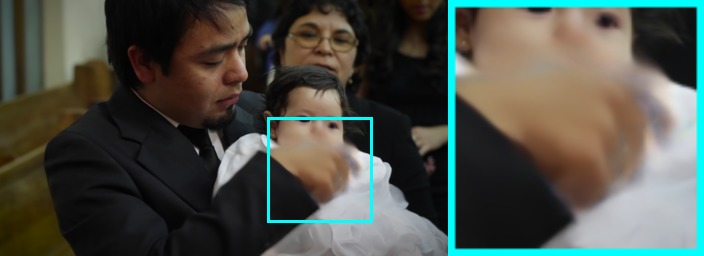}\vspace{1mm}}
		{\includegraphics[width=\linewidth]{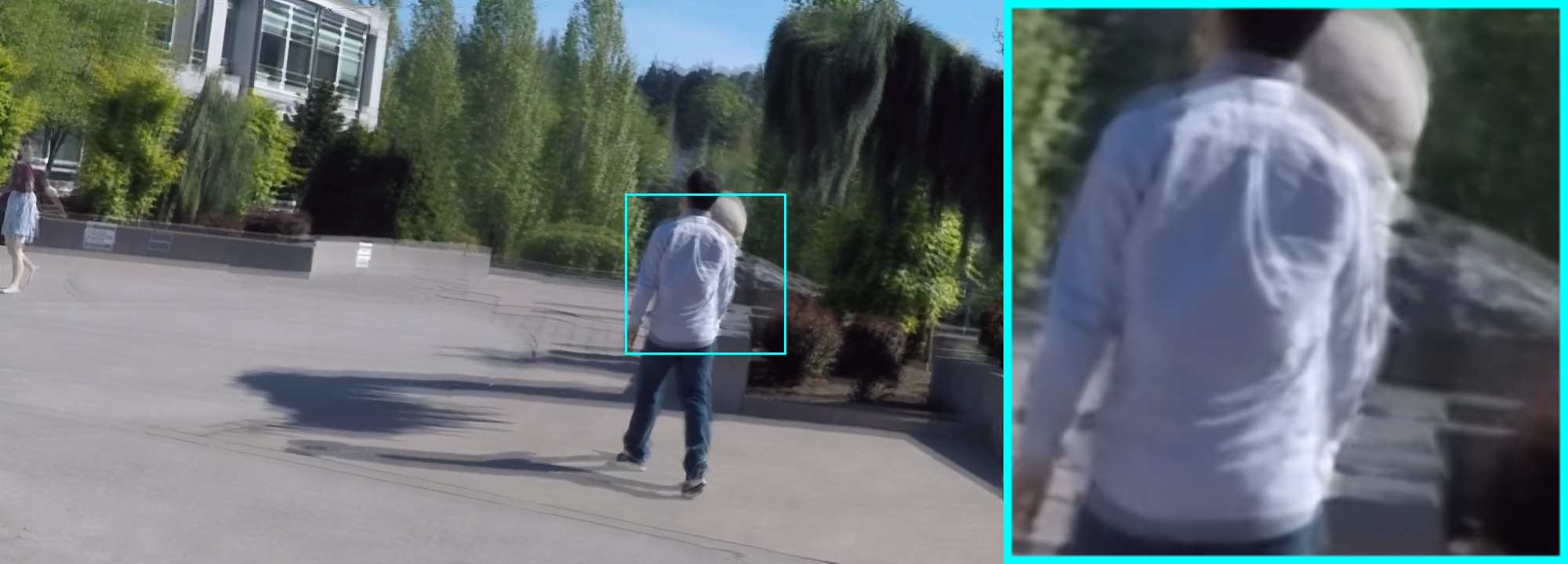}\vspace{1mm}}
		{\includegraphics[width=\linewidth]{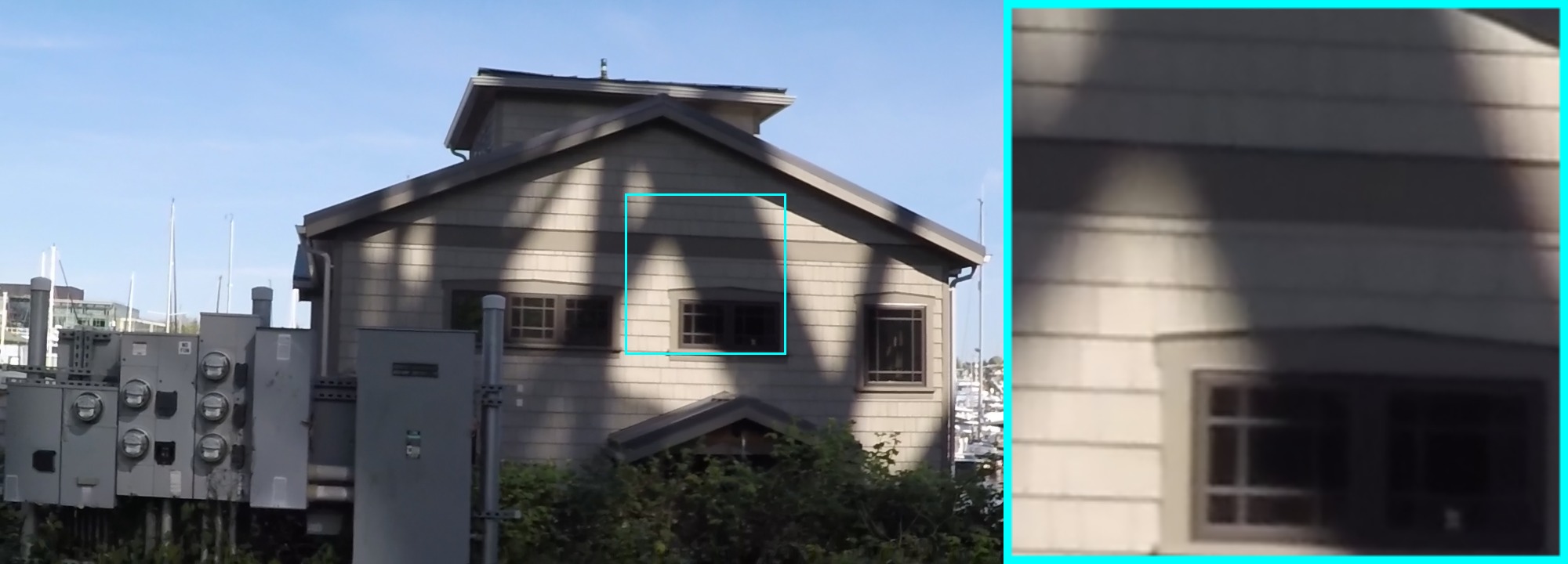}\vspace{1mm}}
		{\includegraphics[width=\linewidth]{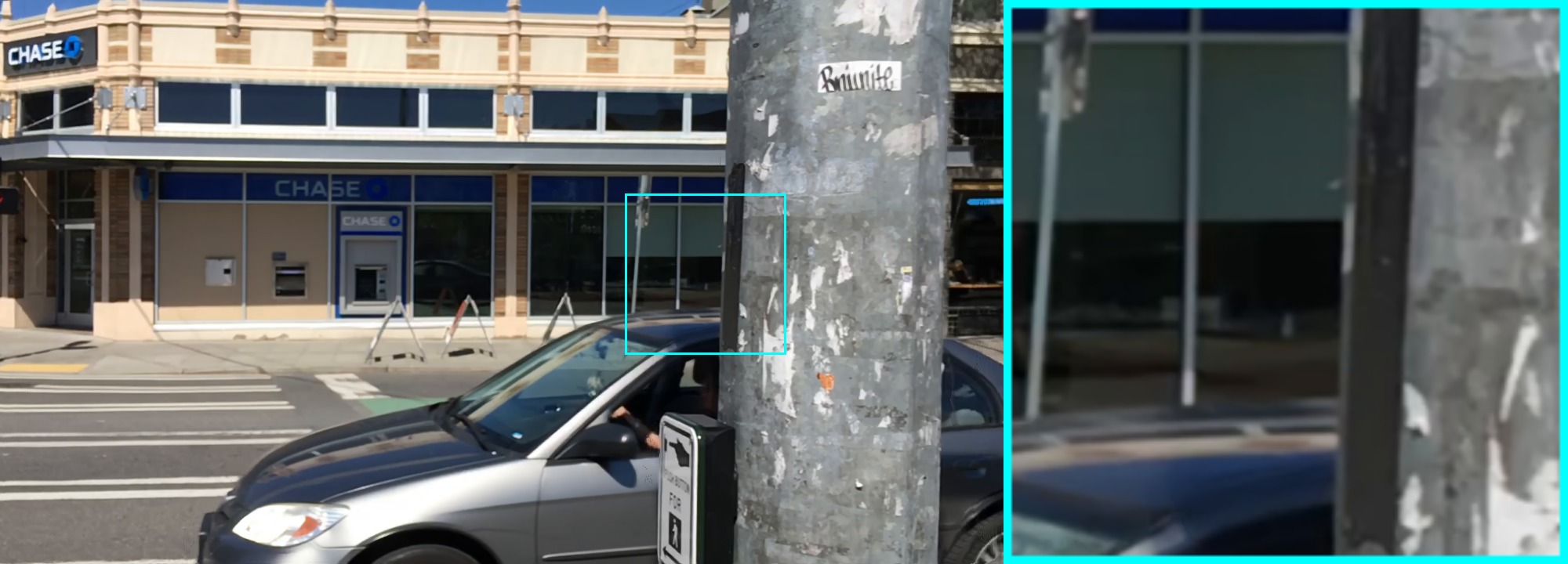}}
		\vspace{1cm}
		\centerline{Ours-Poly}
		\vspace{0.1cm}
	\end{minipage}
	\vspace{-12mm}
	\caption{Qualitative comparisons of different VFI algorithms.}
	\label{fig:qualitative comparisons}
\end{figure*}

Fig.~\ref{fig:beyond limits} provides visual examples of the results. It can be seen that compared to 1D inv, standard polynomial interpolation provides a better reconstruction in the texture regions, which usually contain a fair amount of sampling points beyond limits. In contrast, clamped polynomial interpolation performs considerably worse than the standard one in these regions. Similar phenomena can be observed for images in different datasets. In summary, polynomial interpolation is able to generate sampling points beyond upper and lower limits, and these sampling points contribute positively to the synthesis of the texture regions of the images, which helps to improve the overall performance.

\begin{figure*}[t]
	\centering
	\begin{minipage}[h]{0.45\linewidth}
		\centering
		\includegraphics[width=\linewidth]{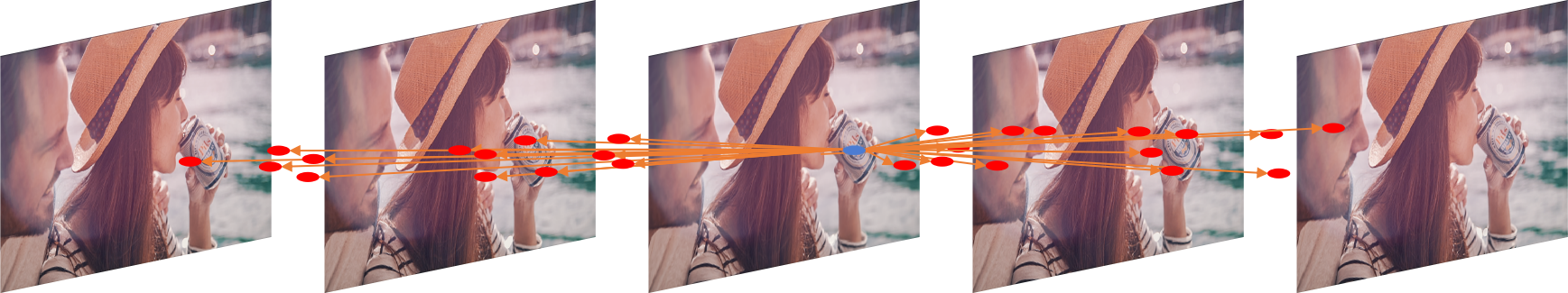}
		\scriptsize{(a)}
	\end{minipage}
	\hspace{5mm}
	\begin{minipage}[h]{0.45\linewidth}
		\centering
		\includegraphics[width=\linewidth]{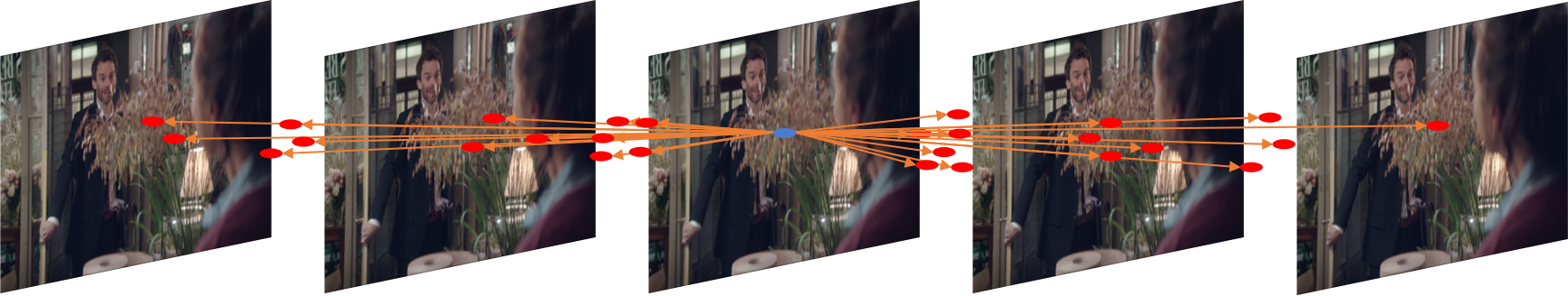}
		\scriptsize{(c)}
	\end{minipage}
	\begin{minipage}[h]{0.45\linewidth}
		\centering
		\includegraphics[width=\linewidth]{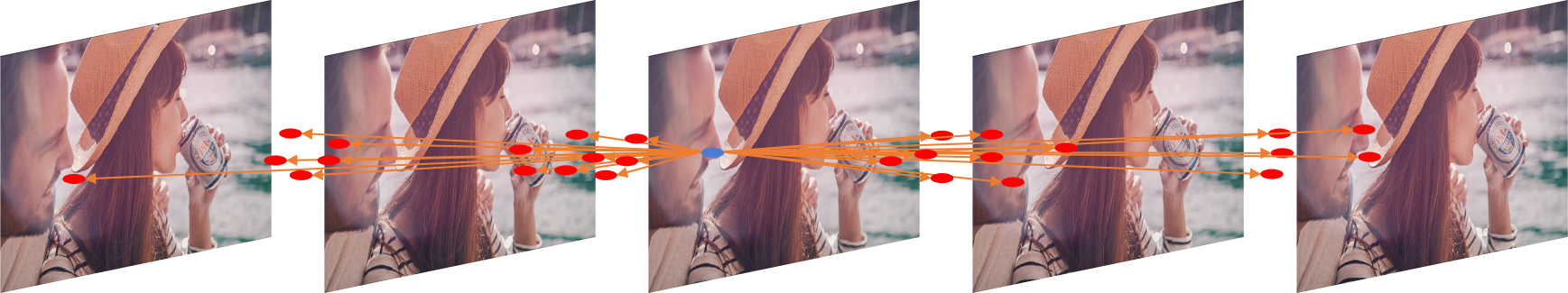}
		\scriptsize{(b)}
	\end{minipage}
	\hspace{5mm}	
	\begin{minipage}[h]{0.45\linewidth}
		\centering
		\includegraphics[width=\linewidth]{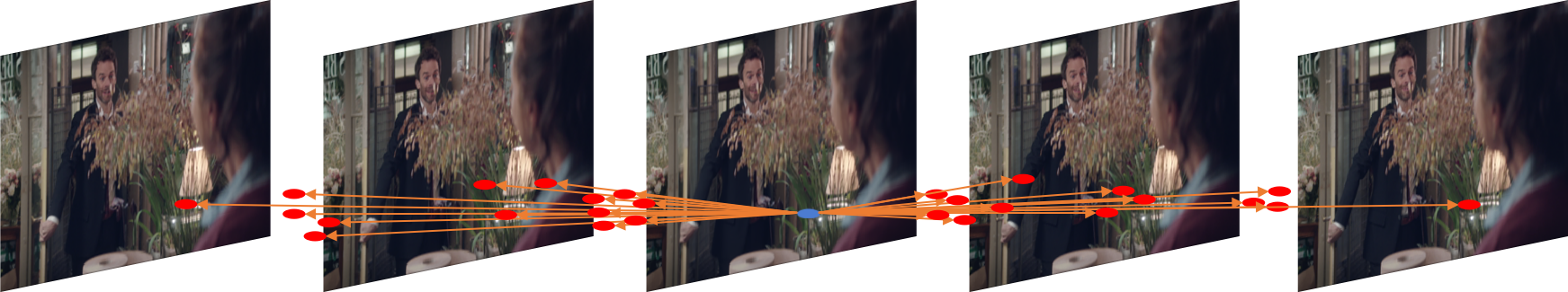}
		\scriptsize{(d)}
	\end{minipage}
	\\
	\vspace{5mm}
	\begin{minipage}[h]{0.45\linewidth}
		\centering
		\includegraphics[width=\linewidth]{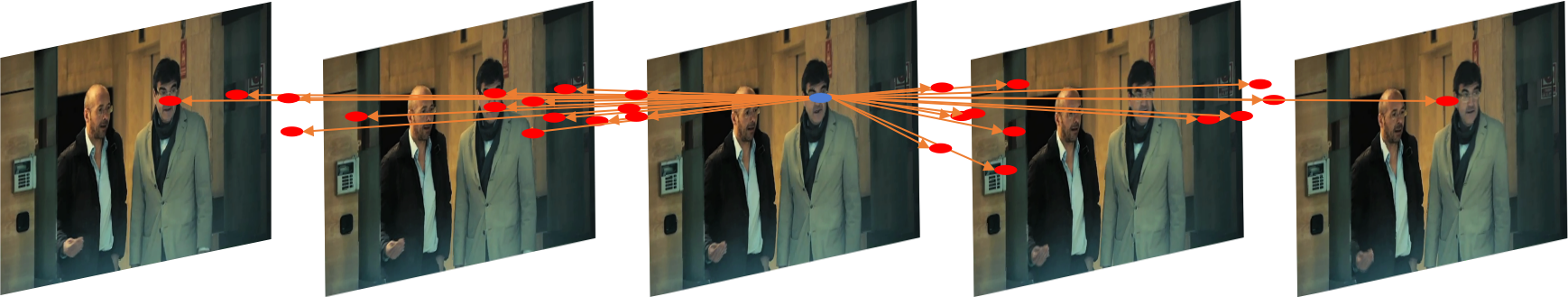}
		\scriptsize{(e)}
	\end{minipage}
	\hspace{5mm}
	\begin{minipage}[h]{0.45\linewidth}
		\centering
		\includegraphics[width=\linewidth]{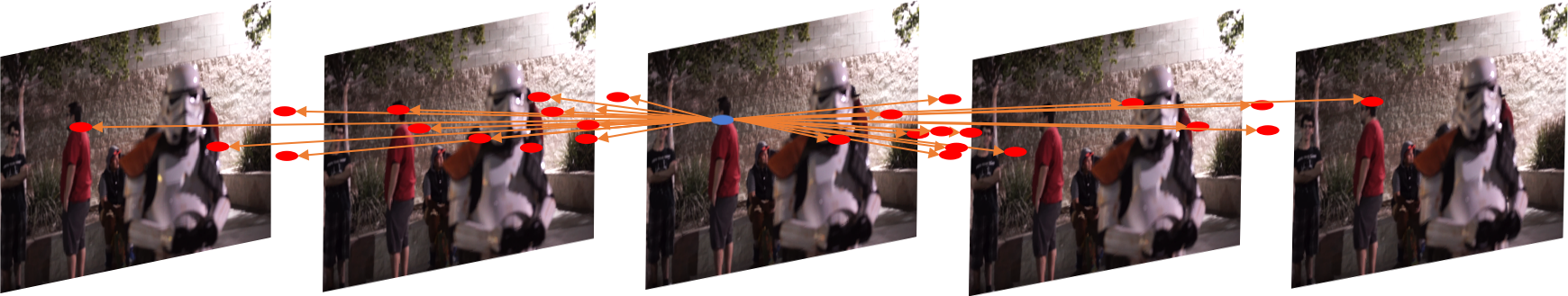}
		\scriptsize{(g)}
	\end{minipage}
	\\
	\begin{minipage}[h]{0.45\linewidth}
		\centering
		\includegraphics[width=\linewidth]{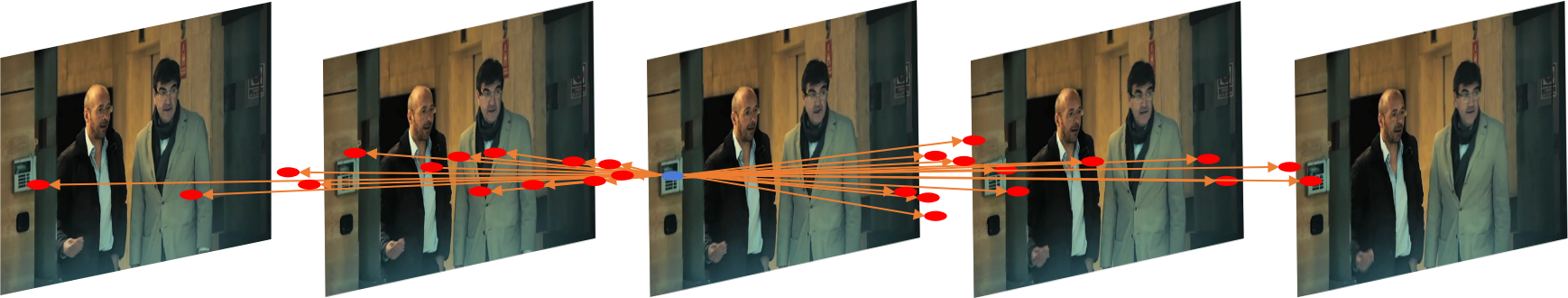}
		\scriptsize{(f)}
	\end{minipage}
	\hspace{5mm}	
	\begin{minipage}[h]{0.45\linewidth}
		\centering
		\includegraphics[width=\linewidth]{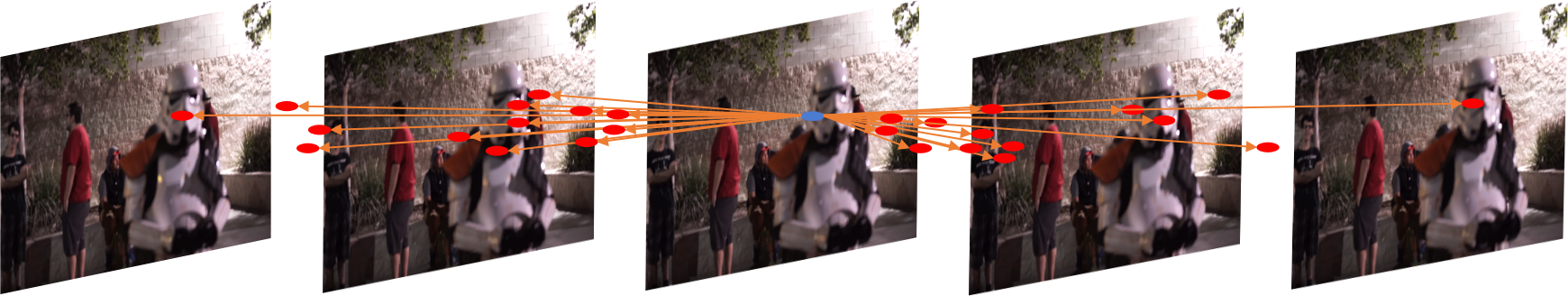}
		\scriptsize{(h)}
	\end{minipage}
	\caption{Visualization of sampling points in GDCM when $t=1.5$. Here ((a), (b)), ((c), (d)), ((e), (f)), and ((g), (h)) illustrate two different pixels in a same target intermediate frame and their associated sampling points respectively. It can be seen that sampling points are not exclusively located between $I_1$ and $I_2$. Indeed, there are some between $I_1$ and $I_2$, and some between $I_2$ and $I_3$. This indicates that the information from $I_1$ and $I_2$ is more significant for synthesizing $I_t$, but $I_0$ and $I_3$ also contribute to the synthesized result.
	}
	\label{fig:visualization}
\end{figure*}

\subsection{Comparison with the State-of-the-Art}
\label{sec:compare}
We compare our best-performing GDConvNet (Ours-Poly) with the state-of-the-art VFI algorithms on the aforementioned three evaluation datasets. Specifically, the following ones are chosen for comparison: the phase-based method (Phase) \cite{meyer2015phase}, separable adaptive convolution (SepConv) \cite{niklaus2018video}, deep voxel flow (DVF) \cite{ilg2017flownet}, SuperSlomo (Slomo) \cite{jiang2018super}, quadratic video interpolation (QVI) \cite{qvi_nips19}, and adaptive collaboration of flows (AdaCoF) \cite{lee2020adacof}. Since these methods just use two frames ($I_1$, $I_2$) to synthesize the target frame\footnote{Although 4 frames are employed in QVI, only 2 of them are directly involved in predicting the target frame.}, we also provide a degraded version of our method (Ours-Poly*) with  4 frames ($I_0$, $I_1$, $I_2$, $I_3$) for offset generation and 2 frames ($I_1$, $I_2$) for target frame prediction. For fair comparison,  DVF, Slomo, QVI, and AdaCoF are retrained on our training dataset. As the SepConv training code is not available, we choose to directly evaluate the original SepConv model.

In Table \ref{tab:comparison with SOTA}, we quantitatively compare our method with the state-of-the-art methods on the evaluation datasets under two well-known objective image quality metrics, PSNR and SSIM. It can be seen that  although it suffers from some performance degradation with respect to Ours-Poly, Ours-Poly* still performs on par with QVI
(which is 6 times as large as Ours-Poly* in terms of model size)
and surpasses other methods by a visible margin.
As for Ours-Poly, it shows a significant improvement over its degraded counterpart due to the complete freedom in exploiting the given frames,
and ranks consistently at the top in Table \ref{tab:comparison with SOTA} (except for  the Gopro dataset on which it comes in a close second in terms of the SSIM value). Overall, our method has a clear advantage under joint consideration
of cost and performance.

Fig.~\ref{fig:qualitative comparisons} shows some qualitative comparisons. It can be seen that our method produces clearer and sharper results. For example, on the first row, our method is capable of generating smooth edges around the hand compared with that of Phase, DVF, SepConv, Slomo, QVI, and AdaCoF.

\subsection{Ablation Study}
In our ablation studies, we adopt polynomial interpolation and consider a simplified version of GDConvNet in which the CEM and the associated GDCM, as well as the PM, are removed. This simplification greatly reduces the training time and, more importantly, enables us to focus on the most essential aspects of GDConvNet. 

\subsubsection{Generalized Deformable Convolution Module}
In order to validate the effectiveness of our design, we 
compare the proposed GDConv with DConv (more precisely, spatially-adaptive DConv or modulated DConv) adopted by \cite{lee2020adacof}, as well as several variants of GDConv.

{\color{orange}\begin{table}[h]
	\begin{tabular}{c|ccc}
		\toprule
		Method & \hspace{3mm} Sampling points & PSNR & SSIM\\
		\hline
		DConv &\hspace{3mm} $25$ & $32.82$ &$0.9236$\\
		\hline
		GDConv &\hspace{3mm} $1$ &$33.40$ &$0.9342$\\
		GDConv &\hspace{3mm} $9$ &$33.98$ &$0.9414$\\
		GDConv &\hspace{3mm} $25$ &$\textbf{34.20}$ &$\textbf{0.9436}$\\
		GDConv &\hspace{3mm} $36$ &$34.17$ &$0.9430$\\
		\hline
		Variant (a) &\hspace{3mm} $25$ & $32.99$& $0.9274$ \\
		Variant (b) &\hspace{3mm} $25$ & $33.24$& $0.9310$ \\
		Variant (c) &\hspace{3mm} $25$ & $33.92$& $0.9410$ \\
		Variant (d) &\hspace{3mm} $25$ & $34.06$& $0.9418$ \\
		Variant (e) &\hspace{3mm} $25$ & $\textbf{34.20}$ &$\textbf{0.9436}$\\
		\bottomrule
	\end{tabular}
	\caption{\label{tab:GDConv}Comparisons of DConv, GDConv with different numbers of sampling points, and some variants of GDConv.}
\end{table}}

\noindent \textbf{Superiority of GDConv over DConv}:
As mentioned earlier, the proposed GDConv is able to fully exploit the given source frames in accordance with their relevance to the target intermediate frame in terms of temporal distance. In contrast, the performance of DConv is limited by the inflexibility in choosing the number of sampling points from each source frame. For instance, consider the case where $4$ consecutive frames are used for VFI and the convolution kernel size is set to $3$. DConv is constrained to select $9$ sampling points from each frame. This is inefficient from the perspective of resource allocation since the source frames closer to the target intermediate frame in time are conceivably more informative and should receive more attention. In this sense, the proposed GDConv is more desirable as it is endowed with complete freedom to select sampling points in space-time. Specifically, in GDConv, the number of sampling points in each frame is adjustable according to the significance of that frame in synthesis. More importantly, sampling points are not even required to lie exactly on the source frames, and are allowed to be anywhere in the spatio-temporal domain specified by their associated parameters $\triangle x_n$, $\triangle y_n$ and $z_n$ (see Fig.~\ref{fig:visualization} for some visual results). This mechanism is especially important for VFI since it is better suited to cope with complex and irregular inter-frame motions.
In Table \ref{tab:GDConv}, we provide quantitative comparisons of DConv and GDConv. Here the number of input source frames is $4$. In GDConv, the number of sampling points is set to $36$. For fair comparison, the kernel size in DConv is chosen to be $3$; thus, there are $3 \times 3 \times 4=36$ sampling points in total, as well. It is evident that the proposed GDConv achieves better performance in terms of the PSNR and SSIM metrics.

\noindent \textbf{Importance of Spatio-Temporal  Freedom}:
We consider the following $4$ variants of GDConv to illustrate the importance of spatio-temporal freedom for sampling points.
	
	\noindent (a) No spatio-temporal freedom: $(\triangle x_n^i,\triangle y_n^i)$, $i\in\{0,1,2,3\}$, are identical and fixed to be a distinct point in a $5\times 5$ grid $\{(-2,-2),(-2,-1),\cdots,(2,2)\}\}$, and $z_n = 1.5$. 
	
	\noindent (b) Limited spatial freedom, no temporal freedom: $(\triangle x_n^i,\triangle y_n^i)$, $i\in\{0,1,2,3\}$, are identical but adaptive, and $z_n = 1.5$.
	
	\noindent (c)  Limited spatial freedom, complete temporal freedom: $(\triangle x_n^i,\triangle y_n^i)$, $i\in\{0,1,2,3\}$, are identical but adaptive, and $z_n$ is adaptive.
	
	\noindent (d) Complete spatial freedom, no temporal freedom: $(\triangle x_n^i,\triangle y_n^i)$, $i\in\{0,1,2,3\}$, can be different from each other and are individually adaptive, and $z_n=1.5$.
	
	\noindent (e) Complete spatio-temporal freedom: $(\triangle x_n^i,\triangle y_n^i)$, $i\in\{0,1,2,3\}$, can be different from each other and are individually adaptive, and $z_n$ is also adaptive.
	
	The results of the experiment are shown in Table~\ref{tab:GDConv}. One can easily find that the performance rises progressively with the availability of every additional degree of freedom. It is worth noting that the temporal parameter $z_n$ is better interpreted as being effective time instead of physical time. Indeed, forcing $z_n=1.5$ limits the degrees of freedom and jeopardizes the performance.

\noindent \textbf{Choice of the Number of Sampling Points}:
We further investigate how to choose the number of sampling points in GDConv. As shown in Table \ref{tab:GDConv}, as the number of sampling points increases, the performance improves initially, but becomes saturated eventually. In particular, using more than $36$ sampling points does not further enhance the quality of synthesized frames.

\begin{table}[h]
	\caption{\label{tab:reference} Comparisons for different numbers of reference frames (with the number of generation frames set to be the same as that of reference frame). }
	\center
	\begin{tabular}{c|cc}
		\hline
		{Reference Frames}
		& \hspace{5mm} PSNR & \hspace{5mm} SSIM \\    %
		\hline
		$I_1$, $I_2$ &  \hspace{5mm} $33.69$ & \hspace{5mm} $0.9416$\\
		$I_0$, $I_1$, $I_2$ & \hspace{5mm} $33.97$  & \hspace{5mm} $0.9427$ \\
		$I_0$, $I_1$, $I_2$, $I_3$ & \hspace{5mm} $\textbf{34.20}$ & \hspace{5mm} $\textbf{0.9436}$\\
		\hline
	\end{tabular}
\end{table}

\begin{table}[h]
	\caption{\label{tab:generation} Comparisons for different numbers of generation frames (with the reference frames fixed to be $I_1$ and $I_2$). }
	\center
	\begin{tabular}{c|cc}
		\hline
		{Generation Frames} & 
		\hspace{5mm} PSNR & \hspace{5mm}
		SSIM\\  %
		\hline
		$I_1$, $I_2$ & \hspace{5mm} $33.69$ & \hspace{5mm} $0.9416$\\
		$I_0$, $I_1$, $I_2$ & \hspace{5mm} $33.84$  & \hspace{5mm} $0.9418$ \\
		$I_0$, $I_1$, $I_2$, $I_3$ & \hspace{5mm} $\textbf{34.05}$ & \hspace{5mm} $\textbf{0.9434}$\\
		\hline
	\end{tabular}
\end{table}

\begin{table*}[t]
	\renewcommand\arraystretch{1}
	\caption{\label{tab:comparison on vimeo and ucf}Quantitative comparisons on Vimeo90K interpolation test set, UCF101 dataset and Middlebury-Other dataset, where the first place and second place are highlighted in red and blue, respectively.}
	\footnotesize
	\center
	\renewcommand\tabcolsep{15.0pt}
	\resizebox{\textwidth}{!}{\begin{tabular}{cccccccc}
			\toprule
			\multicolumn{1}{c}{\multirow{2}*{Method}} 
			&{\#Parameters} 
			&\multicolumn{2}{c}{UCF101} 
			&\multicolumn{2}{c}{Vimeo90K}
			&\multicolumn{1}{c}{Middlebury} \\  
			\cmidrule(r){3-4} 
			\cmidrule(r){5-6}
			\cmidrule(r){7-7}
			& (million) & PSNR & SSIM & PSNR & SSIM & IE\\
			\hline
			MIND & $7.60$  & $33.93$ & $0.9661$ & $33.50$ & $0.9429$ &$3.35$\\
			DVF & $3.80$ & $34.12$ & $0.9631$ & $31.54$ & $0.9462$ &$7.75$\\
			ToFlow & $1.07$ & $34.58$ & $0.9667$ & $33.73$ & $0.9682$ &$2.51$\\
			SepConv-Lf & $21.6$ & $34.69$ & $0.9655$ & $33.45$ & $0.9674$ &$2.44$\\
			SepConv-L1 & $21.6$   & $34.78$ & $0.9669$ & $33.79$ & $0.9702$ &$2.27$\\
			MEMC-Net & $70.3$   & $34.96$ & $0.9682$ & $34.29$ & $0.9739$ &$2.12$\\
			DAIN & $24.0$  & $\textbf{{\color{blue}{34.99}}}$ & $\textbf{{\color{red}{0.9683}}}$ & $\textbf{{\color{blue}{34.71}}}$ & $\textbf{{\color{red}{0.9756}}}$ & $\textbf{{\color{blue}{2.04}}}$ \\
			AdaCoF & $21.8$  & $\textbf{{\color{blue}{34.99}}}$& $0.9682$ & $33.43$ & $0.9677$ &$2.43$\\
			\hline
			Ours & $5.6$&  $\textbf{{\color{red}{35.16}}}$& $\textbf{{\color{red}{0.9683}}}$ & $\textbf{{\color{red}{34.99}}}$& $\textbf{{\color{blue}{0.9750}}}$ & $\textbf{{\color{red}{2.03}}}$\\
			\bottomrule
			
	\end{tabular}}
	
\end{table*}

\subsubsection{Input Length and Offset Generation}
So far, except for the degraded version in Section \ref{sec:compare}, we have assumed that all $4$ source frames  $I_0$, $I_1$, $I_2$, and $I_3$  participate in generating offsets (as well as $z_n$ and $\triangle m_n$) and in predicting the target intermediate frame $I_{1.5}$. It is interesting to study how the proposed method performs if one only utilizes a subset of source frames. In fact, our framework is flexible enough to allow the use of different subsets of source frames for offset generation and frame prediction separately. For clarity, we shall refer to source frames used for generating offsets as generation frames and those directly involved in predicting the target intermediate frame as reference frames. For example, if we use $I_0$, $I_1$, $I_2$ to generate offsets for $I_1$ and $I_2$, which are subsequently leveraged to predict $I_{1.5}$, then $I_0$, $I_1$, $I_2$ are generation frames while the latter two are reference frames. We first study the scenario with the same subset of source frames used for both purposes. It is clear from Table \ref{tab:reference} that the VFI result improves progressively with the increase in the number of reference frames (as well as generation frames). We further investigate the scenario where reference frames and generation references are not necessarily the same. Specifically, we fix $I_1$ and $I_2$ to be reference frames, and consider various combinations of generation frames. It can be seen from Table \ref{tab:generation} that increasing the number of generation frames leads to better performance. This is consistent with a similar finding regarding flow-based methods: namely, it is profitable to have three or more generation frames as that opens the door for exploiting higher-order approximation of motion trajectories (instead of relying on linear approximation, which is basically the only available choice in the case with just two generation frames). Finally, comparing the corresponding rows in Table \ref{tab:reference} and Table \ref{tab:generation} reveals that VFI can also benefit from an increase in the number of reference frames (when the number of generation frames is fixed). 

\subsection{Failure Case Analysis}
Our method is trained in a purely data-driven manner to learn motion estimation. As such, it is able to handle complex motion patterns that cannot be characterized by simple mathematical models.
On the other hand, the success of our method depends critically on the quality of the training dataset, which should ideally contain extensive motion patterns to ensure sufficient coverage. The performance of our method tends to degenerate when the motion patterns encountered in the evaluation dataset deviate significantly from those in the training dataset. Fig.~\ref{fig:failure case} provides some examples where object motions are atypical with respect to the training dataset.
It can be seen that the VFI results produced by our method are somewhat blurry (albeit still slightly better than those of QVI, which is the best known mathematical-model-based method).

\begin{figure}[h]
	
	\centering
	\begin{minipage}[b]{0.45\linewidth}
		\centering
		\includegraphics[width=\linewidth]{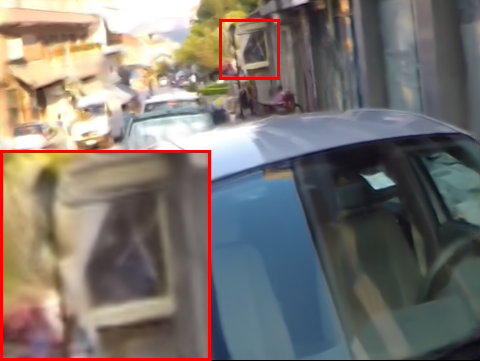}\\
		\vspace{1mm}
		\includegraphics[width=\linewidth]{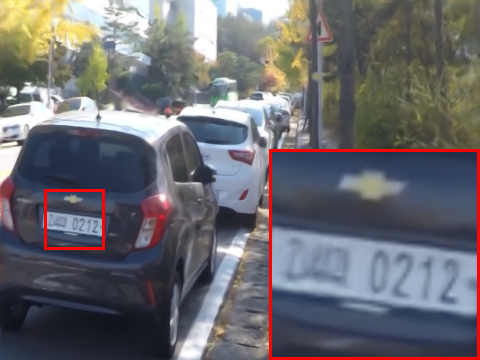}
		\centerline{(a) Ours}
		\vspace{0.0001cm}	
	\end{minipage}
	\begin{minipage}[b]{0.45\linewidth}
		\centering
		\includegraphics[width=\linewidth]{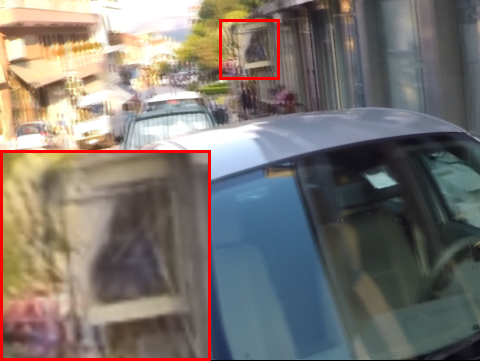}\\
		\vspace{1mm}
		\includegraphics[width=\linewidth]{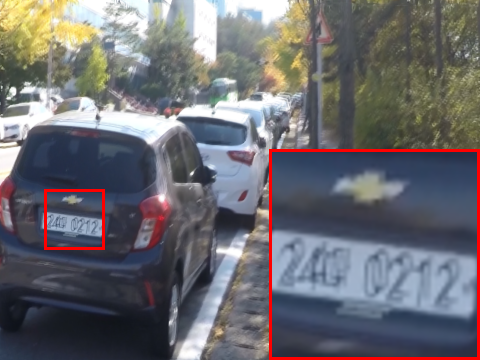}
		\centerline{(b) QVI}
		\vspace{0.0001cm}	
	\end{minipage}
	\figvspace
	\caption{Visualization of failure cases.}
	\label{fig:failure case}
\end{figure}

\section{Two-Frames VFI Experiments} \label{Sec:2-frame}
As described earlier, our method is able to handle an arbitrary number of frames. To substantiate this claim, here we conduct two-frame VFI experiments (i.e., using $I_0$ and $I_1$ to predict $I_{0.5}$).

\subsection{Implementation Details}
We adopt polynomial interpolation (or linear interpolation) and set the number of sampling points for each warped pixel to be $25$ in GDConv. The training dataset and the training strategy are described below.

\noindent \textbf{Training Dataset:}
The Vimeo90k interpolation training dataset \cite{xue2019video} is used to train our model. This training dataset is composed of $51,312$ triplets with resolution  $256 \times 448$. We use the first frame and the third frame (corresponding to $I_0$ and $I_1$, respectively) of each triplet to predict the second one (corresponding to $I_{0.5}$). We randomly crop image patches of size $256 \times 256$ for training. Horizontal and vertical flipping, as well as temporal order reversal, are performed for data augmentation.

\noindent \textbf{Training Strategy:}
This is the same as the four-frame case, except that we train our network for $20$ epochs in total. The initial learning rate remains to be $10^{-3}$, and the learning rate is reduced by a factor of two every $4$ epochs for the first $12$ epochs and by a factor of five every $4$ epochs for the last $8$ epochs. The whole training process takes about 3 days on our hardware.

\subsection{Evaluation Datasets}

Following \cite{bao2019depth}, we evaluate the proposed
 GDConvNet on three public datasets (Vimeo90k Interpolation Test Set \cite{xue2019video}, UCF101 Test Dataset\cite{soomro2012ucf101}, and Middlebury-Other Dataset \cite{baker2011database}) and compare it with the state-of-the-art.


\noindent \textbf{Vimeo90k Interpolation Test Set \cite{xue2019video}:}
This dataset consists of $3,782$ video sequences, each with $3$ frames. As in the case of the Vimeo90K interpolation training dataset, the first frame and the third frame of each sequence are leveraged to synthesize the second one. The image resolution of this dataset is $256 \times 448$.

\noindent \textbf{UCF101 Test Dataset\cite{soomro2012ucf101}:} 
The UCF101 dataset contains $379$ triplets with a large variety of human actions. The image resolution of this dataset is $256 \times 256$. 

\noindent \textbf{Middlebury-Other Dataset \cite{baker2011database}:} 
The Middlebury-Other dataset is another commonly used benchmark for VFI, which contains $12$ triplets in total. Most of the images in this dataset are of resolution $640 \times 480$. Again, we use the first frame and the third frame to predict the second one.

\subsection{Experimental Results}
We compare our GDConvNet with the state-of-the-art VFI algorithms on the aforementioned datasets. Specifically, the following ones are chosen for comparison: MIND \cite{long2016learning}, DVF \cite{ilg2017flownet}, SepConv \cite{niklaus2018video}, CtxSyn \cite{niklaus2018context}, ToFlow \cite{xue2019video}, SuperSlomo \cite{jiang2018super}, MEMC-Net \cite{bao2019memc}, DAIN \cite{bao2019depth}, and AdaCoF \cite{lee2020adacof}. 

In Table \ref{tab:comparison on vimeo and ucf}, we quantitatively compare our method with the state-of-the-art on  Vimeo90k and UCF101 under PSNR and SSIM, while Interpolation Error \cite{simonyan2014very} (IE) is used as the performance measure for the Middlebury-Other dataset. It can be seen that the proposed method performs favorably against those under consideration. Overall, our method has a clear advantage under joint consideration of cost and performance. In particular, although DAIN \cite{bao2019depth} also shows very competitive performance, its model size is about $5$ times that of our model. In addition, our method can be trained from scratch, while DAIN \cite{bao2019depth} needs to rely on a pre-trained model.

\section{Conclusion}
In this paper, a new mechanism named generalized deformable convolution is proposed to tackle the VFI problem. This mechanism unifies the essential ideas underlying flow-based and kernel-based methods and resolves some performance-limiting issues. It should be noted that the proposed mechanism is largely generic in nature, and is potentially applicable to a wide range of problems, especially those involving video data (e.g., video super-resolution, enhancement, and quality mapping). Exploring such applications is an endeavor well worth undertaking.

\section{Acknowledgment}
The authors would like to thank Prof. Tim Davidson for proofreading the manuscript.

\ifCLASSOPTIONcaptionsoff
\newpage
\fi

%

%
%
%
%
%
%


%
%
\bibliographystyle{IEEEtran}
\bibliography{IEEEabrv,egbib}

\end{document}